\documentclass{article}

\PassOptionsToPackage{numbers, compress}{natbib}
\usepackage[preprint]{neurips_2024}

\usepackage{amsmath,amsfonts,bm}

\def\eqref#1{(\ref{#1})}

\def\1{\bm{1}}

\DeclareMathAlphabet{\mathsfit}{\encodingdefault}{\sfdefault}{m}{sl}
\SetMathAlphabet{\mathsfit}{bold}{\encodingdefault}{\sfdefault}{bx}{n}

\DeclareMathOperator*{\argmax}{arg\,max}
\DeclareMathOperator*{\argmin}{arg\,min}

\usepackage[utf8]{inputenc} 
\usepackage[T1]{fontenc}    
\usepackage{hyperref}       
\usepackage{url}            
\usepackage{booktabs}       
\usepackage{amsfonts}       
\usepackage{nicefrac}       
\usepackage{microtype}      
\usepackage{xcolor}         

\usepackage{setspace}
\usepackage{enumitem}
\usepackage{subcaption}
\usepackage{graphicx}
\usepackage{wrapfig}
\usepackage{algorithm}
\usepackage{algpseudocode}
\usepackage{bbm}
\usepackage{multirow}	
\usepackage{color, colortbl}
\usepackage{comment}

\definecolor{Gray}{gray}{0.9}
\definecolor{Red}{RGB}{139, 0, 0}
\definecolor{Green}{rgb}{0,0.4,0.7}
\definecolor{Blue}{RGB}{0, 100, 150}

\hypersetup{
    colorlinks=true,
    citecolor=Blue,
    linkcolor=Red,
    urlcolor=Green,
    }

\title{Formatting Instructions For NeurIPS 2024}

\title{Cost-Sensitive Multi-Fidelity Bayesian Optimization with Transfer of Learning Curve Extrapolation}

\author{%
  Dong Bok Lee$^{1}$,
  \quad Aoxuan Silvia Zhang$^{2}$\thanks{\ Equal contribution}$\ \ $, 
  \quad Byungjoo Kim$^{1*}$,
  \quad Junhyeon Park$^{1*}$, \\
  \textbf{
  Juho Lee$^{1}$,
  \quad Sung Ju Hwang$^{1,3}$,
  \quad Hae Beom Lee} 
  \\
  $^1$KAIST,$\quad$ $^2$Korea University,$\quad$ $^3$DeepAuto,$\quad$ South Korea
  \\
  \texttt{markhi@kaist.ac.kr,\quad hae-beom.lee@mila.quebec}
   \vspace{-0.05in}
}

\begin{document}

\maketitle

\begin{abstract}
  In this paper, we address the problem of \emph{cost-sensitive multi-fidelity} Bayesian Optimization (BO) for efficient hyperparameter optimization (HPO). Specifically, we assume a scenario where users want to early-stop the BO when the performance improvement is not satisfactory with respect to the required computational cost. Motivated by this scenario, we introduce \emph{utility}, which is a function predefined by each user and describes the trade-off between cost and performance of BO. This utility function, combined with our novel acquisition function and stopping criterion, allows us to dynamically choose for each BO step the best configuration that we expect to maximally improve the utility in future, and also automatically stop the BO around the maximum utility. Further, we improve the sample efficiency of existing learning curve (LC) extrapolation methods with transfer learning, while successfully capturing the correlations between different configurations to develop a sensible surrogate function for multi-fidelity BO. We validate our algorithm on various LC datasets and found it outperform all the previous multi-fidelity BO and transfer-BO baselines we consider, achieving significantly better trade-off between cost and performance of BO.  
\end{abstract}
\section{Introduction}
\vspace{-0.05in}

Hyperparameter optimization (HPO)~\cite{bergstra2012random,bergstra2011algorithms,hutter2011sequential,snoek2012practical,cowen2022hebo,li2018hyperband,franceschi2017forward} stands as a crucial challenge in the domain of deep learning, given its importance of achieving optimal empirical performance. Unfortunately, the field of HPO for deep learning remains relatively underexplored, with many practitioners resorting to simple trial-and-error methods~\cite{bergstra2012random,li2018hyperband}. Moreover, traditional black-box Bayesian optimization (BO) approaches for HPO~\cite{bergstra2011algorithms,snoek2012practical,cowen2022hebo} have limitations when applied to deep neural networks due to the impracticality of evaluating a vast number of hyperparameter configurations until convergence, each of which may take several days. 

Recently, multi-fidelity (or gray-box) BO~\cite{li2018hyperband,falkner2018bohb,ijcai2021p296,swersky2014freeze,wistuba2022supervising,arango2023quick,kadra2024scaling,rakotoarison2024context} receives more attention to improve the sample efficiency of traditional black-box BO. Multi-fidelity BO makes use of lower fidelity information (e.g., validation accuracies at fewer training epoches) to predict and optimize the performances at higher or full fidelities (e.g., validation accuracies at the last training epoch). Unlike black-box BO, multi-fidelity BO dynamically select hyperparameter configurations even before finishing a single training run, demonstraing its ability of finding better configurations in a more sample efficient manner than black-box BO.

However, one critical limitation of the conventional multi-fidelity BO frameworks is that they are not aware of the trade-off between the cost and performance of BO.  For instance, some users may want to strongly penalize the cost of BO with respect to its performance, and in this case the BO process should focus on exploiting the current belief on good hyperparameter configurations than trying to explore new configurations.  Yet, existing multi-fidelity BO methods tend to over-explore because they usually assume a sufficiently large budget for the BO and aim to obtain the best asymptotic performance on a validation set, hence are not able to properly penalize the cost~\cite{swersky2014freeze,kadra2024scaling}. One may argue that we could lower the total BO budget and maximize the performance at that specific budget, but in practice it is hard to specify the target budget in advance as it is difficult to accurately predict the future performances of BO. Rakotoarison et al.~\cite{rakotoarison2024context} propose to randomize the target budget, but it implicitly assumes that a single optimal configuration can achieve strong performances at any stages of training, which is unrealistic~\cite{wistuba2022supervising} (e.g., regularization).

\begin{wrapfigure}{R}{0.58\textwidth}
\vspace{-0.1in}
\begin{subfigure}[c]{0.285\textwidth}
\centering
    \includegraphics[width=0.99\textwidth]{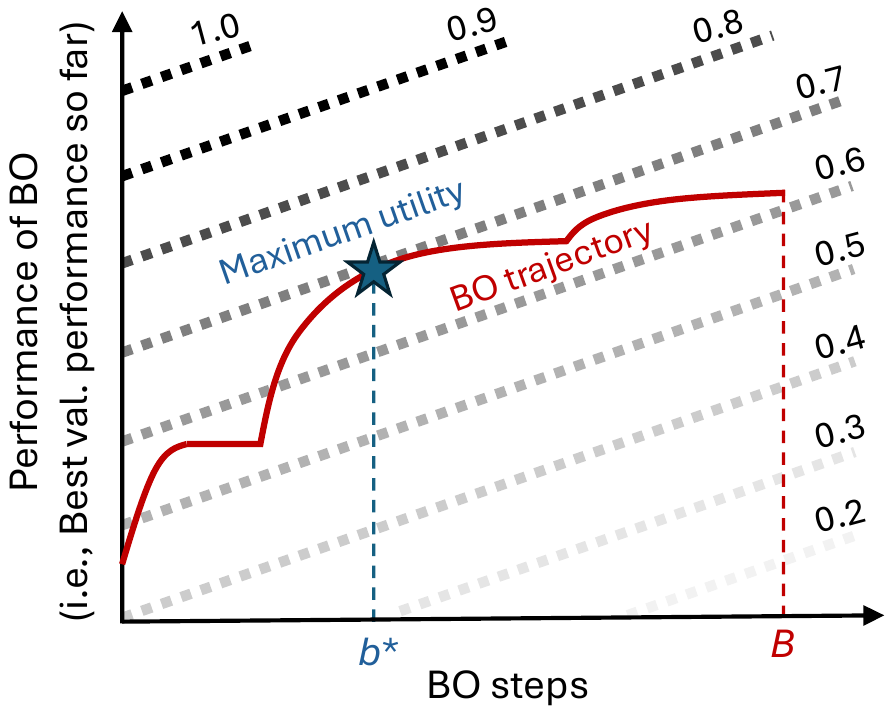}
    \vspace{-0.22in}
\subcaption{}
\label{fig:csbo}
\end{subfigure}
\begin{subfigure}[c]{0.285\textwidth}
\centering
    \includegraphics[width=0.99\textwidth]{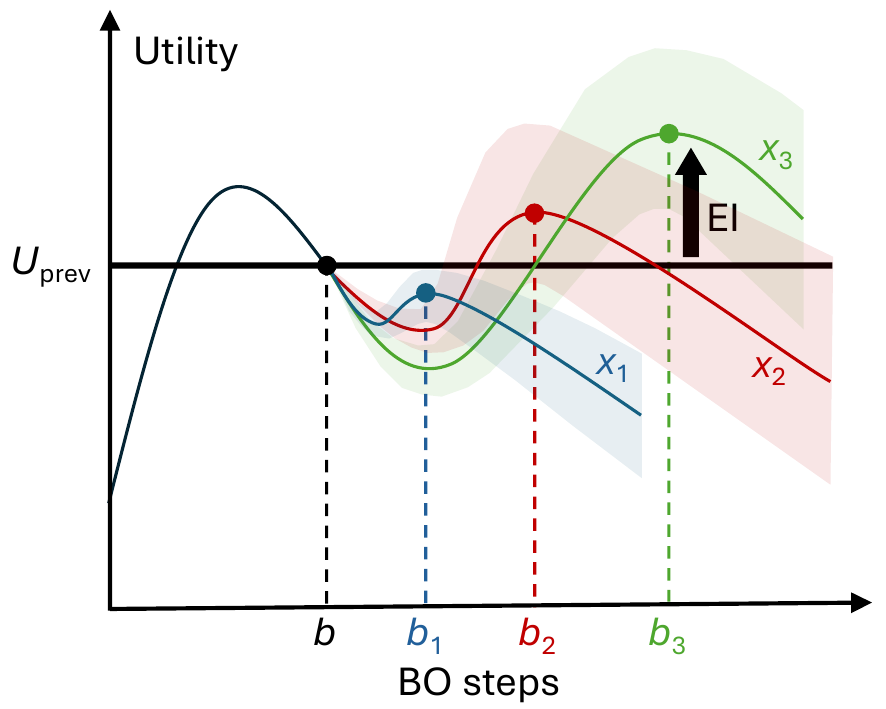}
    \vspace{-0.22in}
\subcaption{}
\label{fig:utility}
\end{subfigure}
\vspace{-0.1in}
\caption{\small \textbf{(a)} A utility function shown in the dotted black lines. The red curve shows a BO trajectory from which we determine the maximum utility ($\approx$ 0.7) and when to stop ($b^*$). \textbf{(b)} An illustration of selecting the best configuration at each BO step. Notice, the y-axis is utility. Starting from the current BO step $b$, we extrapolate the LCs with the three configurations $x_1,x_2,x_3$ (shown in the solid curves with colors and the shaded area), and then select $x_3$ which achieves at $b_3$ the maximum expected improvement (EI) of utility over the previous utility $U_\text{prev}$.}
\vspace{-0.1in}
\end{wrapfigure}

Therefore, in this paper we introduce a more sophisticated notion of cost-sensitivity for multi-fidelity BO. Specifically, we assume that it is easier to specify the trade-off between cost and performance of multi-fidelity BO, than to know the proper target budget in advance or resort to the unrealistic assumption mentioned above. We call this trade-off \emph{utility}. It is a function predefined by each user and describes users' own preferences about the trade-off. It has higher values as cost decreases and performance increases, and vice versa (Fig.~\ref{fig:csbo}). Also, some utility functions may strongly penalize the amount of BO budgets spent, while others may weakly penalize or not penalize at all as with the conventional multi-fidelity BO. We explicitly maximize this utility by dynamically selecting hyperparamter configurations which are expected to maximally improve it in future (Fig.~\ref{fig:utility}), and also automatically terminating the BO around the maximum utility (Fig.~\ref{fig:csbo}), instead of terminating at an arbitrary target budget. 

Solving this problem requires our multi-fidelity BO frameworks to have the following capabilities. Firstly, they should support \textbf{freeze-thaw BO}~\cite{swersky2014freeze,rakotoarison2024context}, an advanced form of multi-fidelity BO in which we can dynamically pause (freeze) and resume (thaw) hyperparameter configurations based on future performances extrapolated from a set of partially observed learning curves (LCs) with various configurations. Such efficient and sensible allocation of computational resources suits well for our purpose of finding the best trade-off between cost and performance of multi-fidelity BO. 
Secondly, freeze-thaw BO requires its surrogate function to be equipped with a \textbf{good LC extrapolation} mechanism~\cite{adriaensen2024efficient,rakotoarison2024context,kadra2024scaling}. In our case, it is crucial for making a good probabilistic inference on future utilities with which we dynamically select the best configuration and accurately early-stop the BO. Lastly, since we assume that users want to stop the BO as early as possible, LC extrapolation should be accurate even at the early stages of BO. Therefore, we should make use of \textbf{transfer learning} to maximally improve the sample efficiency of BO~\cite{arango2023quick} and to prevent inaccurate early-stopping when there are only few or even no observations in the BO.

Based on those criteria, we introduce our novel \textbf{C}ost-sensitive \textbf{M}ulti-fidelity \textbf{BO} (CMBO) that makes use of transfer learning and in-context LC extrapolation. We first introduce the acquisition function and stopping criteria specifically developed for our framework, and explain how to achieve with them a good trade-off between cost and performance of multi-fidelity BO (\S\ref{sec:ustar}).
Also, based on a recently introduced Prior-Fitted Networks (PFNs)~\cite{muller2021transformers,adriaensen2024efficient} for in-context Bayesian inference, we explain how to train a PFN with the existing LC datasets to develop a sample efficient in-context surrogate function for freeze-thaw BO that can also effectively capture the correlations between different hyperparameter configurations (\S\ref{sec:transfer_learning}). Lastly, we empirically demonstrate the superiority of CMBO on a set of diverse user preferences and three HPO benchmarks, showing that it significantly outperforms all the previous multi-fidelity BO and the transfer-BO baselines we consider (\S\ref{sec:experiment}).



\vspace{-0.05in}
\section{Related Work}
\vspace{-0.05in}

\paragraph{Multi-fidelity HPO.} 
Unlike traditional black-box approaches for HPO~\cite{bergstra2012random,hutter2011sequential,bergstra2011algorithms,snoek2012practical,snoek2015scalable,snoek2014input,cowen2022hebo,muller2023pfns4bo}, multi-fidelity (or gray-box) HPO aims to optimize hyperparameters in a sample efficient manner by utilizing low fidelity information (e.g., validation set performances with smaller training dataset) as a proxy for higher or full fidelities~\cite{swersky2013multi,klein2017fast,poloczek2017multi,li2020multi}, dramatically speeding up the HPO. In this paper, we focus on making use of performances at fewer training epochs to better predict/optimize the performances at longer training epochs. One of the well-known examples is Hyperband~\cite{li2018hyperband}, a bandit-based method that randomly selects a set of random hyperparameter configurations, and stops poorly performing ones using successive halving~\cite{karnin2013almost} even before reaching the last training epoch. While Hyperband shows much better performance than random search~\cite{bergstra2012random}, its computational or sample efficiency can be further improved by replacing random sampling of configurations with Bayesian optimization~\cite{falkner2018bohb}, adopting evolution strategy to promote internal knowledge transfer~\cite{ijcai2021p296}, or making it asynchronously parallel~\cite{li2020system}. 

\vspace{-0.05in}
\paragraph{Freeze-thaw BO.} Whereas the form of knowledge transfer of Hyperband (and its variants) from lower to higer fidelity is indirect, freeze-thaw BO~\cite{swersky2014freeze} transfers knowledge more directly by explicitly modeling a GP kernel to jointly model interactions between different training budgets and configurations. It then dynamically pauses (freezes) and resumes (thaws) configurations based on the last epoch performances extrapolated from a set of partially observed LCs obtained from other configurations, leading to an efficient and sensible allocation of computational resources. DyHPO~\cite{wistuba2022supervising} and its transfer version~\cite{arango2023quick} improve the computational efficiency of freeze-thaw BO~\cite{swersky2014freeze} with deep kernel GP~\cite{wilson2016deep}, but their multi-fidelity version of expected improvement (EI) acquisition extrapolates the LCs only a one-step forward, producing a greedy strategy for dynamically selecting configurations. Other recent variants of freeze-thaw BO include DPL~\cite{kadra2024scaling} and ifBO~\cite{rakotoarison2024context} which are not greedy and show competitive performances, but they are either lack of transfer learning~\cite{kadra2024scaling,rakotoarison2024context}, resort to too strong assumptions on the shape of LCs~\cite{rakotoarison2024context}, or incur the cost of retraining the surrogate function for each BO step~\cite{rakotoarison2024context}. On the other hand, we maximize the sample efficiency of freeze-thaw BO with transfer learning, while getting rid of any need for such strong assumptions or retraining costs.

\vspace{-0.05in}
\paragraph{Learning curve extrapolation.}
Freeze-thaw BO requires the ability of dynamically updating predictions on future performances from a growing set of partially observed LCs, thus heavily relies on the ability of LC extrapolation~\cite{baker2017accelerating,gargiani2019probabilistic,wistuba2020learning}. The LC extrapolation used in DyHPO~\cite{wistuba2022supervising} and Quick-Tune~\cite{arango2023quick} is based on deep kernel GP~\cite{wilson2016deep}, but extrapolates only a single step forward, making it hard to be used for our case - maximizing utilites at any possible future time steps. Freeze-thaw BO~\cite{swersky2014freeze} and DPL~\cite{kadra2024scaling} use non-greedy extrapolations but limit the shape of LCs with exponential decay kernel or power law functions, which is questionable if such strong inductive biases are applicable to more general cases. Domhan et al.~\cite{domhan2015speeding} consider a broader set of basis functions to define a prior on LCs and infer its posterior, but requires computationally expensive MCMC, and also do not consider correlations between different configurations. Klein et al.~\cite{klein2017learning} models interactions between configurations with a Bayesian neural network (BNN), but suffers from the same computational inefficiency of MCMC and also the additional cost for online retraining of the BNN. LC-PFNs~\cite{adriaensen2024efficient} are an in-context Bayesian LC extrapolation method without retraining, based on recently introduced Prior-Fitted Networks (PFNs)~\cite{muller2021transformers}. However, as with Domhan et al.~\cite{domhan2015speeding}, LC-PFNs do not consider interactions between configurations and hence is suboptimal to use as a surrogate function for freeze-thaw BO. Recently, Rakotoarison et al.~\cite{rakotoarison2024context} further combine LC-PFN with PFN4BO~\cite{muller2023pfns4bo} to develop an in-context surrogate function for freeze-thaw BO, but they train PFNs only with a prior distribution similarly to the original PFN training scheme. On the other hand, we use transfer learning, i.e., train PFNs with the existing LC datasets, to improve the sample efficiency of freeze-thaw BO while successfully encoding the correlations between configurations at the same time.

\vspace{-0.05in}
\paragraph{Transfer BO.} Transfer learning can be used for improving the sample efficiency of BO~\cite{bai2023transfer}, and here we list a few of them. Some of recent works explore scalable transfer learning with deep neural networks~\cite{perrone2018scalable,wistuba2020few}. Also, different components of BO can be transferred such as observations~\cite{swersky2013multi}, surrogate functions~\cite{golovin2017google,wistuba2020few}, hyperparmater initializations~\cite{wistuba2020few}, or all of them~\cite{wei2021meta}. However, most of the existing transfer-BO approaches assume the traditional black-box BO settings. To the best of our knowledge, Quick-Tune~\cite{arango2023quick} is the only recent work which targets multi-fidelity and transfer BO at the same time. However, as mentioned above, their multi-fidelity BO formulation is greedy, whereas our transfer-BO method can dynamically control the degree of greediness during the BO by explicitly taking into consideration the trade-off between cost and performance of BO.

\vspace{-0.05in}
\paragraph{Cost-sensitive BO.}
Multi-fidelity BO 
is inherently cost-sensitive since predictions get more accurate as the gap between the fidelities becomes closer. However, the performance metric of such vanilla multi-fidelity BO monotonically increases as we spend more budget, whereas in this paper we want to find the optimal trade-off between the amount of budget spent thus far and the corresponding intermediate performances of BO, thereby automatically early-stopping the BO around the maximal utility. Quick-Tune~\cite{arango2023quick} also suggests a cost-sensitive BO in multi-fidelity settings, but unlike our work, their primary focus is to trade-off between the performance and the cost of BO associated with pretrained models of various size, which can be seen as a generalization of more traditional notion of cost-sensitive BO~\cite{snoek2012practical,abdolshah2019cost,lee2020cost}, from black-box to multi-fidelity settings.


\vspace{-0.05in}
\section{Approach}
\vspace{-0.05in}

In this section, we introduce CMBO, a novel framework for cost-sensitive multi-fidelity BO. We first introduce notation and background on freeze-thaw BO in \S\ref{sec:background}. We then introduce the overall method and algorithm in \S\ref{sec:ustar}, and explain the transfer learning of surrogate function in \S\ref{sec:transfer_learning}.
\vspace{-0.05in}
\subsection{Notation and Background on freeze-thaw BO}
\label{sec:background}
\vspace{-0.05in}
\paragraph{Notation.} Following the convention, we assume that we are given a finite pool of hyperparameter configurations $\mathcal{X} = \{x_1,\dots,x_N\}$, with $N$ the number of configurations. Let $t \in [T] := \{1,\dots,T\}$ denote the training epochs, $T$ the last epoch, and $y_{n,1},\dots,y_{n,T}$ the validation performances (e.g., validation accuracies) obtained with the configuration $x_n$. 
We further introduce notations for multi-fidelity BO. Let $b=1,\dots,B$ denote the BO steps, $B$ the last BO step, and $\tilde{y}_1,\dots,\tilde{y}_B$ the BO performances, i.e., each $\tilde{y}_b$ is the best validation performance ($y$) obtained until the BO step $b$. 
\vspace{-0.1in}
\paragraph{Freeze-thaw BO.} Freeze-thaw BO~\cite{swersky2014freeze} is an advanced form of multi-fidelity BO. At each BO step, it allows us to dynamically select and evalute the best hyperparameter configuration $x_{n^*}$ with $n^* \in [N]$ denoting the corresponding index, while pausing the evaluation on the previous best configuration. Specifically, given $\mathcal{C}=\{(x,t,y)\}$  that represents a set of partial LCs collected up to a specific BO step, we predict for all $x \in \mathcal{X}$ the remaining part of the LCs up to the last training epoch $T$ with a (pretrained) LC extrapolator, compute the acquisition such as the expected improvement~\cite{mockus1978application} of validation performance at epoch $T$, and select and evaluate the best configuration $x_{n^*}$ that maximizes the acquisition. Note that at any BO step, the partial LCs in $\mathcal{C}$ can have different length across the configurations. Suppose that at BO step $b$ the next training epoch for $x_{n^*}$ is $t_{n^*}$. We then evaluate $x_{n^*}$ a single epoch from the corresponding checkpoint to obtain the validation performance $y_{n^*, t_{n^*}}$ at the next epoch $t_{n^*}$, which we use to update the corresponding partial LC in $\mathcal{C}$ and compute the BO performance $\tilde{y}_b$. We repeat this process $B$ times until convergence. See Alg.~\ref{algo:bo} for the pseudocode (except the red parts).
\vspace{-0.05in}
\subsection{Cost-sensitive Multi-fidelity BO}
\label{sec:ustar}
\vspace{-0.05in}
We next introduce our main method and algorithm based on freeze-thaw BO (\S\ref{sec:background}).
\vspace{-0.08in}
\paragraph{Utility function.} A utility function $U$\footnote{In this paper we assume that the utility function is predefined by a user. One can instead try to learn it from data, but we leave that as a future work.} describes the trade-off between the BO step $b$ and the BO performance $\tilde{y}_b$. Its values $U(b, \tilde{y}_b)$ negatively correlate with $b$ and positively with $\tilde{y}_b$. For instance, we may simply define $U(b, \tilde{y}_b) = \tilde{y}_b - \alpha b$ for some $\alpha > 0$, such that the utility gives linear incentives and penalties to the performance and number of BO steps, respectively.

\vspace{-0.07in}
\paragraph{Acquisition function.}
Let $t_n$ be the next training epoch for the configuration $x_n$ at a BO step $b$.
Further, suppose we have a LC extrapolator $f(\cdot|x_n,\mathcal{C})$ that can probabilistically estimate $x_n$'s remaining part of LC, $y_{n,t_n:T}$, conditioned on $\mathcal{C}$ a set of partial LCs collected up to the step $b$.
Then, based on the expected improvement (EI)~\cite{mockus1978application}, we define the acquisition function $A$ as follows:
\begin{align}
A(n) = \max_{\Delta t \in \{0,\dots, T - t_n\}} \mathbb{E}_{y_{n,t_{n}:T} \sim  f(\cdot|x_n, \mathcal{C})} \left[ \max\left(0, U(b+\Delta t, \tilde{y}_{b+\Delta t}) - U_\text{prev} \right)\right].
\label{eq:acquisition}
\end{align}
In Eq.~\eqref{eq:acquisition}, we first extrapolate $y_{n,t_n:T}$, the remaining part of the LC associated with $x_n$, and compute the corresponding predictive BO performances $\{ \tilde{y}_{b+\Delta t}\ |\ \Delta t = 0,\dots,T-t_n\}$. 
Note that according to the definition in \S\ref{sec:background}, $\tilde{y}_{b+\Delta t}$ is computed by taking the maximum among the last step BO performance $\tilde{y}_{b-1}$ as well as the newly extrapolated validation performances $y_{n,t_n},\dots,y_{n,t_n + \Delta t}$. Then, based on the increased BO step $b + \Delta t$ and the updated BO performance $\tilde{y}_{b+\Delta t}$, we compute the corresponding utility, and its expected improvement over the previous utility $U_\text{prev}$ over the distribution of LC extrapolation with the Monte-Carlo (MC) estimation. The acquisition $A(n)$ for each configuration index $n$ is defined by picking the best increment $\Delta t \in \{0,\dots,T-t_n\}$ that maximizes the expected improvement, and we eventually choose the best configuration index $n$ that maximizes $A$ (see Fig.~\ref{fig:utility}). 


\begin{wrapfigure}{t}{0.55\textwidth}
\vspace{-0.31in}
\begin{minipage}{0.55\textwidth}
\begin{algorithm}[H]
    \caption{{\color{red}Cost-sensitive} Multi-fidelity BO}
\small
\begin{algorithmic}[1]
    \State \textbf{Input:} LC extrapolator $f$, acquisition function $A$, {\color{red}utility function $U$}, maximum BO steps $B$, hyperparameter configuration pool $\mathcal{X}$, number of configurations $N$.
    \State ${\color{red}U_\text{prev}\leftarrow 0},\ \tilde{y}_0\leftarrow-\infty,\ \mathcal{C}\leftarrow\emptyset,\ t_1,\dots,t_N \leftarrow 1$
    \For{$b=1, \ldots, B$}
    \State $n^* \leftarrow \argmax_n {\color{red}A(n)}$ \Comment{Acquisition func., {\color{red}Eq.~\eqref{eq:acquisition}}}   
    \If{{\color{red}Eq.~\eqref{eq:stop} and $b>1$}} \Comment{{\color{red}Stopping criterion}}
    \State {\color{red}Break the for loop} \Comment{{\color{red}Stop the BO}}
    \EndIf
    \State Evaluate $y_{n^*,t_{n^*}}$ with $x_{n^*}$.
    \State $\mathcal{C}\leftarrow\mathcal{C} \cup \{(x_{n^*}, t_{n^*}, y_{n^*,t_{n^*}})\}$ \Comment{Update the history}
    \State $\tilde{y}_b \leftarrow \max (\tilde{y}_{b-1}, y_{n^*,t_{n^*}})$
    \Comment{Update the BO perf.}
    \State ${\color{red}U_\text{prev}\leftarrow U(b, \tilde{y}_b)}$
    \Comment{Update the prev. utility}
    \State $t_{n^*} \leftarrow t_{n^*}+1$
    \EndFor
\end{algorithmic} 
\label{algo:bo}
\end{algorithm}
\end{minipage}
\vspace{-0.15in}
\end{wrapfigure}

The main differences of our acquisition function in Eq.~\eqref{eq:acquisition} from the EI-based acquisitions used in the previous works are twofold. First, instead of maximizing the expected improvement of validation performance $y$, we maximize the expected improvement of utility. Second, rather than setting the target epoch at which we evaluate the acquisition to the last epoch $T$, we dynamically choose the optimal target epoch that is expected to maximally improve the utility. 

Those aspects allow our BO framework to more carefully select configurations for each BO step, seeking the best trade-off between cost and performance of BO. 
Specifically, the acqusition function initially prefers configurations that are expected to produce good asymptotic validation performances, but as the BO proceeds it will gradually become greedy as the performance of BO saturates and the associated cost dominates the utility function. As a result, the acquisition function will tend to exploit more than explore -- it will try to avoid selecting new configurations but stick to the few current configurations to maximize the short term performances.

Note that $U_\text{prev}$ in Eq.~\eqref{eq:acquisition}, the threshold of EI, is \emph{not} the greatest utility achieved so far, but simply set to the utility value achieved most recently (line 11 in Alg.~\ref{algo:bo}). 
This is because the cost of BO that has previously been incurred is not reversible.
It differs from the typical EI-based BO settings where we assume all the previous evaluations are freely accessible and set the threshold to the maximum among them.  
As a result, $U_\text{prev}$ can either increase or decrease during the BO, and we need to stop the BO when $U_\text{prev}$ starts decreasing monotonically, i.e., when the performance of BO stops improving meaningfully with respect to the associated cost.

\vspace{-0.07in}
\begin{wrapfigure}{R}{0.33\textwidth}
\vspace{-0.15in}
\centering
    \includegraphics[width=0.33\textwidth]{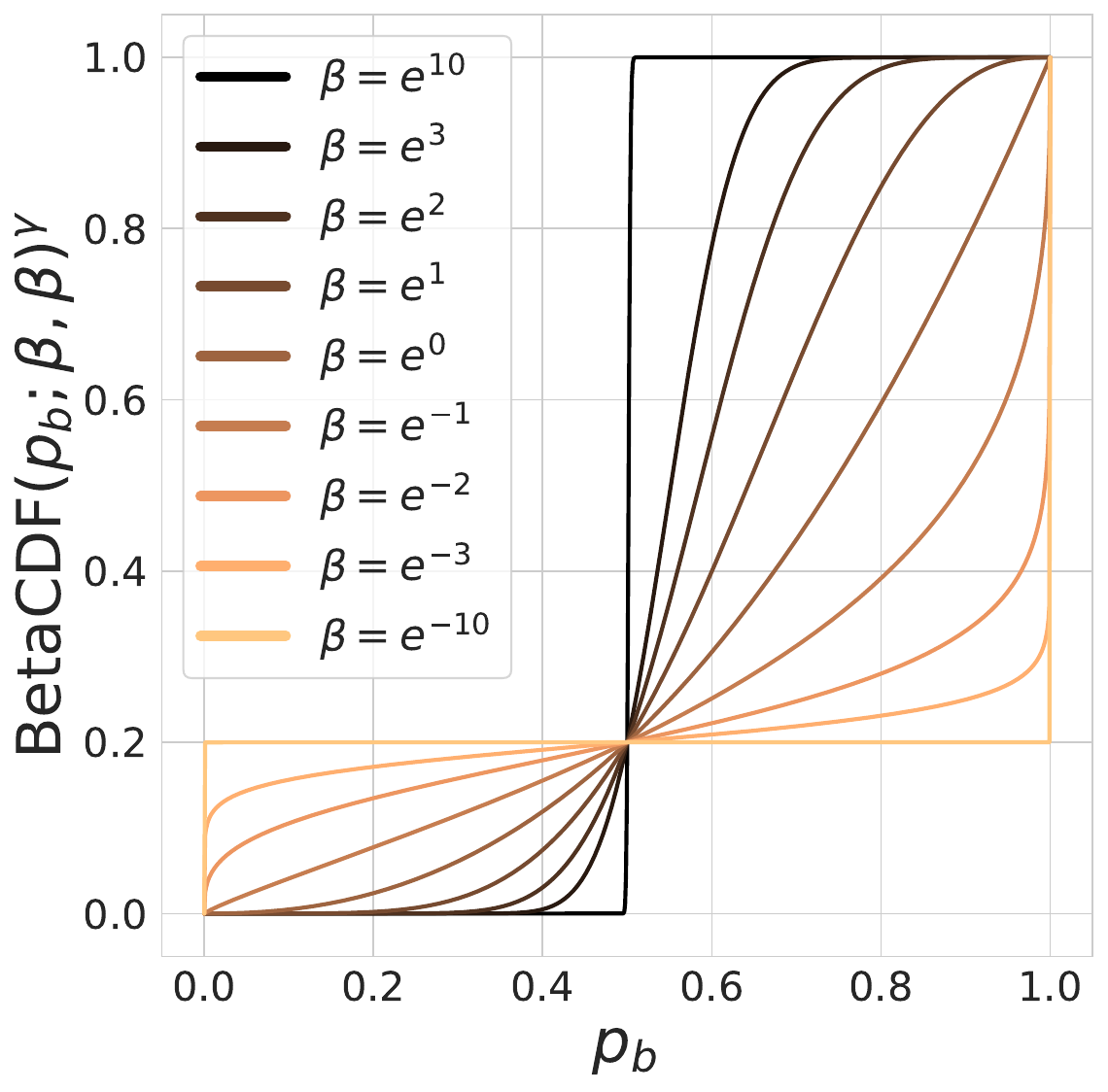}
    \vspace{-0.25in}
\caption{\small Eq.~\eqref{eq:threshold} with $\gamma=\log_2 5$ and the various values for $\beta$.} 
\label{fig:beta}
\vspace{-0.3in}
\end{wrapfigure}
\paragraph{Stopping criterion.}
The next question is how to properly stop the BO around the maximum utility.
We propose to stop when the following criterion is satisfied at each BO step $b > 1$:
\begin{align}
    \frac{\hat{U}_\text{max} - U_\text{prev}}{\hat{U}_\text{max} - \hat{U}_\text{min}} > \delta_b.
    \label{eq:stop}
\end{align}
In Eq.~\eqref{eq:stop}, $U_\text{prev}$ is the utility value at the last step $b-1$, $\hat{U}_\text{max}$ is defined as the maximum utility value seen up to the last step, and $\hat{U}_\text{min} = U(B, \tilde{y}_1)$. The role of $\hat{U}_\text{max}$ and $\hat{U}_\text{min}$ is to roughly estimate the maximum and the minimum utility achievable over the course of BO, respectively. Therefore, the LHS of Eq.~\eqref{eq:stop} can be seen as the \emph{normalized regret} of utility roughly estimated at the current step $b$, and we stop the BO as soon as the current estimation on the regret exceeds some threshold $\delta_b$.

To define $\delta_b$, let $n^* = \argmax_n A(n)$ denote the index of the currently chosen best configuration $x_{n^*}$ based on Eq.~\eqref{eq:acquisition}, $\operatorname{BetaCDF}$ the cumulative distribution function (CDF) of $\operatorname{Beta}$, and $\mathbbm{1}$ the indicator function. Then, we have:
\begin{align}
    \delta_b &= \operatorname{BetaCDF}(p_b; \beta, \beta)^{\gamma},
    \quad \beta, \gamma > 0,
    \label{eq:threshold}
    \\
    p_b &= \max_{\Delta t \in\{1, \ldots, T-t_{n^*}\}} \mathbb{E}_{y_{n,t_{n^*}:T}\sim f(\cdot|x_{n^*}, \mathcal{C})} \left[ \mathbbm{1}\left(U\left(b+\Delta t, \tilde{y}_{b + \Delta t}\right) > U_\text{prev}\right) \right].
    \label{eq:pi}
\end{align}
$p_b$ in Eq.~\eqref{eq:pi} is the probability that the current best configuration $x_{n^*}$ improves on $U_\text{prev}$ in some future BO step (i.e., probability of improvement, or PI~\cite{mockus1978application}). Intuitively, we want to defer the termination as $p_b$ increases, and vice versa. It is considered in Eq.~\eqref{eq:threshold} -- as $p_b$ increases, the threshold $\delta_b$ increases as well because $\operatorname{BetaCDF}(\cdot;\beta,\beta)^\gamma$ is a monotonically increasing function in $[0, 1]$, so we have less motivation to stop according to Eq.~\eqref{eq:stop}. 

Fig.~\ref{fig:beta} plots $\operatorname{BetaCDF}(\cdot;\beta,\beta)^\gamma$ in Eq.~\eqref{eq:threshold} over the various values of $\beta$ and when $\gamma = \log_2 5$. We see that the function becomes vertical as $\beta \rightarrow +\infty$ and horizontal as $\beta \rightarrow 0$.  In the former case, we terminate the BO when $p_b < 0.5$ while ignoring the regret in the LHS of Eq.~\eqref{eq:stop}, whereas in the latter case we ignore $p_b$ and only decide based on the regret, with the threshold $\delta_b$ fixed to a specific value specified by $\gamma$ (e.g., $\gamma=\log_2 5$ corresponds to $\delta_b=0.2$ in Fig.~\ref{fig:beta}). Therefore, the role of $\beta$ is to smoothly interpolate between the two extreme stopping criteria, whereas $\gamma$ decides the range of the overall shape of the interpolated criterion.

\vspace{-0.07in}
\paragraph{Algorithm.}
We summarize the pseudocode of our overall method in Alg.~\ref{algo:bo}, with the red parts corresponding to the specifics of our method.

\vspace{-0.05in}
\subsection{Transfer Learning of LC 
Extrapolation}
\label{sec:transfer_learning}
\vspace{-0.05in}
As explained above, our BO framework is based on freeze-thaw BO (\S\ref{sec:background})
which heavily resorts to accurate LC extrapolation. Since we assume that users may want to early-stop the BO, we should have an even more sample efficient LC extrapolation mechanism for preventing inaccurate early-stopping before collecting a sufficient amount of BO observations. Therefore, we propose to make use of transfer learning to maximally improve the sample efficiency of the LC extrapolator.

\vspace{-0.07in}
\paragraph{Transfer learning with LC mixup.} Among many plausible options, in this paper we propose to use Prior Fitted Networks (PFNs)~\cite{muller2021transformers} for LC extrapolation. PFNs are an in-context Bayesian inference method based on Transformer architectures~\cite{vaswani2017attention}, and show good performances on LC extrapolation~\cite{adriaensen2024efficient,rakotoarison2024context} without the computationally expensive online retraining~\cite{kadra2024scaling}. A major difficulty of using PFNs for our purpose is that they are not trained with the existing datasets, but trained with a prior distribution, which is essential to perform Bayesian inference. Also, PFNs require relatively a large Transformer architecture~\cite{vaswani2017attention} as well as huge amounts of training examples for good generalization performance~\cite{adriaensen2024efficient}, which makes it risky to train PFNs with a finite set of examples. 

Here we explain our novel transfer learning method for PFNs that can circumvent those difficulties with the mixup strategy~\cite{zhang2018mixup}. Suppose we have $M$ different datasets and the corresponding $M$ sets of LCs collected from $N$ hyperparameter configurations. Define $l_{m,n}=(y^{m}_{n,1},\dots,y^{m}_{n,T})$, the $T$-dimensional row vector of validation performances ($y$'s) collected from the $m$-th dataset and the $n$-th configuration, forming a complete LC of length $T$. Further define the matrix $L_m = [l_{m,1}^\top ; \dots ; l_{m,N}^\top]^\top$, the row-wise stack of those LCs. In order to augment training examples, we propose to perform the following two consecutive mixups~\cite{zhang2018mixup}:
\vspace{-0.05in}
\begin{enumerate}[itemsep=1mm, parsep=0pt, leftmargin=*]
\item Across datasets: $L' = \lambda_1 L_m + (1-\lambda_1) L_{m'}$,\quad with $\lambda_1 \sim \operatorname{Unif}(0,1),\quad\text{for all } m, m' \in [M]$.
\item Across configurations: $(x'', l'') = \lambda_2 (x_n, l'_{n}) + (1-\lambda_2) (x_{n'}, l'_{n'})$ 
\vspace{2pt}
\newline
{\color{white}.} $\qquad\qquad\qquad\qquad$ with $l'_n \text{ the } n\text{-th row of } L' ,\ \  \lambda_2 \sim \operatorname{Unif}(0,1),\quad \text{for all } L' \text{ and } n,n' \in [N]$.
\end{enumerate}
\vspace{-0.05in}
In this way, we can sample infinitely many training examples $\{(x'',l'')\}$ by interpolating between the LCs, leading to a robust LC extrapolator with less overfitting.
Note that in the first step, we do not individually perform the mixup over the configurations but apply the same $\lambda_1$ to all the configurations, in order to preserve the correlations between the configurations encoded in the given datasets.

As for the network architecture and the training objective, we mostly follow Rakotoarison et al.~\cite{rakotoarison2024context}. We use a similar Transformer architecture that takes a set of partial LCs and the corresponding configurations as an input and extrapolates the remaining part of the LCs. The training objective then maximizes the likelihood of those predictions conditioned on the partial LCs. We defer more details on the training to \S\ref{sec:pretraining_details}. Also, see \S\ref{sec:mixup_related_work} for more discussion about the connection of our transfer learning method with ifBO~\cite{rakotoarison2024context} and Transformer Neural Processes (TNPs)~\cite{nguyen2022transformer}. 

\vspace{-0.05in}
\section{Experiments}
\vspace{-0.05in}
\label{sec:experiment}
We next validate the efficacy of our method on various multi-fidelity HPO settings. We will publicly release our code upon acceptance.

\begin{figure}[t]
\vspace{-0.3in}
\centering
\includegraphics[height=3.3cm]{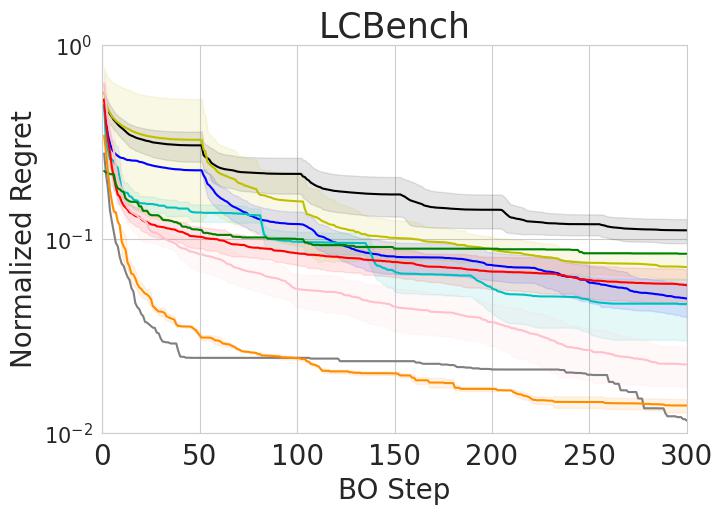}
\includegraphics[height=3.3cm]{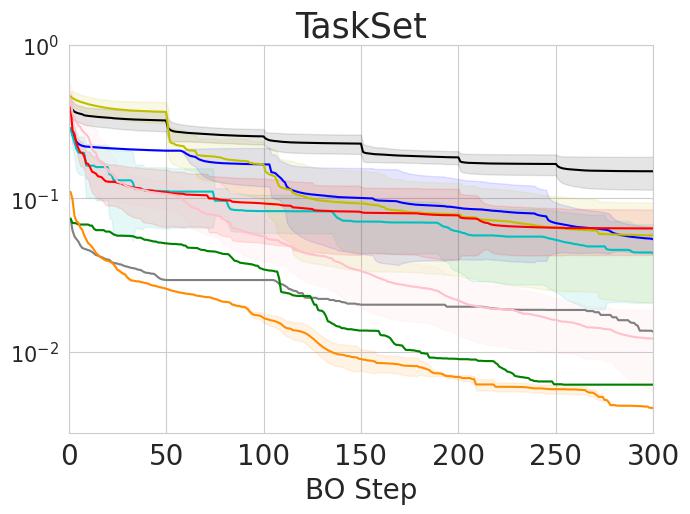}
\includegraphics[height=3.3cm]{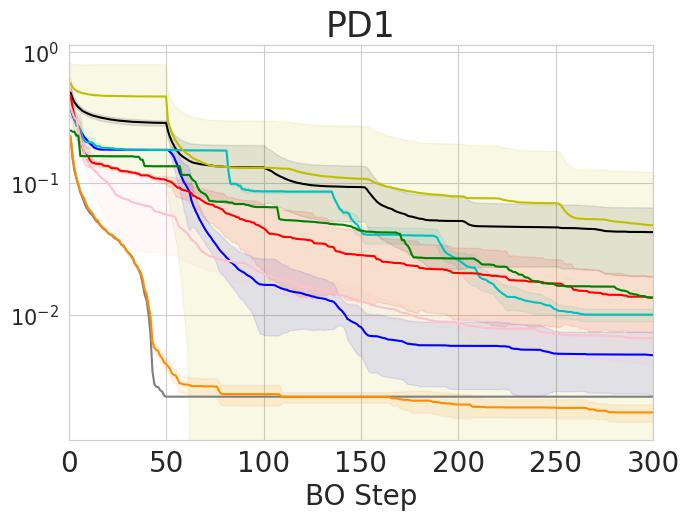}
\vspace{-0.15in}
\medskip
\includegraphics[width=1.0\textwidth]{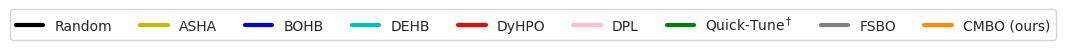}
\caption{\small \textbf{The results on the conventional multi-fidelity HPO setup} ($\alpha=0$). For each benchmark, we report the normalized regret of utility aggregated over all the test datasets.}
\vspace{-0.15in}
\label{fig:alpha_0}
\end{figure}

\vspace{-0.07in}
\paragraph{Datasets.} We use the following benchmark datasets for multi-fidelity HPO. \textbf{LCBench} \cite{zimmer2021auto}: A LC dataset that evaluates the performance of 7 different hyperparameters on 35 different tabular datasets. The LCs are collected by training MLPs with 2,000 hyperparameter configurations, each for 51 epochs. We train our LC prediction model on 20 datasets and evaluate on the remaining 15 datasets. \textbf{TaskSet} \cite{metz2020using}: A LC dataset that consists of a diverse set of 1,000 optimization tasks drawn from various domains. We select 30 natural language processing (text classification and language modeling) tasks, train our LC extrapolator on 21 tasks, and evaluate on the remaining 9 tasks. 
Each task include 8 different hyperparameters and 1,000 their configurations. Each LC is collected by training models for 50 epochs. \textbf{PD1} \cite{wang2021pre}: A LC benchmark that includes the performance of modern neural architectures (including Transformers) run on large vision datasets such as CIFAR-10, CIFAR-100~\cite{krizhevsky2009learning}, ImageNet~\cite{ILSVRC15}, as well as statistical modeling corpora and protein sequence datasets from bioinformatics. We select 23 tasks with 4 different hyperparameters based on SyneTune \cite{salinas2022syne} package, train our LC extrapolator on 16 tasks, and evaluate on the remaining 7 tasks. For easier transfer learning, we preprocess the data by excluding hyperparameter configurations with their training diverging (e.g., LCs with \texttt{NaN}), and linearly interpolate the LCs to match their length across different tasks. We then obtain the LCs of 50 epochs over the 240 configurations. See \S\ref{sec:data} for more details.

\vspace{-0.07in}
\paragraph{Baselines.} We compare our method against \textbf{Random Search}~\cite{bergstra2012random} that randomly selects hyperparameter configurations sequentially. We next compare against several variants of Hyperband~\cite{li2018hyperband} such as 
\textbf{ASHA}~\cite{li2020system} the asynchronous parallel version of it, \textbf{BOHB}~\cite{falkner2018bohb} which replaces its random sampling of configurations with BO, and  \textbf{DEHB}~\cite{ijcai2021p296} which promotes internal knowledge transfer with evolution strategy. We also compare against more recent multi-fidelity BO methods such as \textbf{DyHPO}~\cite{wistuba2022supervising} which uses deep kernel GP~\cite{wilson2016deep} and a greedy acquisition function with a short-horizon LC extrapolation, and \textbf{DPL}~\cite{kadra2024scaling} which extrapolates LCs with power law functions and model ensemble. \textbf{Quick-Tune}$\boldsymbol{^\dagger}$, is a modified version of Quick-Tune~\cite{arango2023quick} which is originally developed for dynamically selecting both pretrained models and hyperparmater configurations, with the additional cost term penalizing the non-uniform evaluation wall-time associated with each joint configuration. Since our experimental setup does not consider selecting pretrained models nor non-uniform evaluation wall-time, we only leave the transfer learning part of the model, which corresponds to a transfer learning version of DyHPO, i.e., we train its surrogate function with the same LC datasets used for training our LC extrapolator. Lastly, we compare against \textbf{FSBO}~\cite{wistuba2020few}, a black-box transfer-BO method that uses the same LC datasets to train a deep kernel GP surrogate function. The difference of FSBO from Quick-Tune$^\dagger$ is that its surrogate models the validation performances at the last epoch, whereas the surrogate of Quick-Tune$^\dagger$ predicts the performances at the next epoch for multi-fidelity HPO. See \S\ref{sec:additional_setups} for more details.

\vspace{-0.07in}
\paragraph{Utility function.} While there are many plausible options for the utility function, in this paper we use a linear function for penalizing the cost of multi-fidelity BO, i.e., $U(b,\tilde{y}) = \tilde{y} - \alpha b$ where $\tilde{y}$ is the BO performance, $b$ the BO steps, and $\alpha \in \{0, 4e\text{-}05, 2e\text{-}04\}$. Note that $\alpha=0$ does not penalize the number of BO steps at all, hence the BO does not terminate until the last BO step $B$ as with the conventional multi-fidelity BO setup.

\begin{figure}[t]
\vspace{-0.1in}
\begin{subfigure}[c]{0.245\textwidth}
\centering
    \includegraphics[width=1.\textwidth]{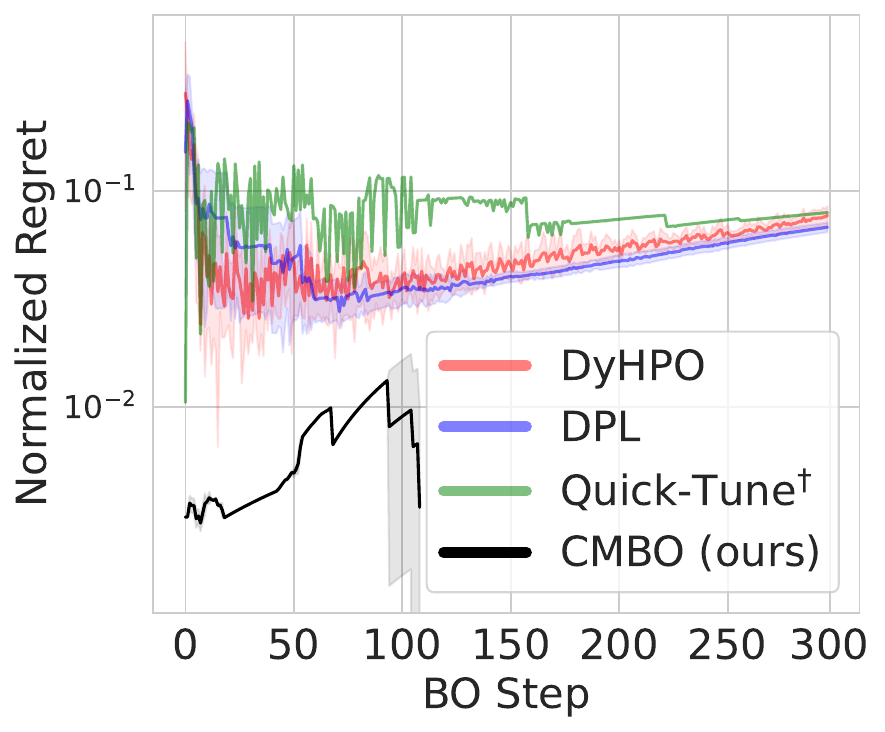}
    \vspace{-0.25in}
\subcaption{}\label{fig:acq1}
\end{subfigure}
\begin{subfigure}[c]{0.245\textwidth}
\centering
    \includegraphics[width=1.\textwidth]{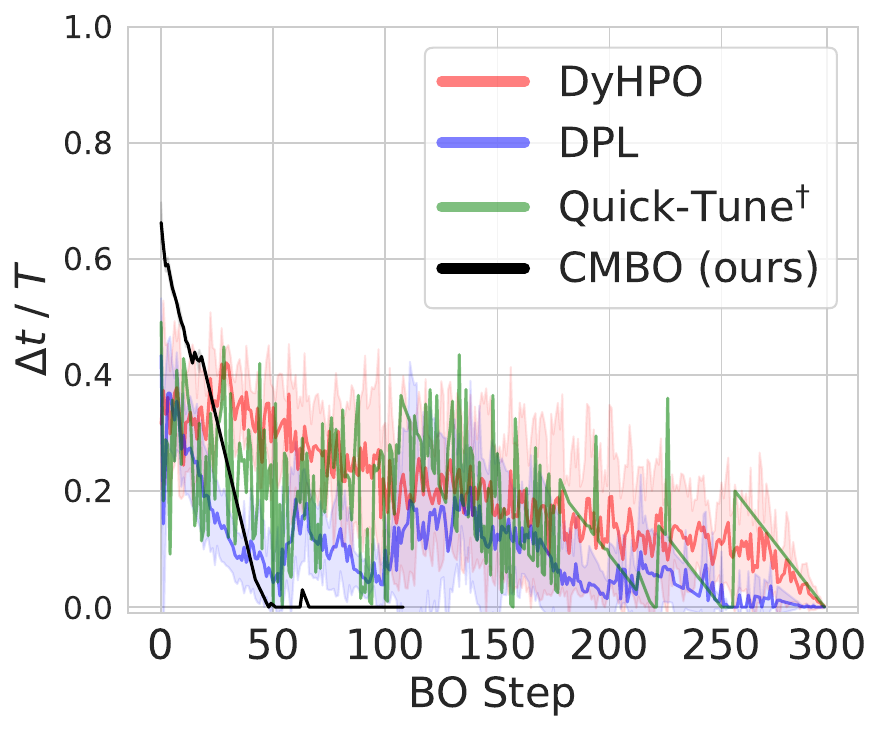}
    \vspace{-0.25in}
\subcaption{}
    \label{fig:acq2}
\end{subfigure}
\begin{subfigure}[c]{0.245\textwidth}
\centering
    \includegraphics[width=1.\textwidth]{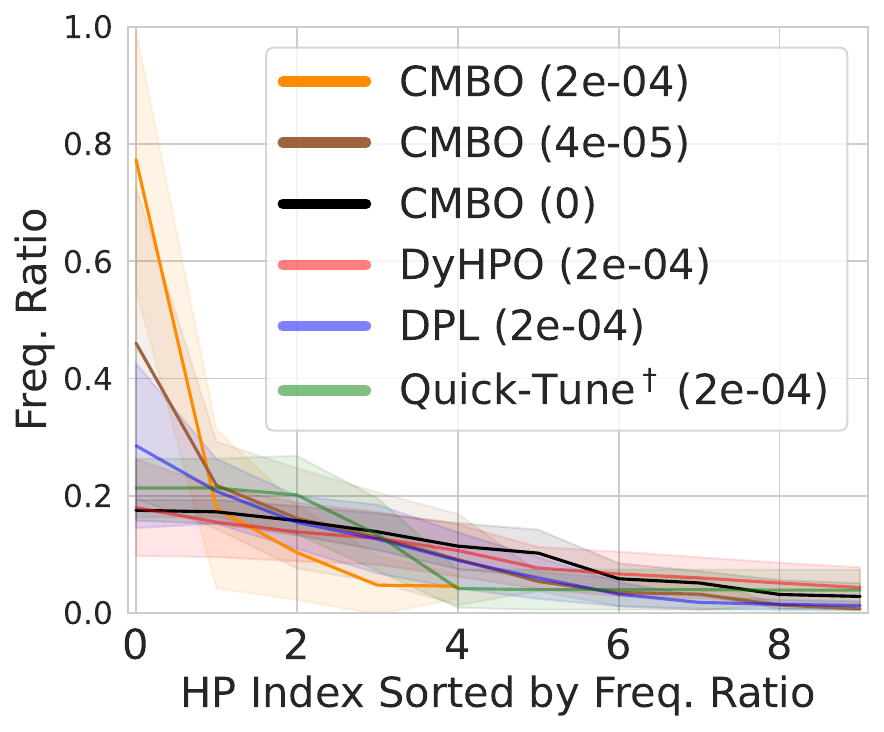}
    \vspace{-0.25in}
\subcaption{}
    \label{fig:acq3}
\end{subfigure}
\begin{subfigure}[c]{0.245\textwidth}
\centering
    \includegraphics[width=1.\textwidth]{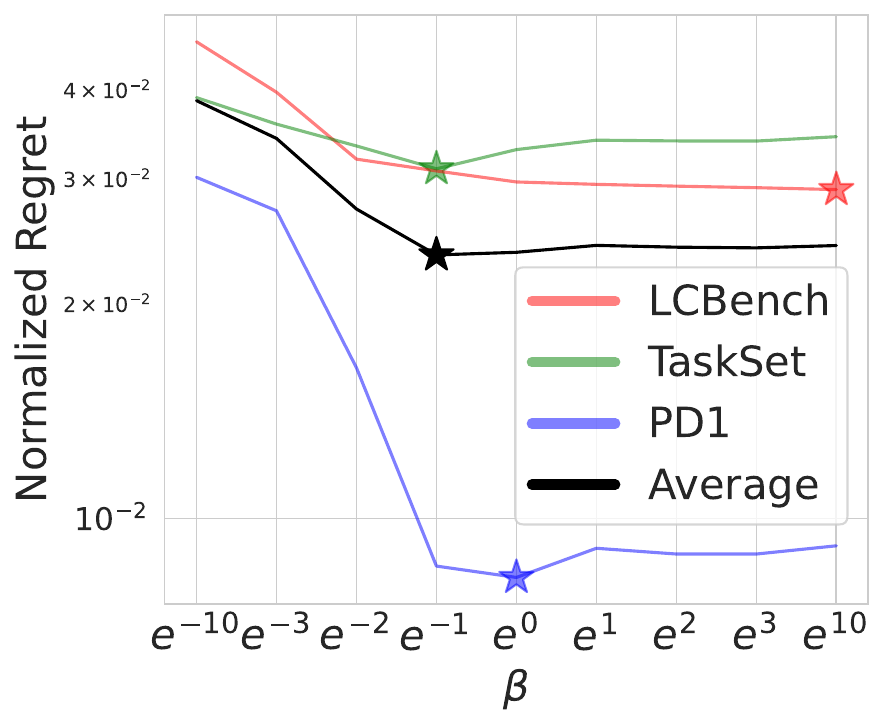}
    \vspace{-0.25in}
\subcaption{}
    \label{fig:stop_ablation}
\label{fig:beta_ablation}
\end{subfigure}
\vspace{-0.1in}
\caption{\small \textbf{(a, b, c)} Additional analysis on the effectiveness of our acquisition function. We use PD1 for the visualization. In \textbf{(c)}, the values of $\alpha$ are shown in the parenthesis. \textbf{(d)} Ablation study on $\beta$, with the minimum regret shown with the asterisks. }
\vspace{-0.1in}
\end{figure}

\vspace{-0.07in}
\paragraph{Stopping criterion.} 
For the baselines, we simply use the fixed threshold $\delta_b = 0.2$ in Eq.~\eqref{eq:stop} as computing the PI in Eq.~\eqref{eq:pi} is not straightforward for them. For our model, we use $\beta = \exp(3)$ and $\gamma = \log_2 5$ 
for all the experiments in this paper, except the ablation study in Fig.~\ref{fig:beta_ablation}.


\vspace{-0.07in}
\paragraph{Evaluation metric.} 
In order to report the average performances over the tasks, we use the normalized regret of utility $(U_\text{max}-U_{b^*})/(U_\text{max} - U_\text{min}) \in [0, 1]$, similarly to Eq.~\eqref{eq:stop}.
$U_{b^*}$ is the utility obtained right after the BO terminates at step $b^*$, and $U_\text{max}$ is the maximum achievable by running a single optimal configuration up to its maximum utility. Computing the exact $U_\text{min}$ is a difficult combinatorial optimization problem, thus we simply approximate it with $U(B, y^{\text{worst}}_{1})$, where $y^{\text{worst}}_{1}$ is the worst 1-epoch validation performance across the configurations -- we simply let $y_{1}^{\text{worst}}$ decay over the maximum BO steps $B$, corresponding to a lower bound of the exact $U_\text{min}$. We then average the normalized regret across all the tasks in each benchmark, and report the mean and standard deviation over 5 runs. Lastly, we also report the rank of each method averaged over the tasks.


\definecolor{Gray}{gray}{0.9}

\begin{table}[t]
\vspace{-0.1in}
\centering
\caption{\small \textbf{Results on the cost-sensitive multi-fidelity HPO setups} ($\alpha=4e\text{-}05, 2e\text{-}04$). For better readability, we multiply 100 to the normalized regret. The transfer learning methods are indicated by blue.}
\label{tab:main}
\resizebox{1.0\textwidth}{!}{%
\begin{tabular}{ccccccccccccc}
\midrule[0.8pt]
\multirow{3}{*}{Method} & \multicolumn{4}{c}{LCBench}                           & \multicolumn{4}{c}{TaskSet}                           & \multicolumn{4}{c}{PD1} \\ \cline{2-13}
                        & \multicolumn{2}{c}{$\alpha=4e\text{-}05$} & \multicolumn{2}{c}{$\alpha=2e\text{-}04$} & \multicolumn{2}{c}{$\alpha=4e\text{-}05$} & \multicolumn{2}{c}{$\alpha=2e\text{-}04$} & \multicolumn{2}{c}{$\alpha=4e\text{-}05$} & \multicolumn{2}{c}{$\alpha=2e\text{-}04$} \\ \cline{2-13}
                        & Regret & Rank & Regret & Rank & Regret & Rank & Regret & Rank & Regret & Rank & Regret & Rank \\
\midrule[0.8pt]

\rowcolor{Gray}
Random~\cite{bergstra2012random} & 14.1\tiny$\pm$1.8 & 7.7 & 18.1\tiny$\pm$1.7 & 7.5 & 18.9\tiny$\pm$4.6 & 7.8 & 22.3\tiny$\pm$4.3 & 7.7 & \phantom{0}5.4\tiny$\pm$2.3 & 7.1 & 11.0\tiny$\pm$5.6 & 7.2 \\

ASHA~\cite{li2020system} & \phantom{0}8.6\tiny$\pm$1.0 & 6.3 & 13.7\tiny$\pm$1.1 & 6.8 & \phantom{0}8.4\tiny$\pm$4.1 & 6.2 & 14.7\tiny$\pm$5.4 & 7.0 & \phantom{0}5.9\tiny$\pm$7.3 & 6.5 & \phantom{0}9.7\tiny$\pm$7.4 & 6.5 \\

\rowcolor{Gray}
BOHB~\cite{falkner2018bohb} & \phantom{0}6.4\tiny$\pm$1.0 & 5.0 & 11.6\tiny$\pm$1.0 & 5.5 & \phantom{0}7.4\tiny$\pm$1.4 & 6.9 & 11.2\tiny$\pm$1.9 & 6.2 & \phantom{0}1.6\tiny$\pm$0.2 & 4.7 & \phantom{0}4.9\tiny$\pm$0.2 & 4.9 \\

DEHB~\cite{ijcai2021p296} & \phantom{0}6.1\tiny$\pm$1.6 & 4.6 & 11.0\tiny$\pm$1.4 & 4.9 & \phantom{0}5.8\tiny$\pm$2.3 & 6.1 & 10.0\tiny$\pm$1.7 & 6.0 & \phantom{0}2.1\tiny$\pm$0.1 & 6.1 & \phantom{0}5.4\tiny$\pm$0.1 & 6.0 \\

\rowcolor{Gray}
DyHPO~\cite{wistuba2022supervising} & \phantom{0}7.2\tiny$\pm$1.2 & 5.7 & 12.1\tiny$\pm$1.6 & 5.9 & \phantom{0}7.5\tiny$\pm$2.1 & 6.4 & 11.1\tiny$\pm$2.0 & 6.3 & \phantom{0}2.5\tiny$\pm$0.6 & 6.2 & \phantom{0}6.2\tiny$\pm$0.9 & 6.7 \\

DPL~\cite{kadra2024scaling} & \phantom{0}3.8\tiny$\pm$0.5 & 3.2 & \phantom{0}9.3\tiny$\pm$0.5 & 4.4 & \phantom{0}2.6\tiny$\pm$0.7 & 3.4 & \phantom{0}7.5\tiny$\pm$0.6 & 4.5 & \phantom{0}1.8\tiny$\pm$0.3 & 4.5 & \phantom{0}5.1\tiny$\pm$0.6 & 4.7 \\

\rowcolor{Gray}
{\color{blue} Quick-Tune}$\boldsymbol{^\dagger}$~\cite{arango2023quick} & \phantom{0}9.6\tiny$\pm$0.0 & 6.9 & 12.7\tiny$\pm$0.0 & 6.1 & \phantom{0}3.7\tiny$\pm$0.0 & 3.7 & \phantom{0}5.6\tiny$\pm$0.0 & 3.1 & \phantom{0}2.4\tiny$\pm$0.0 & 5.4 & \phantom{0}5.5\tiny$\pm$0.0 & 5.0 \\ 

{\color{blue}FSBO}~\cite{wistuba2020few} & \phantom{0}2.6\tiny$\pm$0.0 & 3.0 & \phantom{0}6.4\tiny$\pm$0.0 & 2.7 & \phantom{0}2.9\tiny$\pm$0.0 & 3.0 & \phantom{0}4.9\tiny$\pm$0.0 & 2.6 & \phantom{0}1.3\tiny$\pm$0.0 & 2.6 & \phantom{0}4.2\tiny$\pm$0.0 & 3.1 \\ 
\midrule[0.8pt]

\rowcolor{Gray}
{\color{blue}\bf CMBO (ours)} & \bf\phantom{0}2.3\tiny$\pm$0.1 & \bf2.7 & \bf\phantom{0}3.1\tiny$\pm$0.0 & \bf1.3 & \bf\phantom{0}1.3\tiny$\pm$0.0 & \bf1.5 & \bf\phantom{0}3.1\tiny$\pm$1.0 & \bf1.5 & \bf\phantom{0}0.8\tiny$\pm$0.0 & \bf1.9 & \bf\phantom{0}0.9\tiny$\pm$0.0 & \bf1.0 \\
\midrule[0.8pt]
\end{tabular}
}
\end{table}

\begin{figure}[t]
\vspace{-0.15in}
\centering
\includegraphics[height=3.3cm]{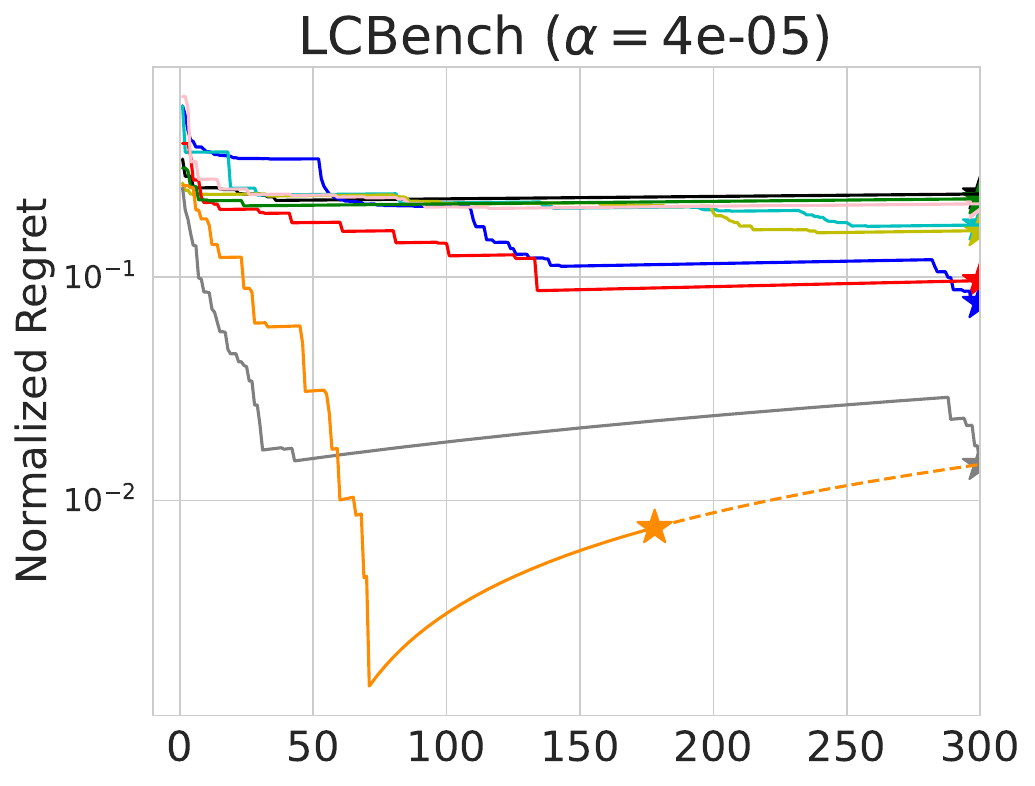}
\includegraphics[height=3.3cm]{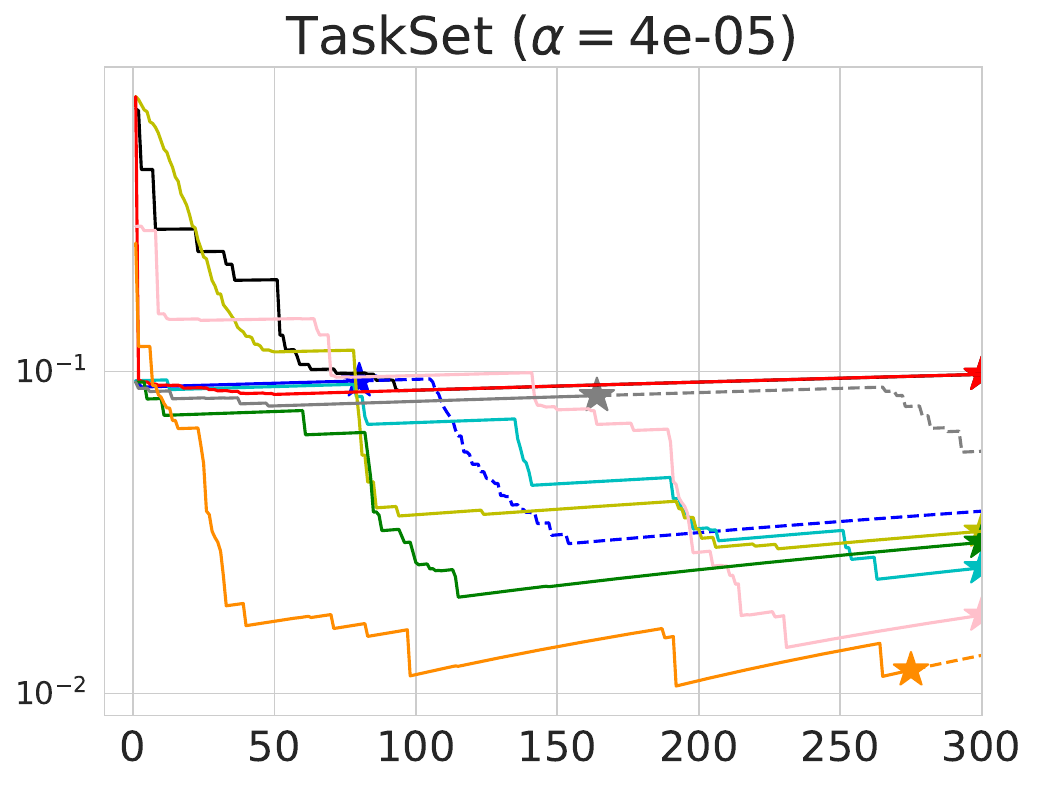}
\includegraphics[height=3.3cm]{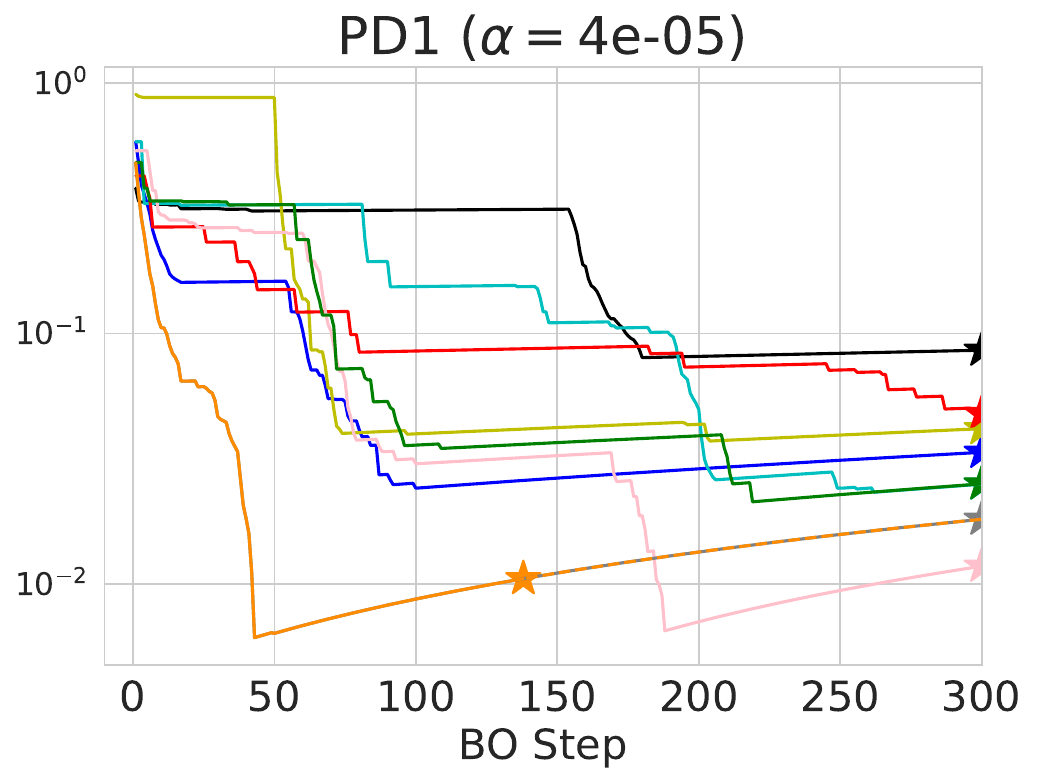}
\includegraphics[height=3.3cm]{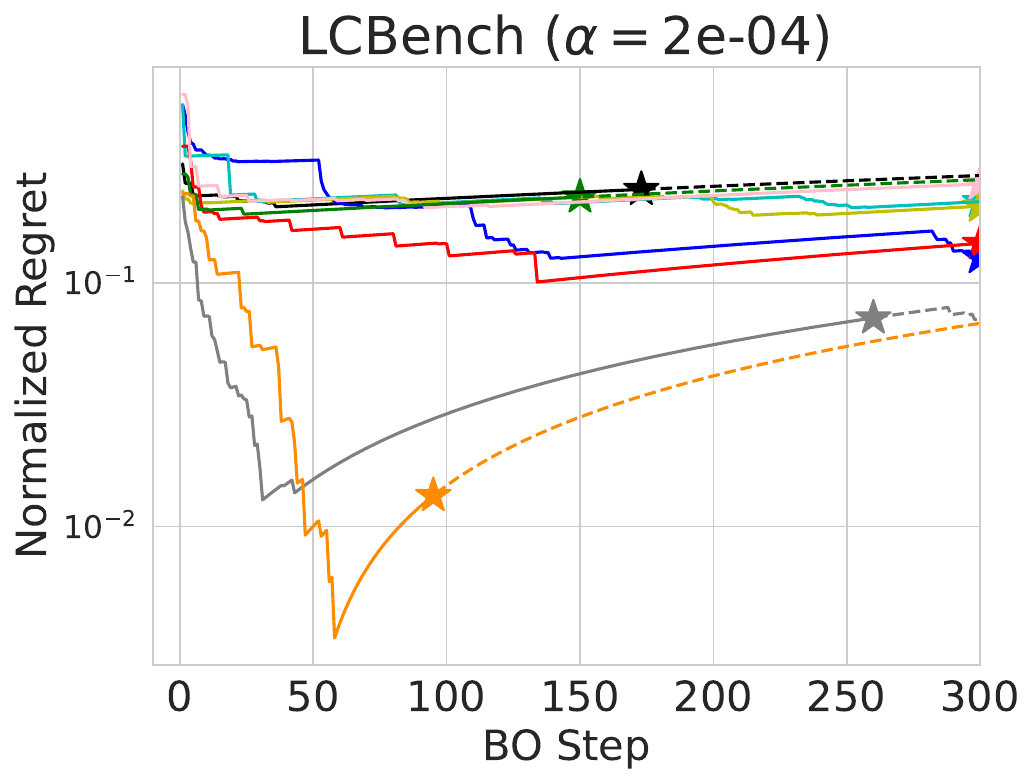}
\includegraphics[height=3.3cm]{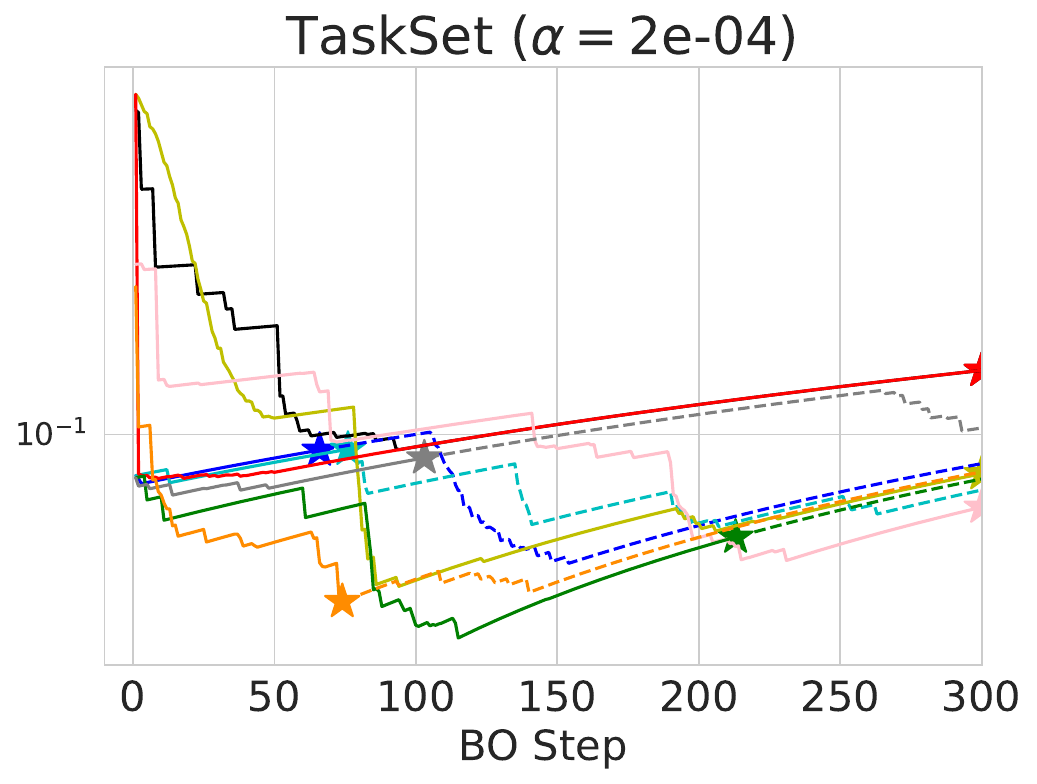}
\includegraphics[height=3.3cm]{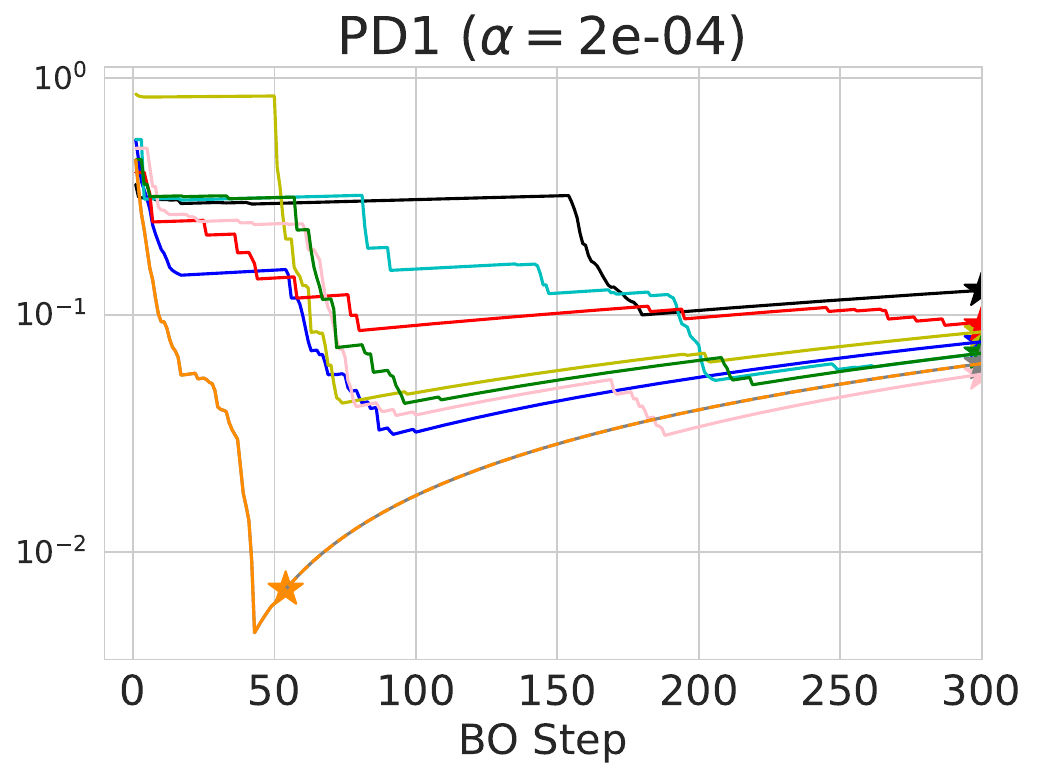}
\vspace{-0.17in}
\medskip
\includegraphics[width=1.0\textwidth]{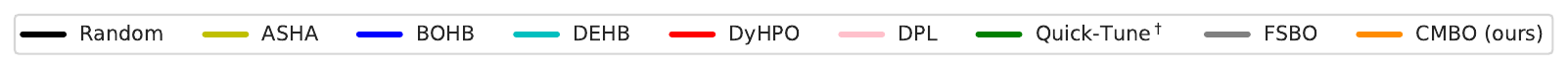}

\caption{\small \textbf{Visualization of the normalized regret over BO steps}. The first and second row correspond to $\alpha=4e\text{-}05$ and $2e\text{-}04$, respectively. Each column shows the results on a cherry-picked task from each benchmark. The asterisks indicate the stopping points, and the dotted lines represent the normalized regret achievable by running each method without stopping. See \S\ref{sec:additional_results} for the results on all the other tasks.}
\label{fig:main}
\vspace{-0.15in}
\label{graph_results}
\end{figure}

\vspace{-0.07in}
\subsection{Analysis}
\vspace{-0.05in}

\begin{wrapfigure}{R}{0.5\textwidth}
\vspace{-0.15in}
\begin{subfigure}[c]{0.247\textwidth}
\centering
    \includegraphics[height=3.5cm]{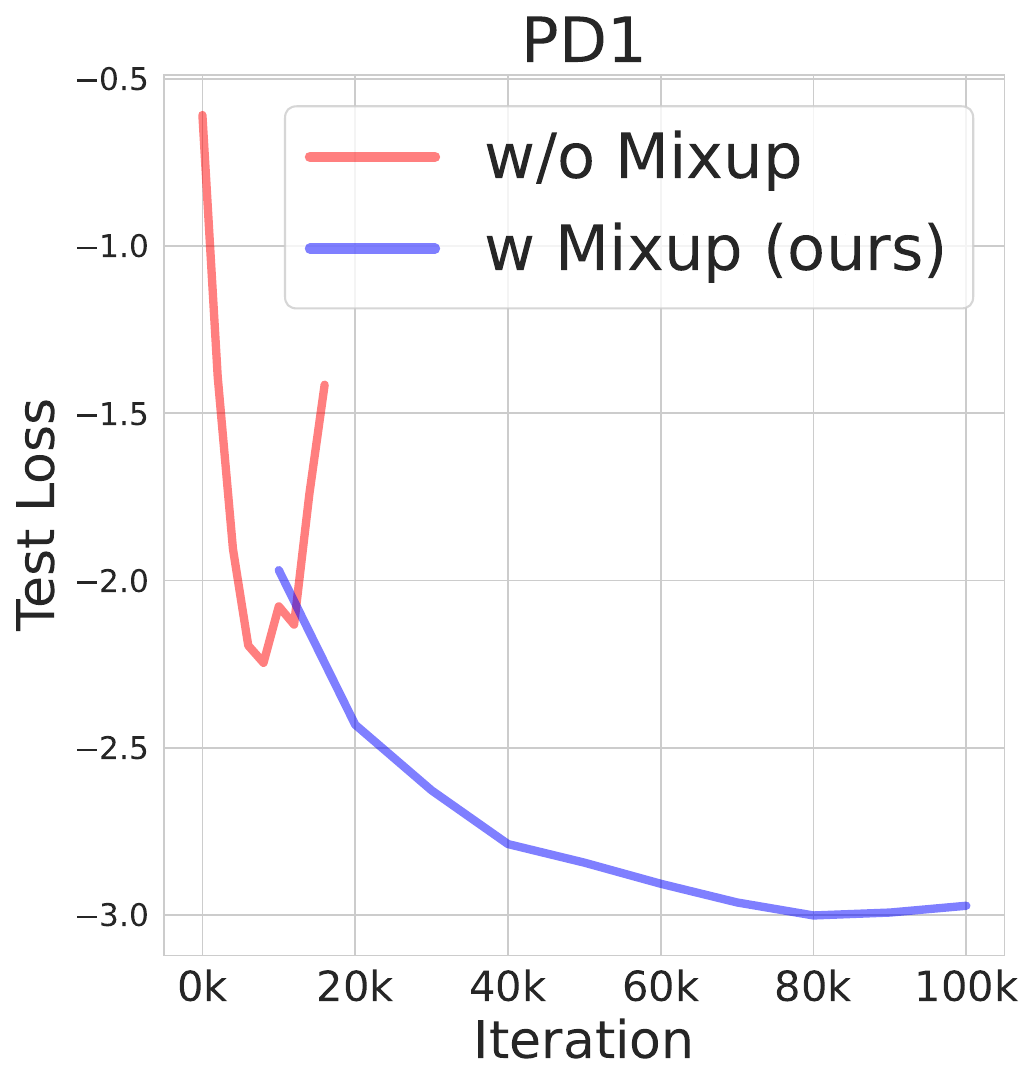}
    \vspace{-0.05in}
\subcaption{Test loss}
\label{fig:overfig_pd1_main}
\end{subfigure}
\begin{subfigure}[c]{0.247\textwidth}
\centering
    \includegraphics[height=3.5cm]{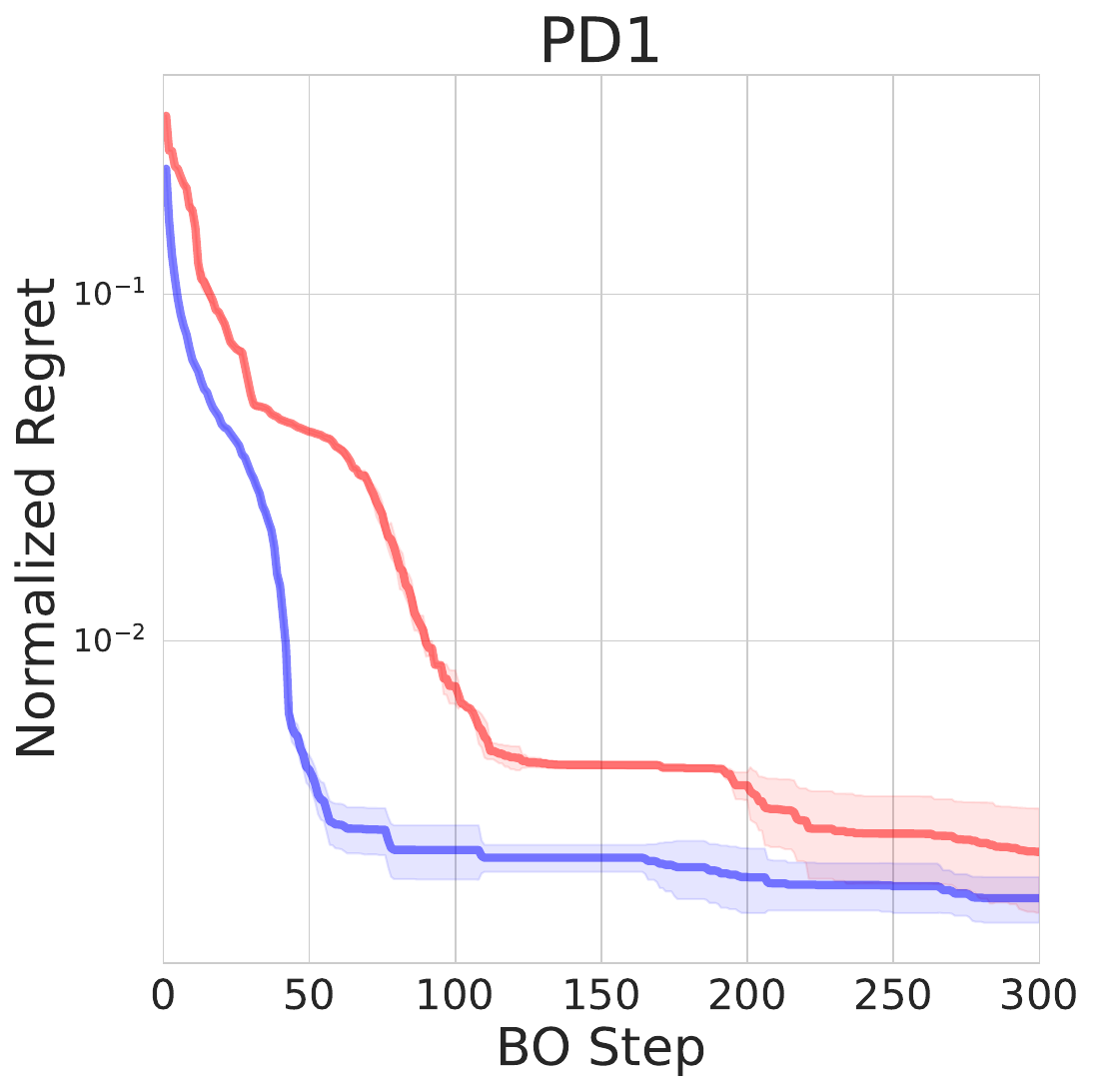}
    \vspace{-0.2in}
\subcaption{Normalized Regret}
\label{fig:overfig_pd1_hpo_main}
\end{subfigure}
\vspace{-0.07in}
\caption{\small Ablation study on the mixup training. We use $\alpha=0$ and PD1 benchmark for the experiments.}
\label{fig:mixup_ablation}
\vspace{-0.1in}
\end{wrapfigure}
\paragraph{Effectiveness of our transfer learning.} We first demonstrate the effectiveness of our transfer learning method. Fig.~\ref{fig:alpha_0} shows the results on the conventional multi-fidelity HPO setting where we do not penalize the cost of BO at all ($\alpha=0$). First of all, note that FSBO, a black-box transfer-BO method which switches its configuration only after a single complete training (e.g., 50 epochs), even outperforms all the other multi-fidelity methods that can change the configurations every epoch. The result clearly shows the importance of transfer learning for improving the sample efficiency of HPO. Quick-Tune$^\dagger$, a transfer version of DyHPO, performs similarly to the other baselines despite of the transfer learning, except on TaskSet benchmark. We attribute this result to its greedy acquisition function, and more importantly its lack of data augmentation. On the other hand, our method is non-greedy (when $\alpha = 0$) and can effectively augment the data with our mixup strategy, thereby showing significantly better performances than all the other multi-fidelity methods. Fig.~\ref{fig:mixup_ablation} shows the ablation study on our mixup training. Fig.~\ref{fig:overfig_pd1_main} shows that we can effectively reduce the risk of overfitting by adding the mixup strategy. As a result, the performance of BO improves significantly (Fig.~\ref{fig:overfig_pd1_hpo_main}). Lastly, our method significantly outperforms FSBO on TaskSet and slightly on LCBench and PD1, showing the superiority of multi-fidelity BO to black-box BO.

\paragraph{Effectiveness of our acquisition function.} 
Next, Table~\ref{tab:main} shows the performance of each method on the cost-sensitive multi-fidelity HPO setup ($\alpha > 0$). We see that our method largely outperforms all the methods on all the settings, including the multi-fidelity HPO and the transfer-BO methods, in terms of both normalized regret and average rank. Notice, our method achieves better average rank as the penalty becomes stronger ($\alpha=2e\text{-}04$). Fig.~\ref{fig:main} visualizes the normalized regret over the course of BO, where our method achieves significantly lower regret upon termination. Our method tends to achieve the minimum regret earlier than the baselines, demonstrating its sample efficiency in searching good hyperparameter configurations by explicitly considering the utility during the BO.

In order to clearly understand the source of improvements, we next analyze the configurations chosen by each method. Specifically, for each BO step $b$, we run the configuration currently selected at step $b$ up to its last epoch $T$, and compute its minimum ground-truth regret achievable at some future step $b + \Delta t$ (Fig.~\ref{fig:acq1}), as well as the corresponding optimal increment $\Delta t$ (Fig.~\ref{fig:acq2}). In Fig.~\ref{fig:acq1}, our method shows much lower minimum regret than the baselines. It means that our acquisition function in Eq.~\eqref{eq:acquisition} works as intended, trying to select at each BO step the best configuration which is expected to maximally improve the utility in future. Fig.~\ref{fig:acq2} shows that the configurations chosen by our method initially correspond to greater $\Delta t$ (i.e., non-greedy), but gradually to the smaller $\Delta t$ (i.e., greedy). It is because as the BO proceeds, the performance improvements of BO saturate, so the cost of BO quickly dominates the trade-off, leading to smaller $\Delta t$ even close to $0$.
On the other hand, the tendencies of the baselines are very noisy and relatively unclear.
Lastly, Fig.~\ref{fig:acq3} shows the distribution of the top-10 most frequently selected configurations during the BO. As expected, our method tend to focus only on a few configurations during the BO to maximize the short-term performances, especially when the penalty is stronger with greater $\alpha$. On the other hand, the baselines tend to overly explore the configurations even when the penalty is the strongest ($\alpha = 2e\text{-}04$).

\vspace{-0.07in}
\paragraph{Effectiveness of our stopping criterion.} Lastly, we analyze the effectiveness of our stopping criterion discussed in Eq.~\eqref{eq:stop}, \eqref{eq:threshold}, and \eqref{eq:pi}. Fig.~\ref{fig:stop_ablation} shows the normalized regret over the different values of $\beta$, a mixing coefficient between the two extreme stopping criteria, as discussed in \S\ref{sec:ustar}. $\beta \rightarrow 0$ corresponds to the criterion used by the baselines which is only based on the estimated normalized regret, whereas $\beta \rightarrow +\infty$ corresponds to the hard thresholding only based on the PI. We can see that the optimal criterion is achieved by smoothly mixing between the two ($\beta = e^{-1}$), demonstrating the superiority of our stopping criterion to the one used by the baselines ($\beta \rightarrow 0$).


\vspace{-0.1in}
\section{Conclusion}\label{sec:conclusion}
\vspace{-0.1in}
In this paper, we discussed cost-sensitive multi-fidelity BO, a novel framework for improving the sample efficiency of HPO. Based on the assumption that users want to early-stop the BO when the utility saturates, we explained how to achieve the maximum utility with our novel acquisition function and the stopping criterion specifically tailored to this problem setup, as well as the novel transfer learning method for training a sample efficient in-context LC extraploator. We empirically demonstrated the effectiveness of our method over the previous multi-fidelity HPO and the transfer-BO methods, with the numerous empirical evidences strongly supporting our claim. 
\vspace{-0.2cm}
\paragraph{Limitations.} Although our method sheds light on improving the efficiency of HPO, there remain a few limitations. 
First, we assumed that the utility function is given, but instead we could learn it from data provided by each user.
Second, our LC extrapolator is prone to overfitting even with the mixup strategy when the training dataset is small. Thus, we need a theoretically grounded way to incorporate the prior on LCs and infer the corresponding posterior. 
Lastly, PFNs assume the conditional independencies between the outputs and thus generate very noisy LCs, which may distort the estimation of future utilites. Solving them can be an interesting extension of our work.

\vspace{-0.05in}

\small
\bibliography{reference}

\newpage
\appendix

\section{Details on Benchmarks and Data Preprocessing}\label{sec:data}
In this section, we elaborate the details on the LC benchmarks and data preprocessing we have done.

\paragraph{LCBench} We use [APSFailure, Amazon\_employee\_access, Australian, Fashion-MNIST, KDDCup09\_appetency, MiniBooNE, adult, airlines, albert, bank-marketing, blood-transfusion-service-center, car, christine, cnae-9, connect-4, covertype, credit-g, dionis, fabert, helena] for training LC extrapolator. We evaluate it on [higgs, jannis, jasmine, jungle\_chess\_2pcs\_raw\_endgame\_complete, kc1, kr-vs-kp, mfeat-factors, nomao, numerai28.6, phoneme, segment, shuttle, sylvine, vehicle, volkert]. Each task contains 2000 LCs with 51 training epochs. We summarize the hyperparameter of LCBench in Table~\ref{tab:lcbench_hp}.

\begin{table}[H]
\centering
\vspace{-0.1in}
\caption{\small The 7 hyperparameters for \textbf{LCBench} tasks.}
\label{tab:lcbench_hp}
\small
\begin{tabular}{cccc}
\midrule[0.8pt]
\textbf{Name} & \textbf{Type} & \textbf{Vaules} & \textbf{Info} \\
\midrule[0.8pt]
\rowcolor{Gray}
batch\_size & integer & $[16, 51]$ & log \\
learning\_rate & continuous & $[0.0001, 0.1]$ & log \\
\rowcolor{Gray}
max\_dropout & continuous & $[0.0, 1.0]$ &  \\
max\_units & integer & $[64, 1024]$ & log \\
\rowcolor{Gray}
momentum & continuous & $[0.1, 0.99]$ &  \\
max\_layers & integer & $[1, 5]$ \\
\rowcolor{Gray}
weight\_decay & continuous & $[1e-05, 0.1]$ & \\
\midrule[0.8pt]
\end{tabular}
\end{table}

\paragraph{TaskSet} We use [rnn\_text\_classification\_family\_seed\{19, 3, 46, 47, 59, 6, 66\}, \newline word\_rnn\_language\_model\_family\_seed\{22, 47, 48, 74, 76, 81\}, char\_rnn\_language\_model\_family\_\{seed19, 26, 31, 42, 48, 5, 74\}] for training LC extrapolator. We evaluate it on [rnn\_text\_classification\_family\_seed\{8, 82, 89\},
word\_rnn\_language\_model\_family\_seed\{84, 98, 99\},
char\_rnn\_language\_model\_family\_seed\{84, 94, 96\}]. Each task contains 1000 LCs with 50 training epochs. We summarize the hyperparameter of TaskSet in Table~\ref{tab:taskset_hp}.

\begin{table}[H]
\centering
\vspace{-0.1in}
\caption{\small The 8 hyperparameters for \textbf{Taskset} tasks.}
\label{tab:taskset_hp}
\small
\begin{tabular}{cccc}
\midrule[0.8pt]
\textbf{Name} & \textbf{Type} & \textbf{Vaules} & \textbf{Info} \\
\midrule[0.8pt]
\rowcolor{Gray}
learning\_rate & continuous & $[1e-09, 10.0]$ & log \\
beta1 & continuous & $[0.0001, 1.0]$ &  \\
\rowcolor{Gray}
beta2 & continuous & $[0.001, 1.0]$ &  \\
epsilon & continuous & $[1e-12, 1000]$ & log \\
\rowcolor{Gray}
l1 & continuous & $[1e-09, 10.0]$ & log \\
l2 & continuous & $[1e-09, 10.0]$ & log \\
\rowcolor{Gray}
linear\_decay & continuous & $[1e-08, 0.0001]$ & log \\
\midrule[0.8pt]
\end{tabular}
\end{table}

\paragraph{PD1} We use [
uniref50\_transformer\_batch\_size\_128, lm1b\_transformer\_batch\_size\_2048, \newline
imagenet\_resnet\_batch\_size\_256, mnist\_max\_pooling\_cnn\_tanh\_batch\_size\_2048, \newline mnist\_max\_pooling\_cnn\_relu\_batch\_size\_\{2048, 256\}, mnist\_simple\_cnn\_batch\_size\_\{2048, 256\}, \newline fashion\_mnist\_max\_pooling\_cnn\_tanh\_batch\_size\_2048, fashion\_mnist\_max\_pooling\_cnn\_relu\_batch\_size\_\{2048, 256\}, fashion\_mnist\_simple\_cnn\_batch\_size\_\{2048, 256\}, svhn\_no\_extra\_wide\_resnet\_batch\_size\_1024, \newline cifar\{100, 10\}\_wide\_resnet\_batch\_size\_2048] for training LC extrapolator. We evaluate it on \newline [imagenet\_resnet\_batch\_size\_512, translate\_wmt\_xformer\_translate\_batch\_size\_64, \newline mnist\_max\_pooling\_cnn\_tanh\_batch\_size\_256, fashion\_mnist\_max\_pooling\_cnn\_tanh\_batch\_size\_256, \newline svhn\_no\_extra\_wide\_resnet\_batch\_size\_256,  cifar100\_wide\_resnet\_batch\_size\_256, \newline cifar10\_wide\_resnet\_batch\_size\_256]. Each task contains 240 LCs with 50 training epochs. We summarize the hyperparameter of PD1 in Table~\ref{tab:pd1_hp}.

\begin{table}[H]
\centering
\vspace{-0.1in}
\caption{\small The 8 hyperparameters for \textbf{PD1} tasks.}
\label{tab:pd1_hp}
\small
\begin{tabular}{cccc}
\midrule[0.8pt]
\textbf{Name} & \textbf{Type} & \textbf{Vaules} & \textbf{Info} \\
\midrule[0.8pt]
\rowcolor{Gray}
lr\_initial\_value & continuous & $[1e-05, 10.0]$ & log \\
lr\_power & continuous & $[0.1, 2.0]$ &  \\
\rowcolor{Gray}
lr\_decay\_steps\_factor & continuous & $[0.01, 0.99]$ &  \\
one\_minus\_momentum & continuous & $[1e-05, 1.0]$ & log \\
\midrule[0.8pt]
\end{tabular}
\end{table}

\paragraph{Data Preprocessing} 
As will be detailed in the \S\ref{sec:additional_setups}, we use the 0-epoch LC value $y_{n, 0}$ which is the performance before taking any gradient steps. The 0-epoch LC values originally are not provided except for LCBench; we use the log-loss of the first epoch as the 0-epoch LC value for TaskSet, as it is already sufficiently large in our chosen tasks. For PD1, we interpolate the LCs to be the length of 51 training epochs, and we take the first performance as the 0-epoch LC value. Furthermore, we take the average over the 0-epoch LC values $\bar{y}_0$ since it is hard to have different initial values among optimizer hyperparameter configurations in a task, without taking any gradient steps. For transfer learning, we follow the convention of PFN~\cite{adriaensen2024efficient} for data preprocessing; we consistently apply non-linear LC normalization\footnote{The details can be found in \href{https://arxiv.org/pdf/2310.20447}{Appendix A} of PFN~\cite{adriaensen2024efficient} and \url{https://github.com/automl/lcpfn/blob/main/lcpfn/utils.py}.} to the LC data of three benchmarks, which not only maps either accuracy or log-loss LCs into $[0, 1]$ but also simply make our optimization as a maiximization problem. To facilitate transfer learning, we use the maximum and minimum values in each task in LCBench and PD1 benchmark for the LC normalization. In TaskSet, we only use the $\bar{y}_0$ for the LC normalization. 

\section{Details on Architecture and Training of LC Extrapolator}\label{sec:pretraining_details}

In the section, we elaborate our LC extrapolator model and how to train it on the learning curve dataset. 

\paragraph{Construction of Context and Query points.} As mentioned earlier in \S \ref{sec:transfer_learning}, the whole training pipeline of our learning curve extrapolator model can be seen an instance of TNPs~\cite{nguyen2022transformer}. Here we can simulate each step of Bayesian Optimization; predicting the remaining part of LC in all configurations conditioned on the set $\mathcal{C}$ of the collected partial LCs. To do so, we construct a training task by randomly sampling context and query points from LC benchmark after the proposed LC mixup as follows:

\begin{enumerate}[itemsep=1mm, parsep=0pt, leftmargin=*]
\item We choose a LC dataset $L_m = [l_{m, 1}^\top ; \dots ; l_{m, N}^\top]^\top \in \mathbb{R}^{N\times T}$ by randomly sampling $m\in[M]$.
\item From $L_m$, we randomly sample $n_1, \ldots, n_C \in [N]$ and $t_1, ..., t_C \in [T]$ and construct context points of $X^{(c)} = [ x_{n_1}^\top, \ldots,  x_{n_C}^\top]^\top \in \mathbb{R}^{C \times d_x}$, $T^{(c)} = [ t_1/T, \ldots, t_C/T]^\top \in \mathbb{R}^{C \times 1}$, and $Y^{(c)} = [y_{n_1, t_1}, \ldots, y_{n_C, t_C}]\in\mathbb{R}^{C \times 1}$.
\item From $L_m$, we exclude $n_1, \ldots, n_C \in [N]$ and $t_1, ..., t_C \in [T]$ and randomly sample $n'_1, \ldots, n'_Q \in [N]$ and $t'_1, ..., t'_Q \in [T]$ and construct query points of $X^{(q)} = [ x_{n'_1}^\top, \ldots,  x_{n'_Q}^\top]^\top \in \mathbb{R}^{Q \times d_x}$, $T^{(q)} = [ t'_1/T, \ldots, t'_C/T]^\top \in \mathbb{R}^{Q \times 1}$, and $Y^{(q)} = [y_{n'_1, t'_1}, \ldots, y_{n'_Q, t'_Q}]\in\mathbb{R}^{Q \times 1}$.
\end{enumerate}

\paragraph{Transformer for Predicting Learning Curves.} From now on, we denote each row vector of the constructed context and query points with the lowercase, e.g., $y^{(q)}$ of $Y^{(q)}$. We learn a Transformer-based learning curve extrapolator model which is a probabilistic model of $f( Y^{(q)} | X^{(c)}, T^{(c)}, Y^{(c)}, X^{(q)}, T^{(q)} )$. Conditioned on any subsets of LCs (i.e., $X^{(c)}, T^{(c)}$, and $Y^{(c)}$), this model predicts a mini-batch of the remaining part of LCs of existing hyperparameter configurations in a given dataset (i.e., $Y^{(q)}$ of $X^{(q)}$ and $T^{(q)}$). For the computational efficiency, we further assume that the query points are independent to each other, as done in PFN~\cite{adriaensen2024efficient}:  
\begin{equation}\label{eq:independency}
f( Y^{(q)} | X^{(c)}, T^{(c)}, Y^{(c)}, X^{(q)}, T^{(q)} ) = \prod_{x^{(q)}, t^{(q)}, y^{(q)}}  f( y^{(q)} | x^{(q)}, t^{(q)}, X^{(c)}, T^{(c)}, Y^{(c)} ).
\end{equation}

Before encoding the input into the Transformer, we first encode the input of $X^{(c)}, T^{(c)}, Y^{(c)}, X^{(q)}$, and $T^{(q)}$ using simple linear layer as follows:
\begin{gather}
H^{(c)} = X^{(c)}W_x + T^{(c)}W_t + Y^{(c)}W_y  \\
H^{(q)} = X^{(q)}W_x + T^{(q)}W_t,    
\end{gather}
where $W_x \ \in \mathbb{R}^{d_x \times d_h}$, $W_t \in \mathbb{R}^{1 \times d_h}$, and  $W_y \in \mathbb{R}^{1 \times d_h}$. Here, we abbreviate the bias term.

Then we concatenate the encoded represnetations of $H^{(c)}$ and $H^{(q)}$, and feedforward it into Transformer layer by treating each pair of each row vector of $H^{(c)}$ and $H^{(q)}$ as a separate position/token as follows:
\begin{gather}
H=\mathtt{Transformer}([H^{(c)} ; H^{(q)}, \mathtt{Mask}]) \in \mathbb{R}^{(M+N)\times d_h} \\
\hat{Y}=\mathtt{Head}(H)\in\mathbb{R}^{(M+N)\times d_o},
\end{gather}
where $\mathtt{Transformer}(\cdot)$ and $\mathtt{Head}(\cdot)$ denote the Transformer layer and multi-layer perceptron (MLP) for the output prediction, respectively. $\mathtt{Mask} \in \mathbb{R}^{(N_c+N_q)\times(N_c+N_q)}$ is the mask of transformer that allows all the tokens to attend context tokens only. Here, the output dimension $d_o$ is specified by output distribution of $y$. Following PFN~\cite{adriaensen2024efficient}, we discretize the domain of $y$ by $d_o = 1000$ and use the categorical distribution. Finally, we only take the output of the last $N_q$ tokens as output, i.e., $\hat{Y}^{(q)}=\hat{Y}[:, N_c:(N_c+N_q)] \in \mathbb{R}^{N_q\times d_h}$ (PyTorch-style indexing operation), since we only need the outputs of query tokens for modeling $\prod f( y^{(q)} | x^{(q)}, t^{(q)}, X^{(c)}, T^{(c)}, Y^{(c)} )$.

\paragraph{Training Objective.} Our pre-training objective is then defined as follows:
\begin{equation}\label{eq:loss}
\argmin_f \mathbb{E}_p\left[-\sum_{x^{(q)}, t^{(q)}, y^{(q)}} \log f( y^{(q)} | x^{(q)}, t^{(q)}, X^{(c)}, T^{(c)}, Y^{(c)} ) \right] \\
+ \lambda_{\text{PFN}}\mathcal{L}_{\text{PFN}}
,
\end{equation}
where $\mathcal{D}_{KL}$ is the Kullback–Leibler divergence, and $p$ is the empirical LC data distribution. We additionally minimize $\mathcal{L}_{\text{PFN}}$ with coefficient $\lambda_{\text{PFN}}$, which is the LC extrapolation loss in each LC \cite{adriaensen2024efficient}. We found $\lambda_{\text{PFN}}=0.1$ works well for most cases. We use the stochastic gradient descent algorithm to solve the above optimization problem. 

\paragraph{Training Details.} We sample 4 training tasks for each iteration, i.e., the size of meta mini-batch is set to 4. We uniformly sample the size $C$ of context points from 1 to 300, and the size of query points $Q$ is set to 2048. Following PFN~\cite{adriaensen2024efficient}, the hidden size of each Transformer block $d_h$, the hidden size of feed-forward networks, the number of layers of Transformer, dropout rate are set 1024, 2048, 12, 0.2. We use GeLU~\cite{hendrycks2016gaussian}. We train the extrapolator for 10,000 iterations on training split of each benchmark with Adam~\cite{kingma2014adam} optimizer. The $\ell_2$ norm of meta mini-batch gradient is clipped to 1.0. The learning rate is linearly increased to 2e-05 for 25000 iterations, and it is decreased with a cosine scheduling until the end. The whole training process takes roughly 10 hours in one NVIDIA Tensor Core A100 GPU.

\section{Additional Details on Experimental Setups}\label{sec:additional_setups}
In this section, we elaborate additional details on the experimental setups.

\paragraph{0-epoch LC value.} We assume the access of the 0-epoch LC value $\bar{y}_0$ in \S\ref{sec:data} which is the model performance before taking gradient steps. This is also plausible for realistic scenarios since in most deep-learning models one evaluation cost is acceptable in comparison to training costs. The 0-epoch LC value $\bar{y}_0$ is always conditioned on our LC extrapolator $f$ for both pretraining and BO stage.

\paragraph{Monte-Carlo (MC) sampling for reducing variance of LCs.} As mentioned in \S\ref{sec:ustar}, we estimate the expectation of proposed acquisition function $A$ in Eq.~(\ref{eq:acquisition}) with 1000 MC samples. We found that each LC $y_{n,t_{n:T}}$ sampled from LC extrapolator $f(\cdot|x_n, \mathcal{C})$ is noisy, due to the assumption that query points of $y_{n,t_{n:T}}$ are independent to each other in Eq.~(\ref{eq:independency}). We compute $\tilde{y}_{b+\Delta t}$ by taking the maximum among the last step BO performance (i.e., cumulative max operation), therefore, the quality of estimation highly degenerates due to the noise in the small $\Delta t$. To prevent this, we reduce the variance of MC samples by taking the average of the sampled LCs. For example, we sample 5000 LC samples from the LC extrapolator $f$, then we divide them into 1000 groups and take the average among the 5 LC samples in each group. We empirically found that this stabilize the estimation of not only acquisition function $A$ and probability of utility improvement $p_b$ in Eq.~(\ref{eq:pi}).

\paragraph{Inference Time for BO.} The most of time for each BO step in our method is spent during LC extrapolation. In Table~\ref{tab:wallclock}, we report the wall-clock time spent on LC extrapolation for 100 mini-batches of LCs. The wall-clock times vary depending on the context size. We measure all the wall-clock times in one one NVIDIA Tensor Core A100 GPU.
\definecolor{Gray}{gray}{0.9}

\begin{table}[H]
\centering
\caption{\small \textbf{Wall-clock time for Inference} on 100 mini-batches of LCs. }
\label{tab:wallclock}
\small
\begin{tabular}{cc}
\midrule[0.8pt]
\textbf{Context Size} & \textbf{Inference Time (s)} \\
\midrule[0.8pt]
\rowcolor{Gray}
1 & 0.00921010971069336 \\
10 & 0.01493692398071289 \\
\rowcolor{Gray}
20 & 0.01413583755493164 \\
50 & 0.017796993255615234 \\
\rowcolor{Gray}
100 & 0.01770782470703125 \\ 
200 & 0.025087356567382812 \\
\rowcolor{Gray}
300 & 0.027765989303588867 \\
\midrule[0.8pt]
\end{tabular}
\end{table}

\newpage

\paragraph{Details on Baseline Implementation.} We list the implementation detils for baselines as follows:

\begin{enumerate}[itemsep=1mm, parsep=0pt, leftmargin=*]
\item \textbf{Random Search.} Instead of randomly selecting a hyperparameter configuration for each BO step, we run the selected configuration until the last epoch $T$.

\item \textbf{ASHA, BOHB, and DEHB.} We follow the most recent implementation of these algorithms in Quick-Tune~\cite{arango2023quick}. We slightly modify the official code\footnote{\url{https://github.com/releaunifreiburg/QuickTune}}, which is heavily based on SyneTune~\cite{salinas2022syne} package. 

\item \textbf{DyHPO and Quick-Tune$\boldsymbol{^\dagger}$.} We follow the official code\footnote{\url{https://github.com/releaunifreiburg/DyHPO}} provided the authors of DyHPO~\cite{wistuba2022supervising}, and slightly modify the benchmark implementation to incorporate our experimental setups. For Quick-Tune$\boldsymbol{^\dagger}$, we pretrain the deep kernel GP for 50000 iterations with Adam optimizer with mini-batch size of 512. The initial learning rate is set to 1e-03 and decayed with cosine scheduling. To leverage the transfer learning scenario, we use the best configuration among the LC datasets which is used for training the GP as an initial guess of BO.

\item \textbf{DPL.} We follow the official code\footnote{\url{https://github.com/releaunifreiburg/DPL}} provided the authors of DPL~\cite{kadra2024scaling}, and slightly modify the benchmark implementation to incorporate our experimental setups.

\item \textbf{FSBO.} FSBO does not provide an official code, therefore, we follow an available code in the internet\footnote{\url{https://github.com/releaunifreiburg/fsbo}}. We also slightly modify the benchmark implementation, and use the best configuration among the LC datasets as an initial guess. 
\end{enumerate}


\section{Connection between our Mixup Strategy with ifBO and TNP}
\label{sec:mixup_related_work}
Our mixup strategy is reminiscent of the data generation scheme of ifBO~\cite{rakotoarison2024context}, a variant of PFNs for in-context freeze-thaw BO. 
Similarly to our ancestral sampling, ifBO first samples random weights for a neural network (i.e., a prior distribution) to sample a correlation between configurations (the first mixup step), and then linearly combines a set of basis functions to generate LCs (the second mixup step). Our training method differs from ifBO in that our prior distribution is implicitly defined by LC datasets and the mixup strategy, whereas ifBO resorts to a manually defined distribution. 

Indeed, our training method is more similar to Transformer Neural Processes (TNPs)~\cite{nguyen2022transformer}, a Transformer variant of Neural Processes (NPs)~\cite{garnelo2018neural}. Similarly to PFNs, TNPs directly maximize the likelihood of target data given context data with a Transformer architecture, which differs from the typical NP variants that summarize the context into a latent variable and perform variational inference on it. 
Moreover, as with the other NP variants, TNPs meta-learn a model over a distribution of tasks to perform sample efficient probabilistic inference. In this vein, the whole training pipeline of our LC extrapolator can be seen as an instance of TNPs -- we also meta-learn a sample efficient Transformer-based LC extrapolator over the distribution of LCs induced by the mixup strategy.

\section{Additional Experimental Results}\label{sec:additional_results}

\begin{figure}[H]
\vspace{-0.15in}
\centering
\includegraphics[width=0.32\textwidth]{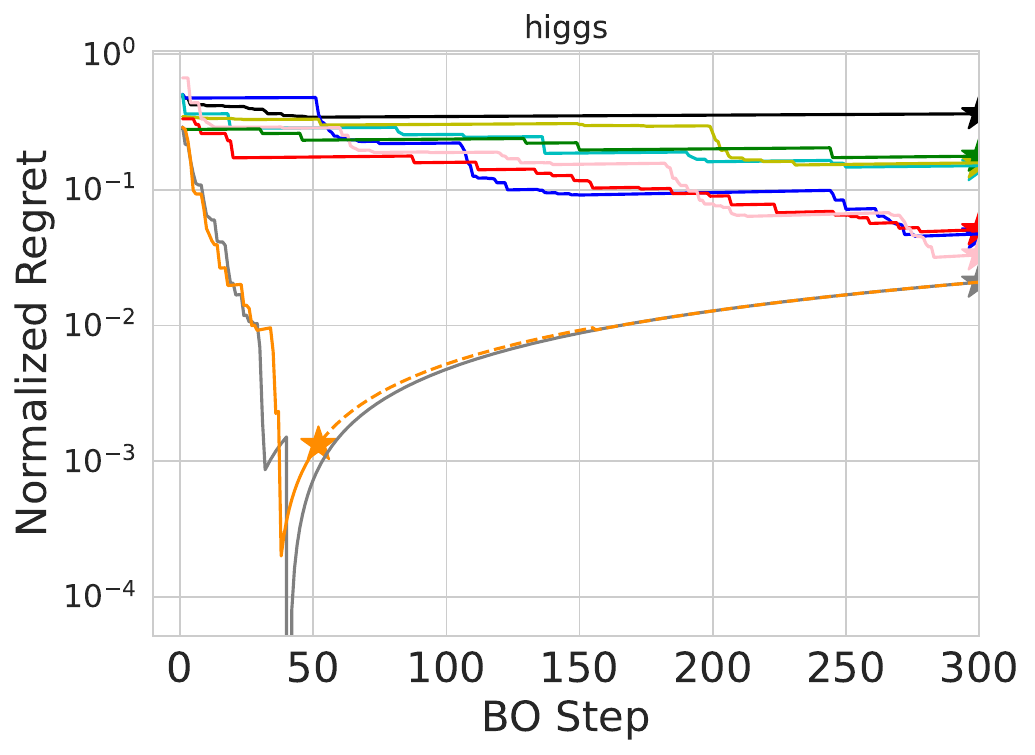}
\includegraphics[width=0.32\textwidth]{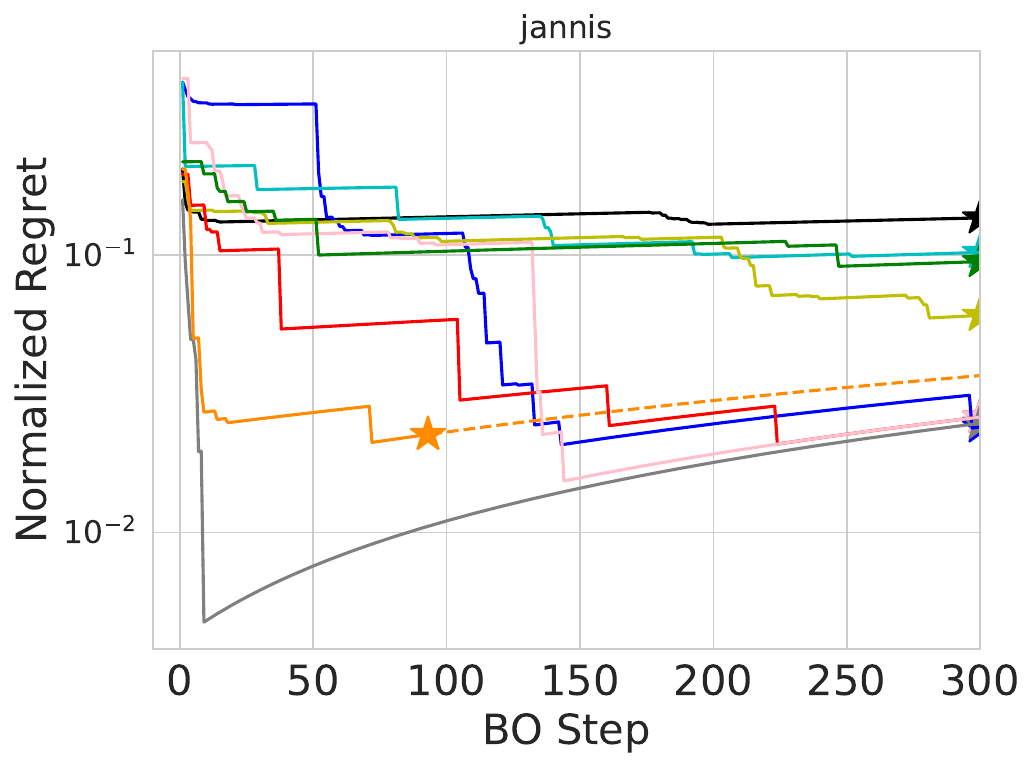}
\includegraphics[width=0.32\textwidth]{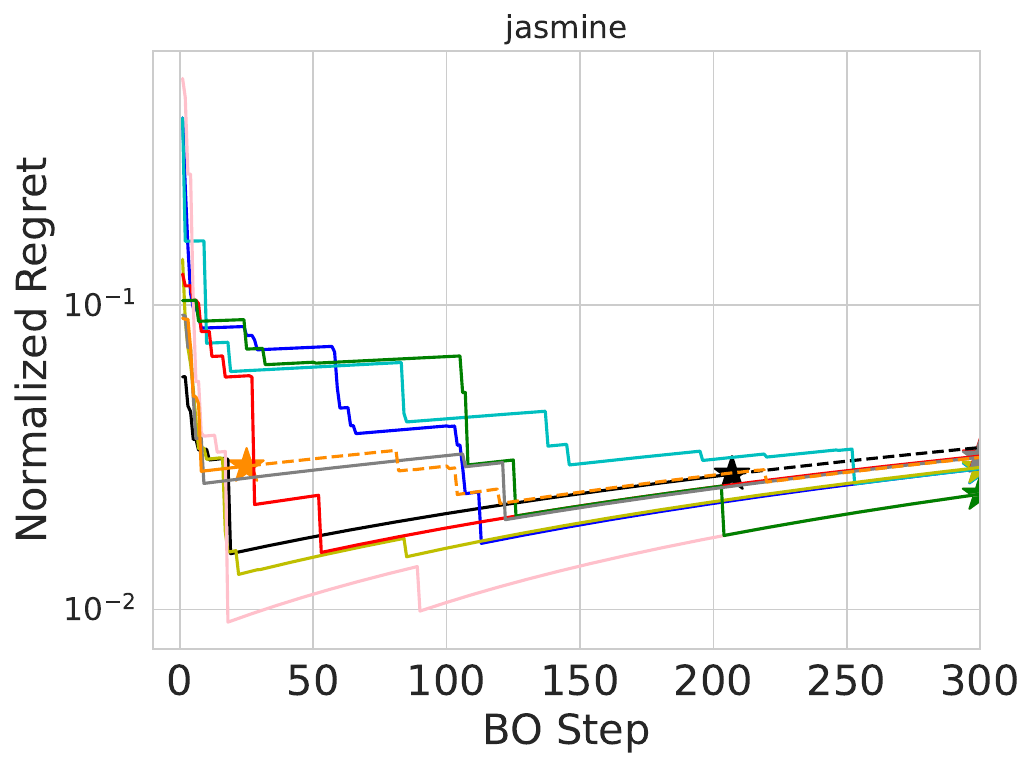}
\includegraphics[width=0.32\textwidth]{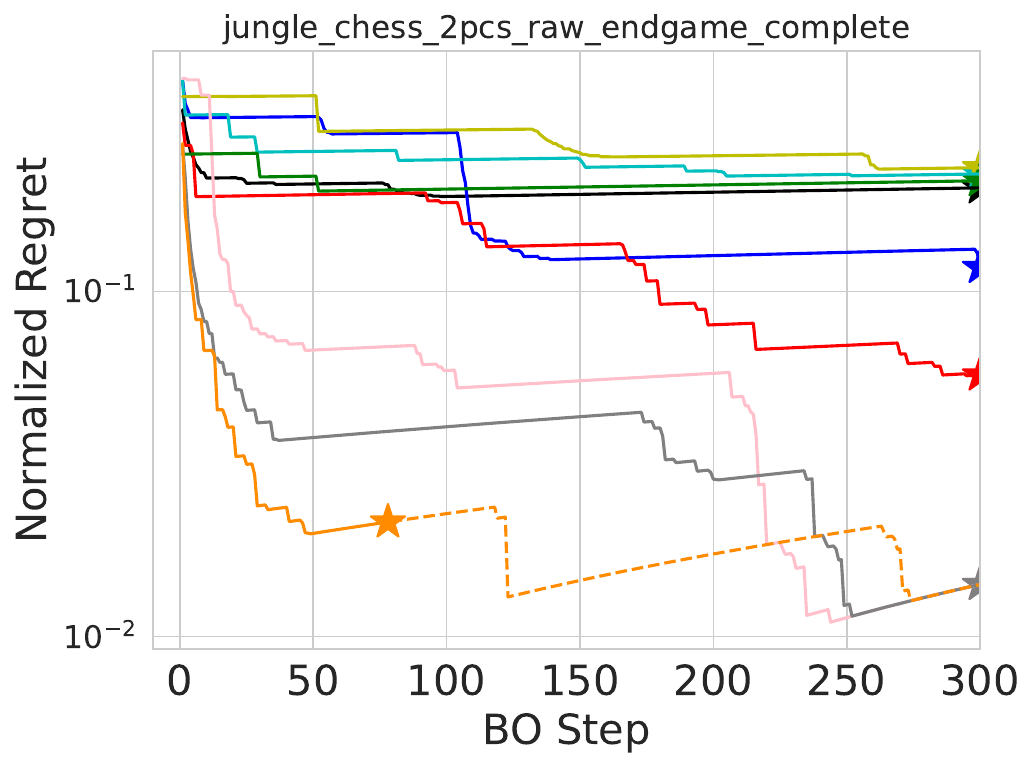}
\includegraphics[width=0.32\textwidth]{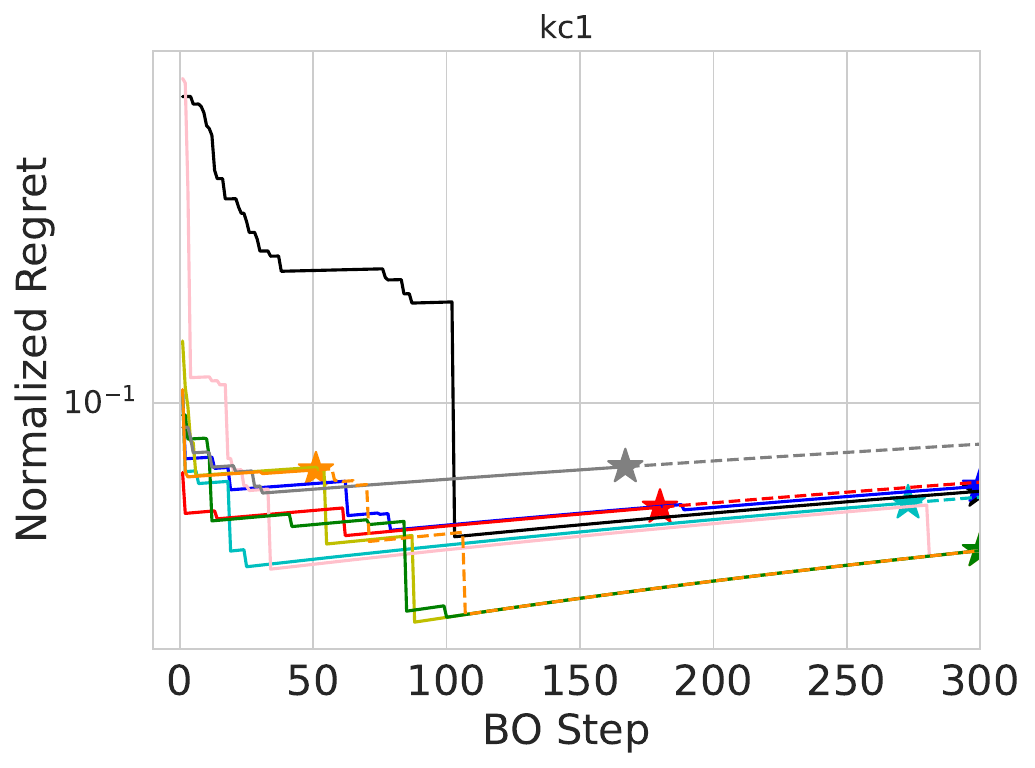}
\includegraphics[width=0.32\textwidth]{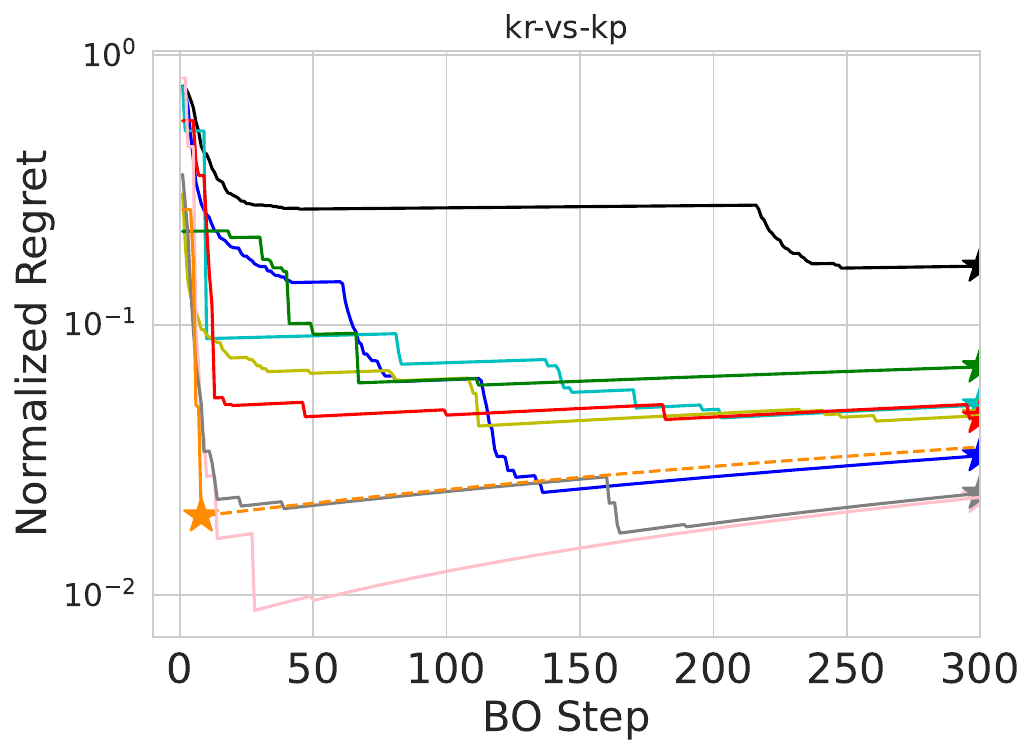}
\includegraphics[width=0.32\textwidth]{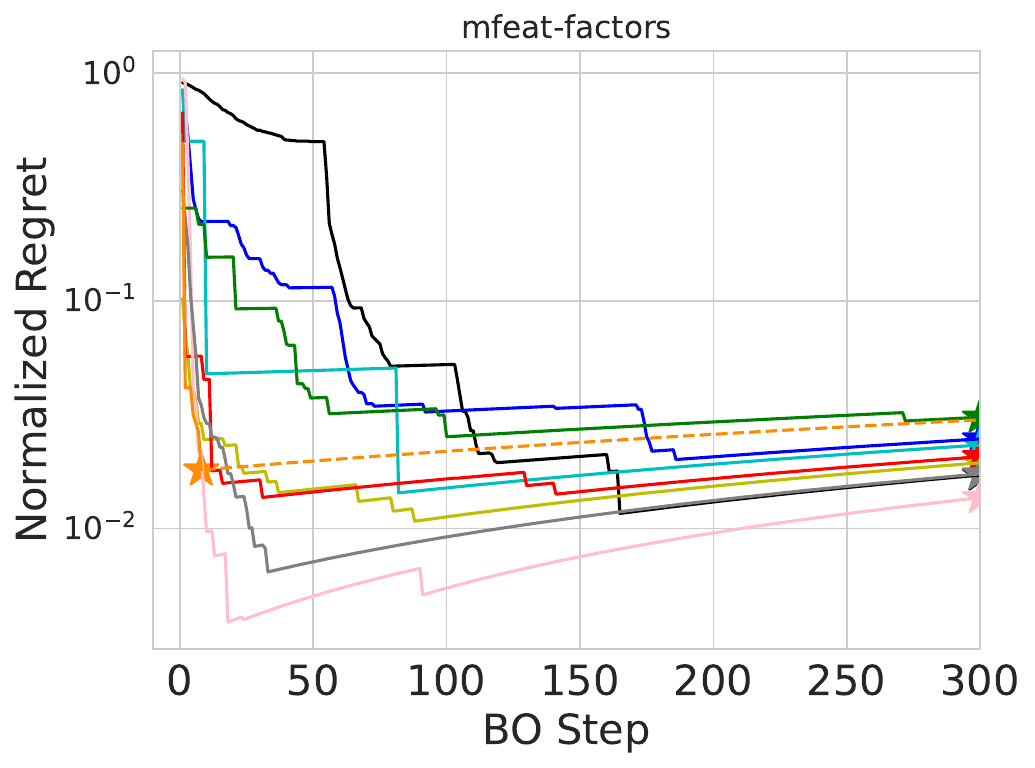}
\includegraphics[width=0.32\textwidth]{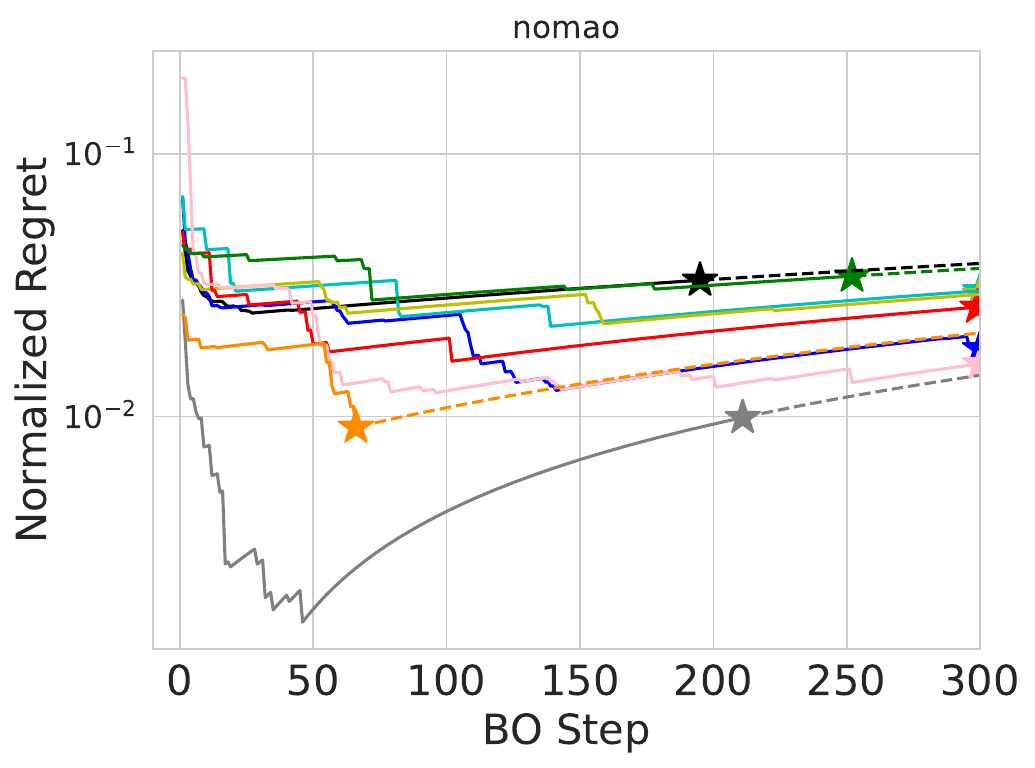}
\includegraphics[width=0.32\textwidth]{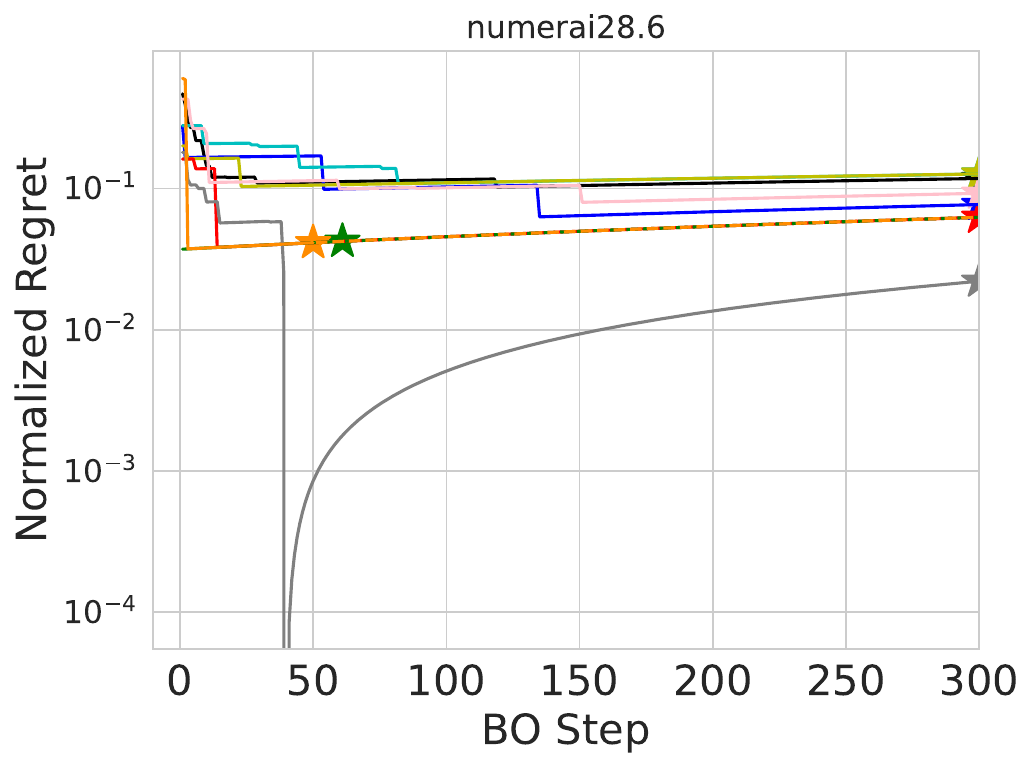}
\includegraphics[width=0.32\textwidth]{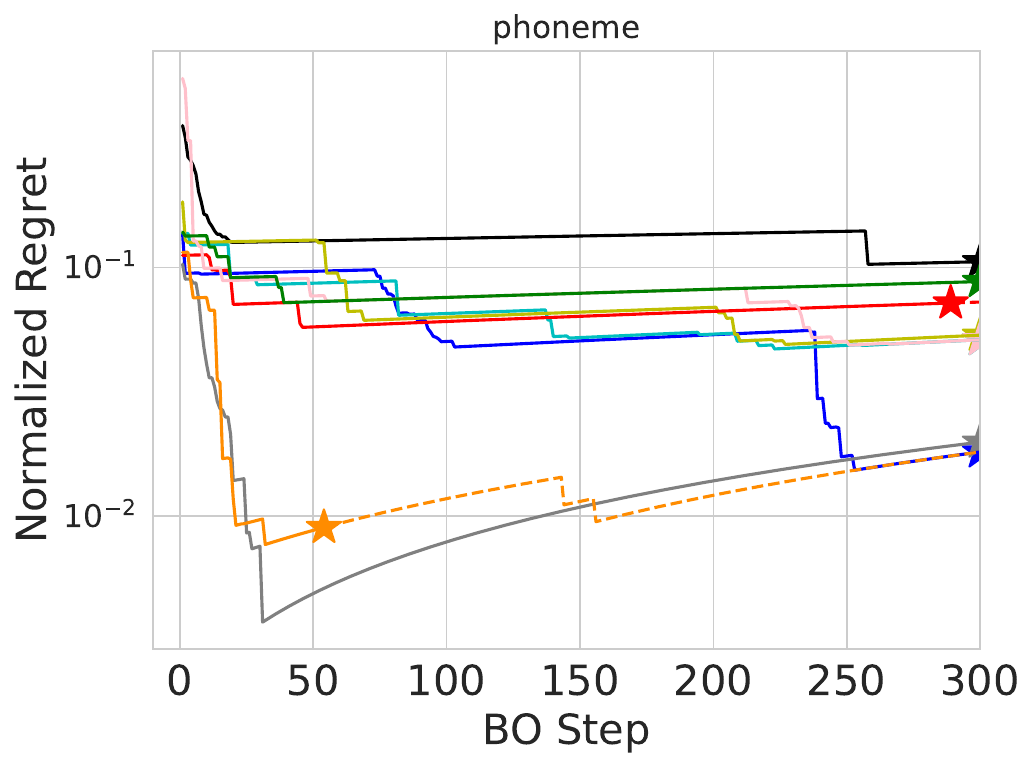}
\includegraphics[width=0.32\textwidth]{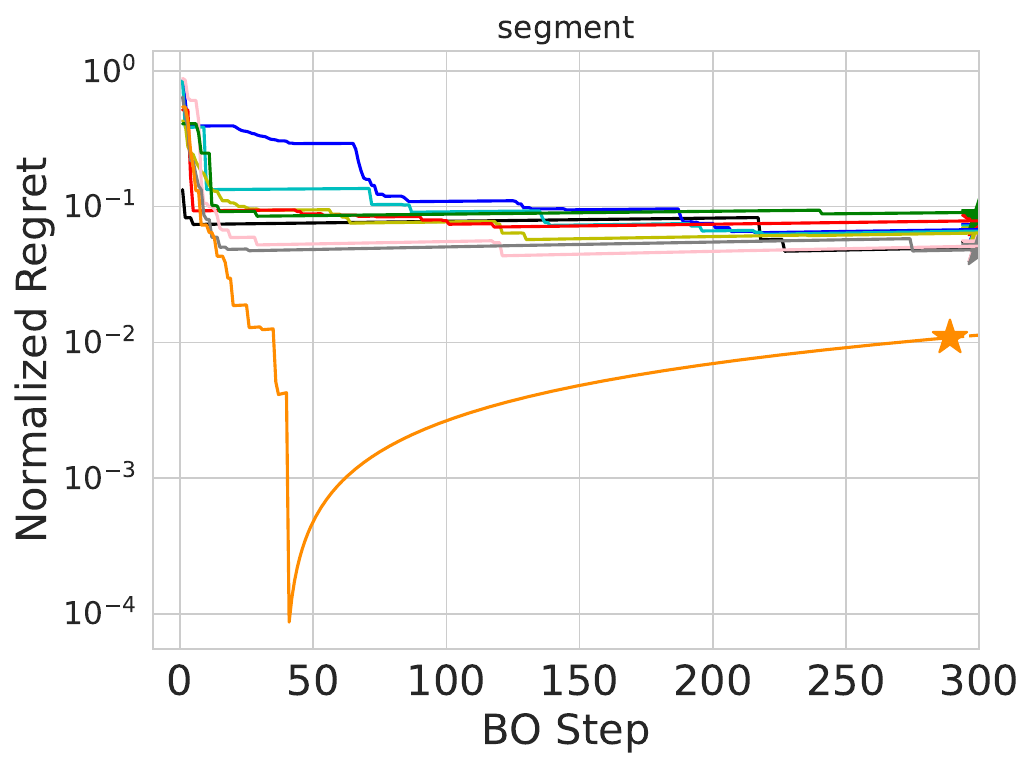}
\includegraphics[width=0.32\textwidth]{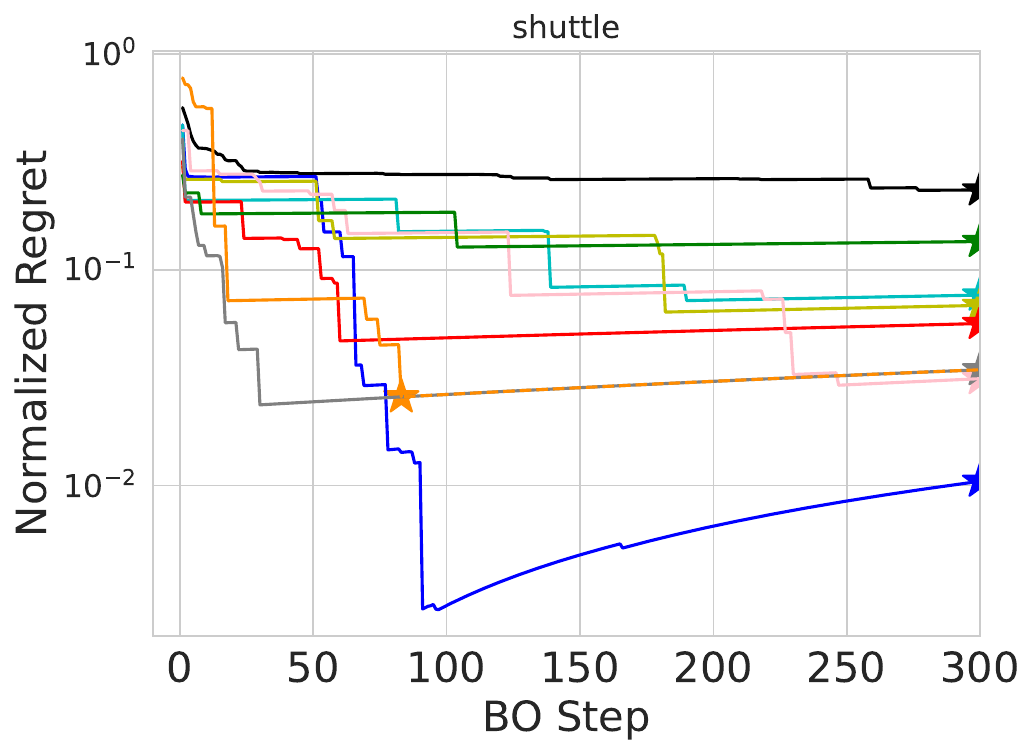}
\includegraphics[width=0.32\textwidth]{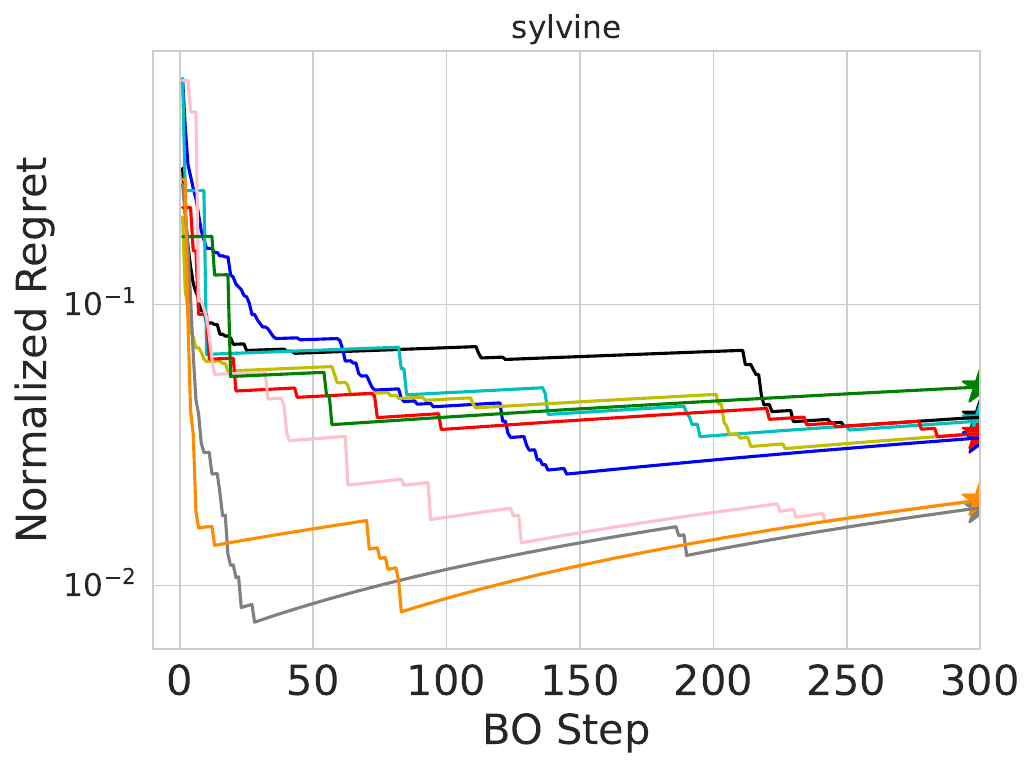}
\includegraphics[width=0.32\textwidth]{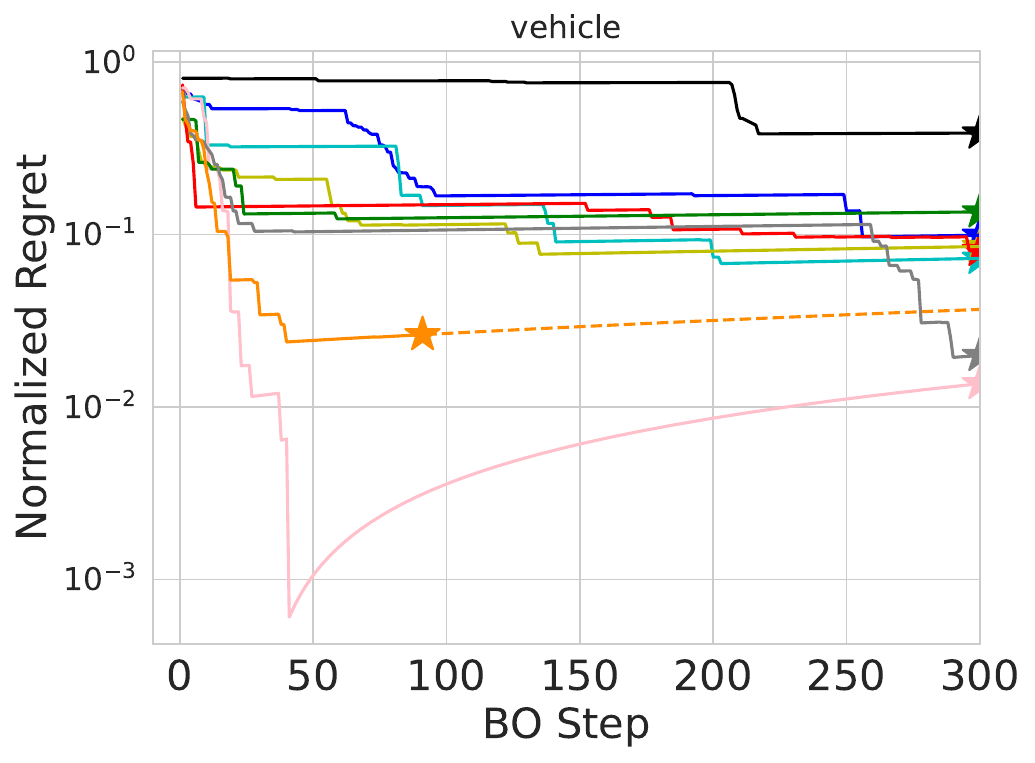}
\includegraphics[width=0.32\textwidth]{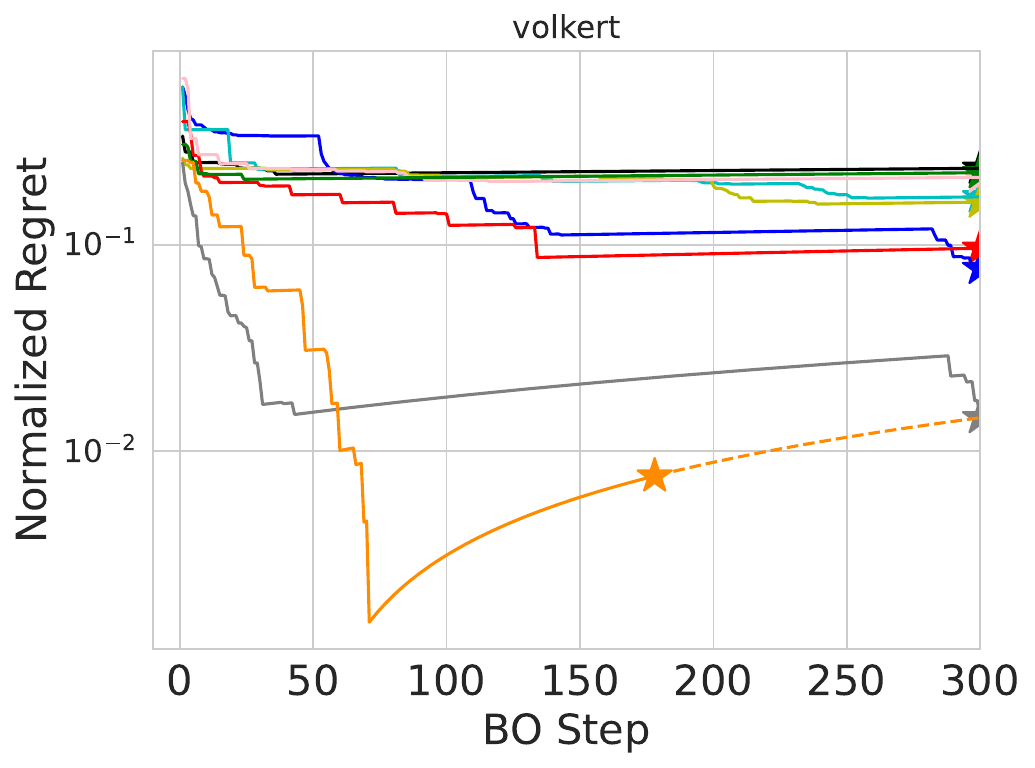}
\vspace{-0.17in}
\medskip
\includegraphics[width=1.0\textwidth]{figures/legend.pdf}
\caption{\small Visualization of the normalized regret over BO steps on \textbf{LCBench ($\alpha=$4e-05)}.}
\label{fig:4e-05_lcbench}
\vspace{-0.15in}
\end{figure}
\begin{figure}[H]
\vspace{-0.15in}
\centering
\includegraphics[width=0.32\textwidth]{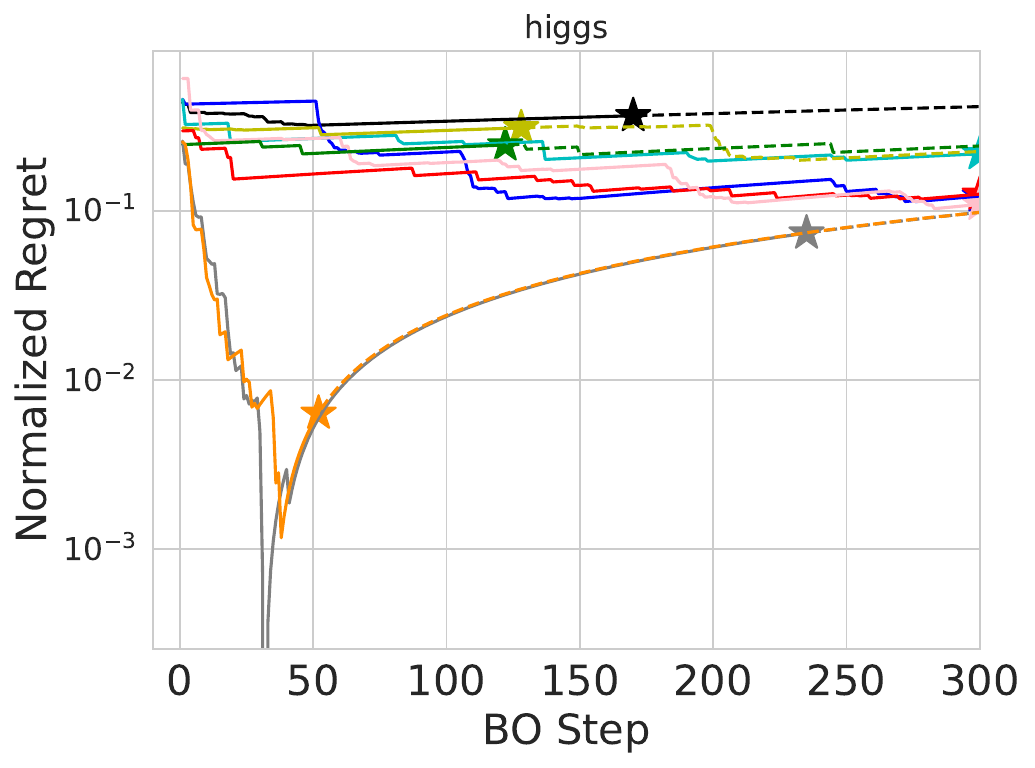}
\includegraphics[width=0.32\textwidth]{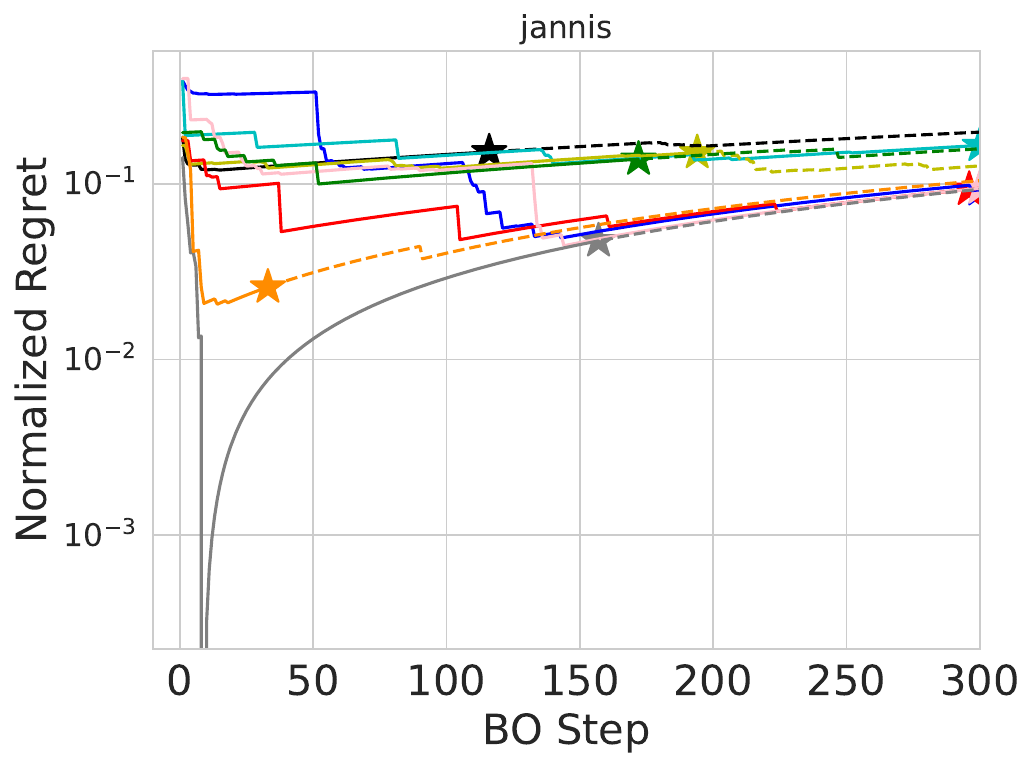}
\includegraphics[width=0.32\textwidth]{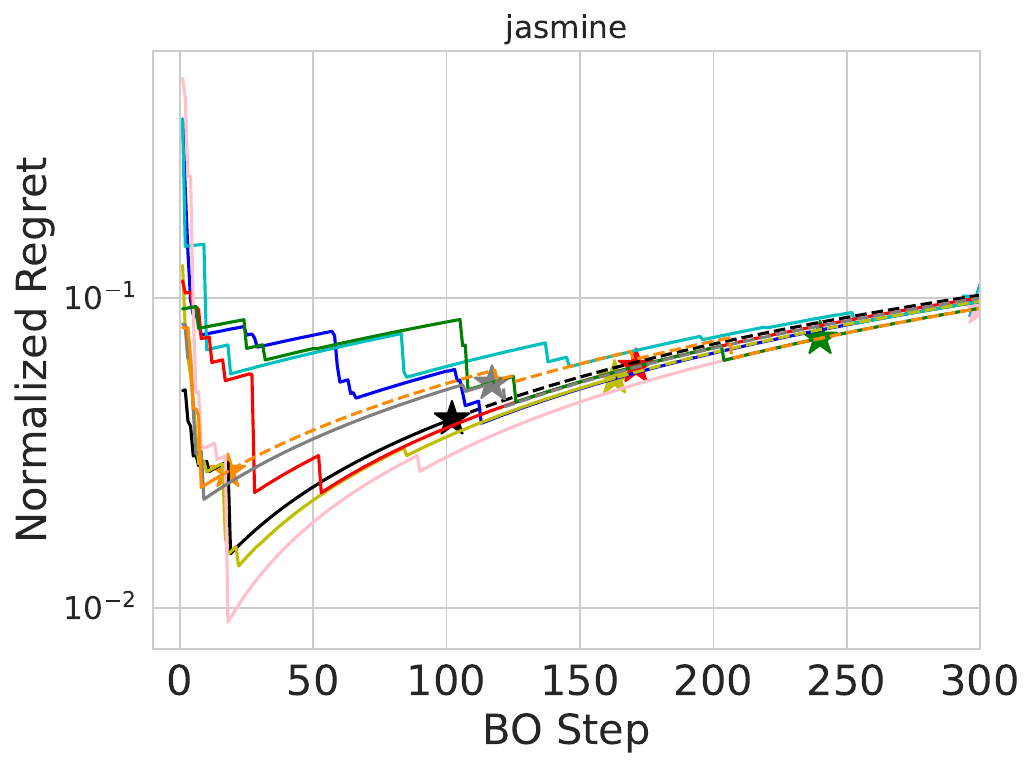}
\includegraphics[width=0.32\textwidth]{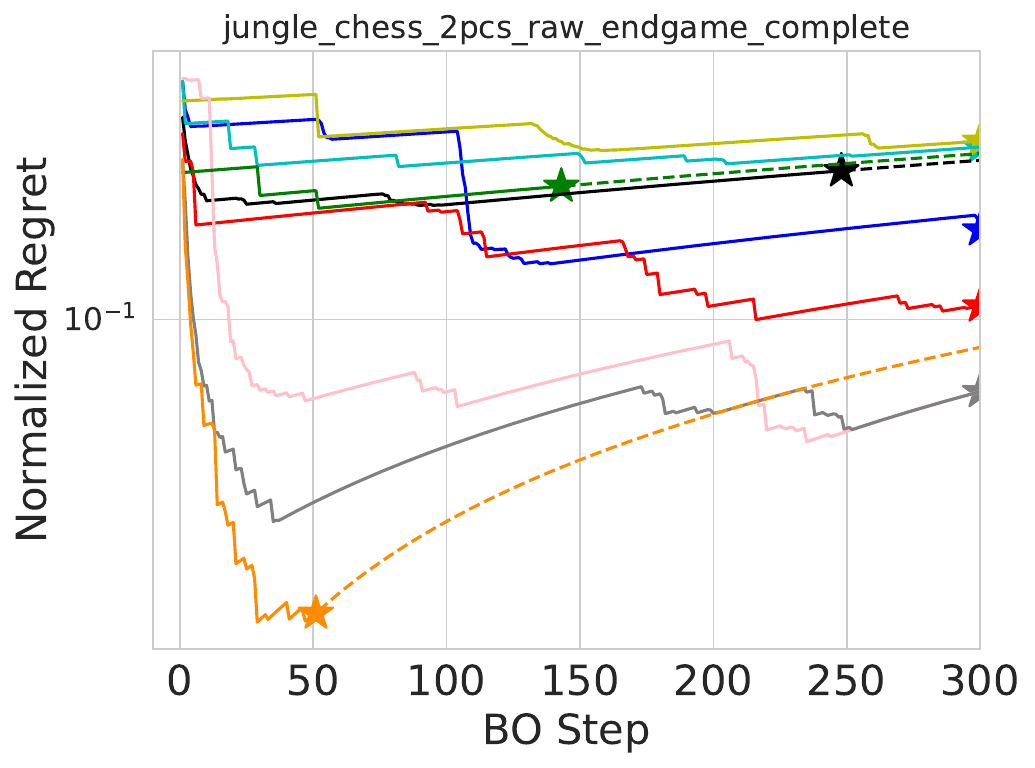}
\includegraphics[width=0.32\textwidth]{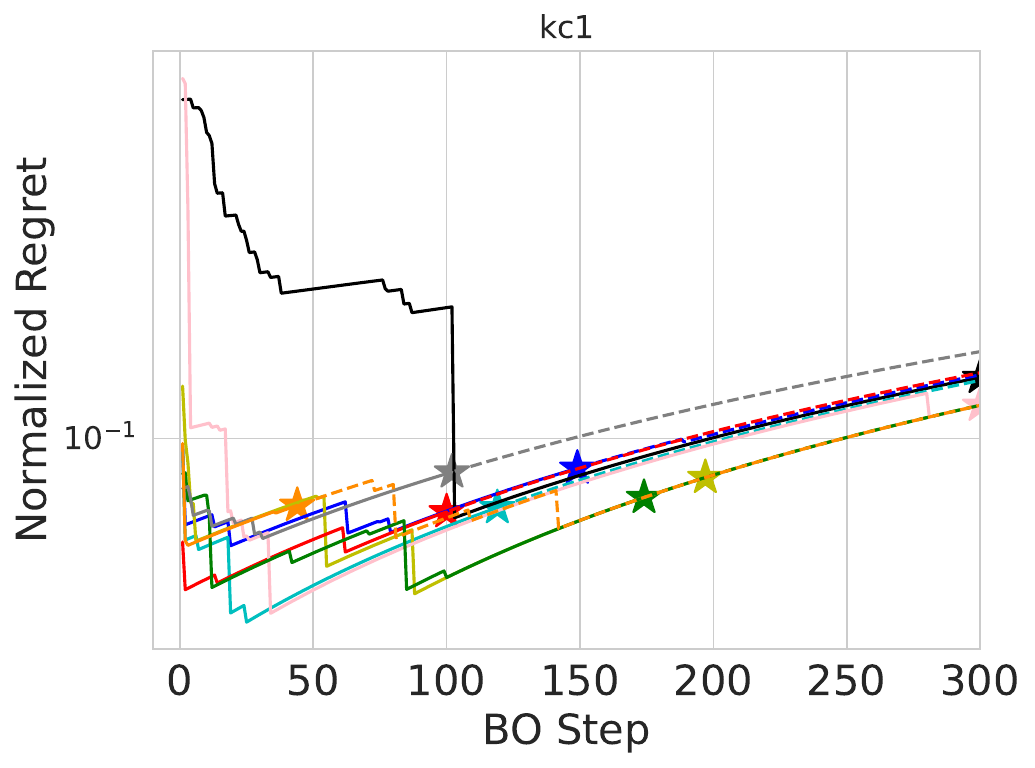}
\includegraphics[width=0.32\textwidth]{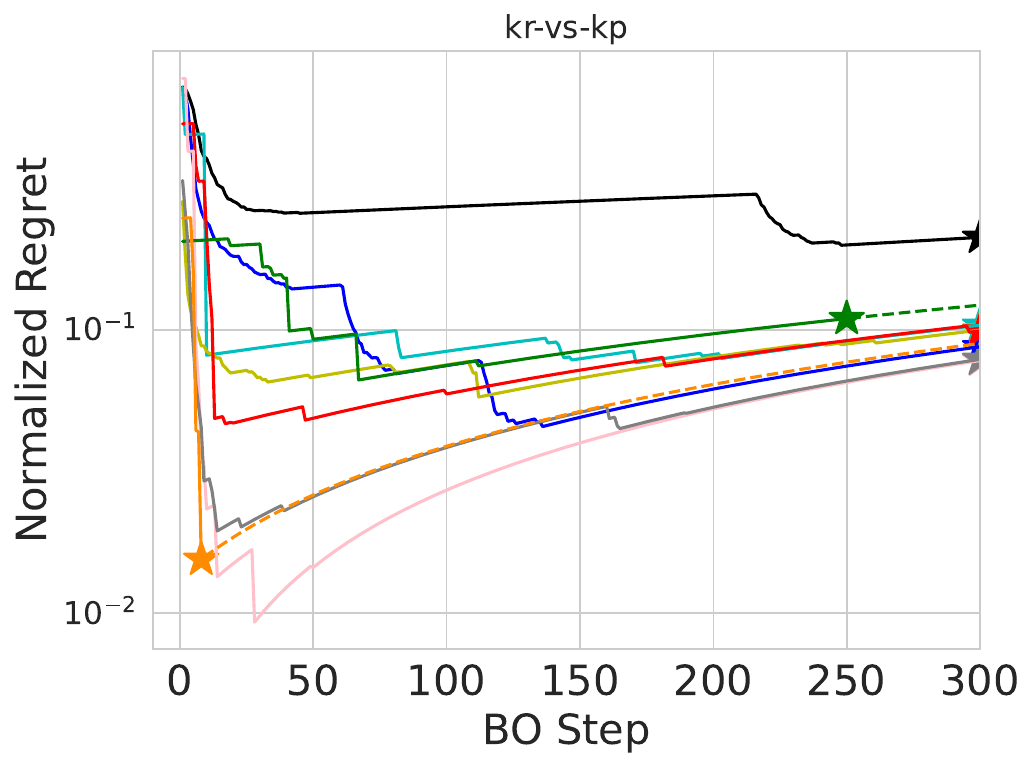}
\includegraphics[width=0.32\textwidth]{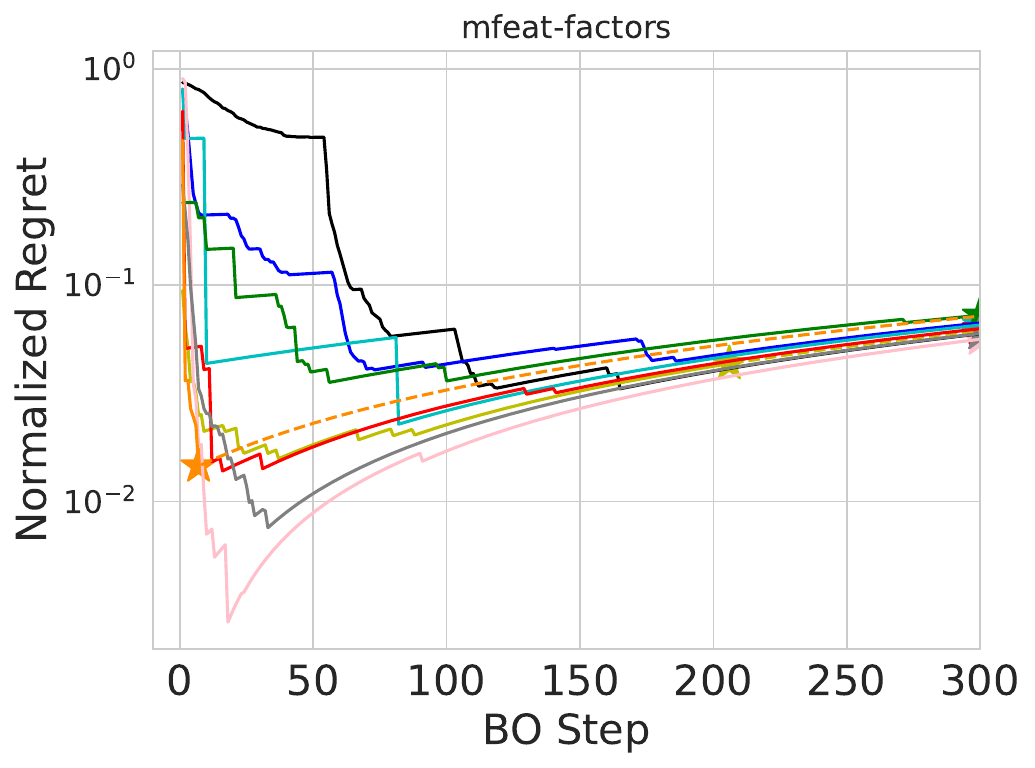}
\includegraphics[width=0.32\textwidth]{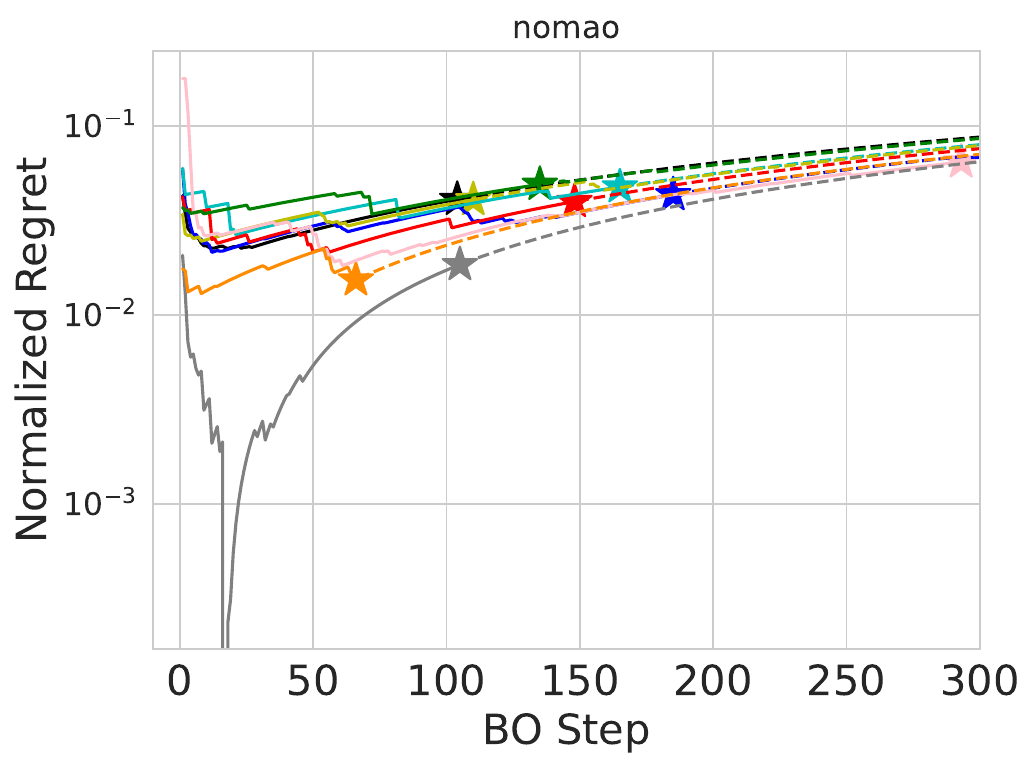}
\includegraphics[width=0.32\textwidth]{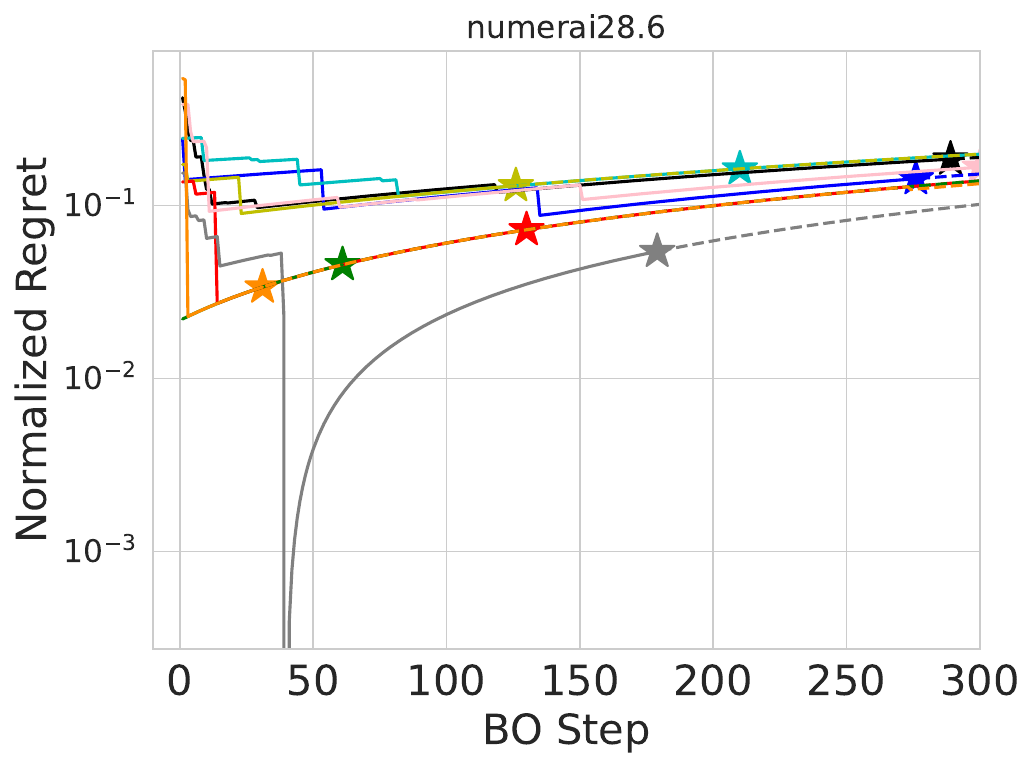}
\includegraphics[width=0.32\textwidth]{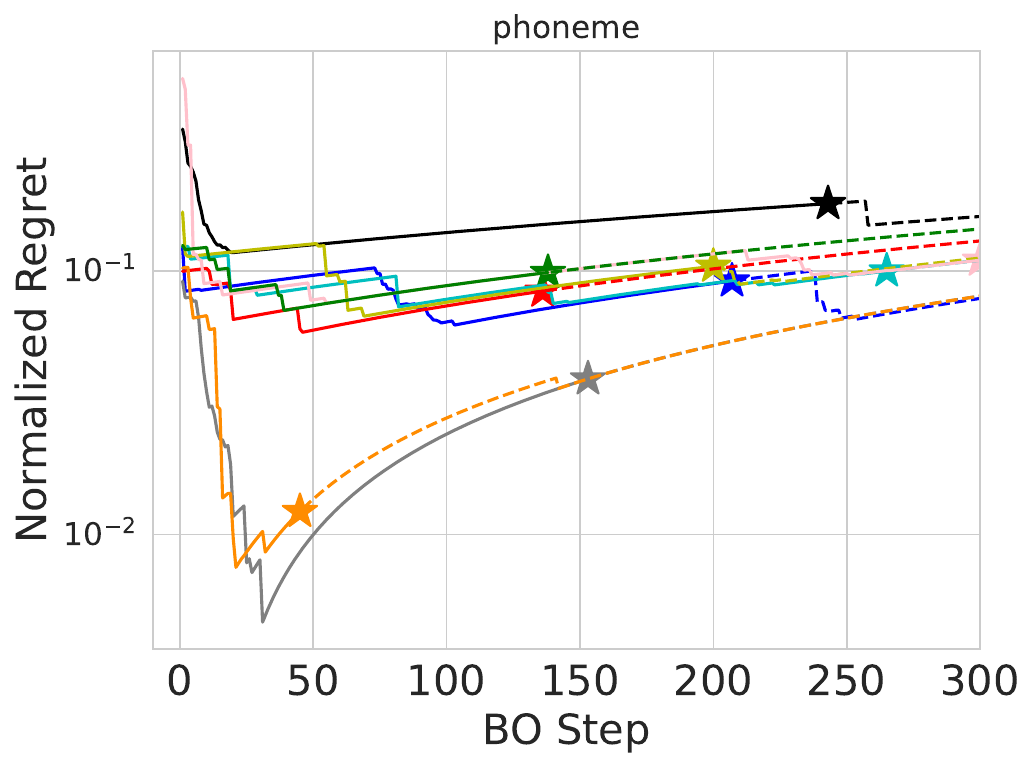}
\includegraphics[width=0.32\textwidth]{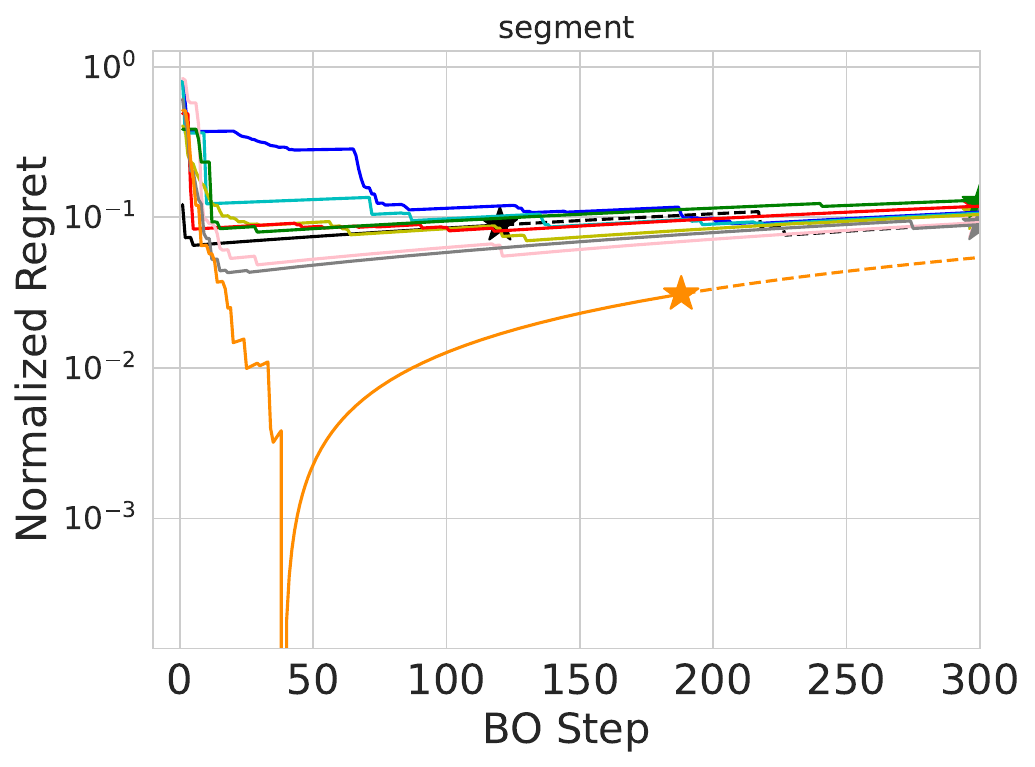}
\includegraphics[width=0.32\textwidth]{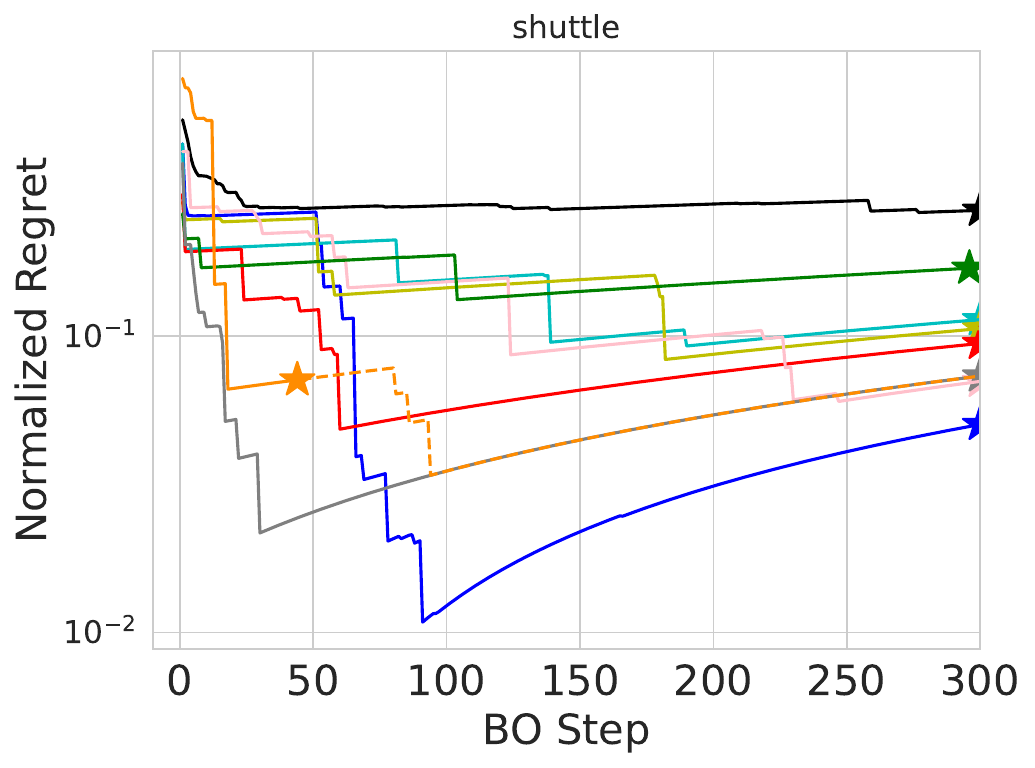}
\includegraphics[width=0.32\textwidth]{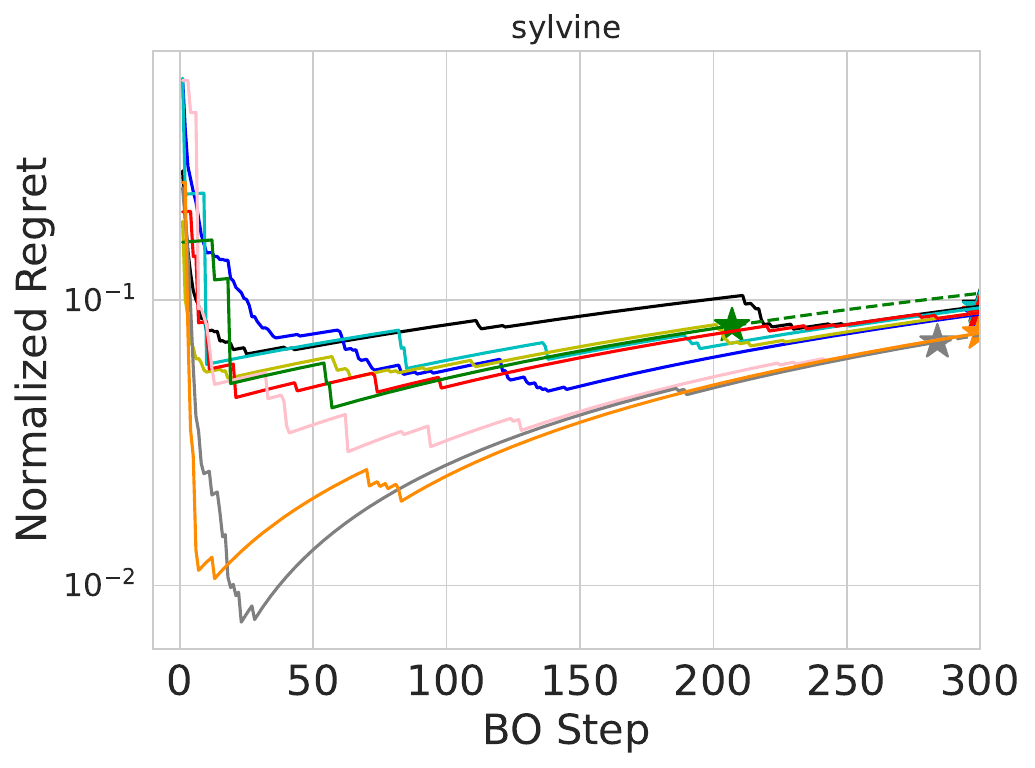}
\includegraphics[width=0.32\textwidth]{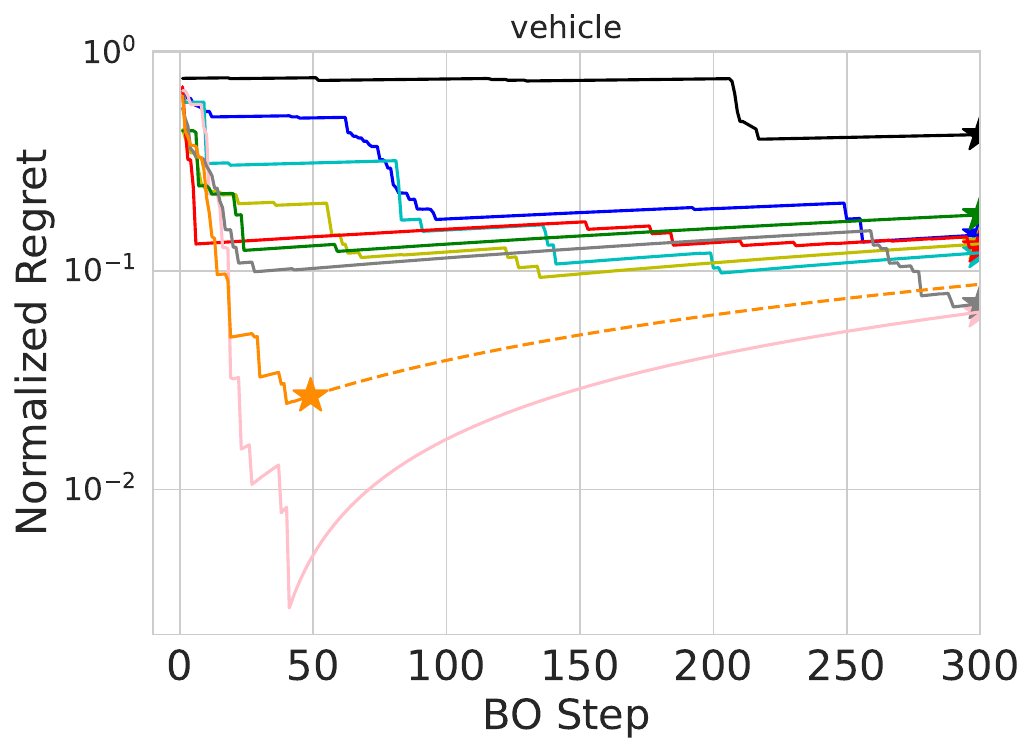}
\includegraphics[width=0.32\textwidth]{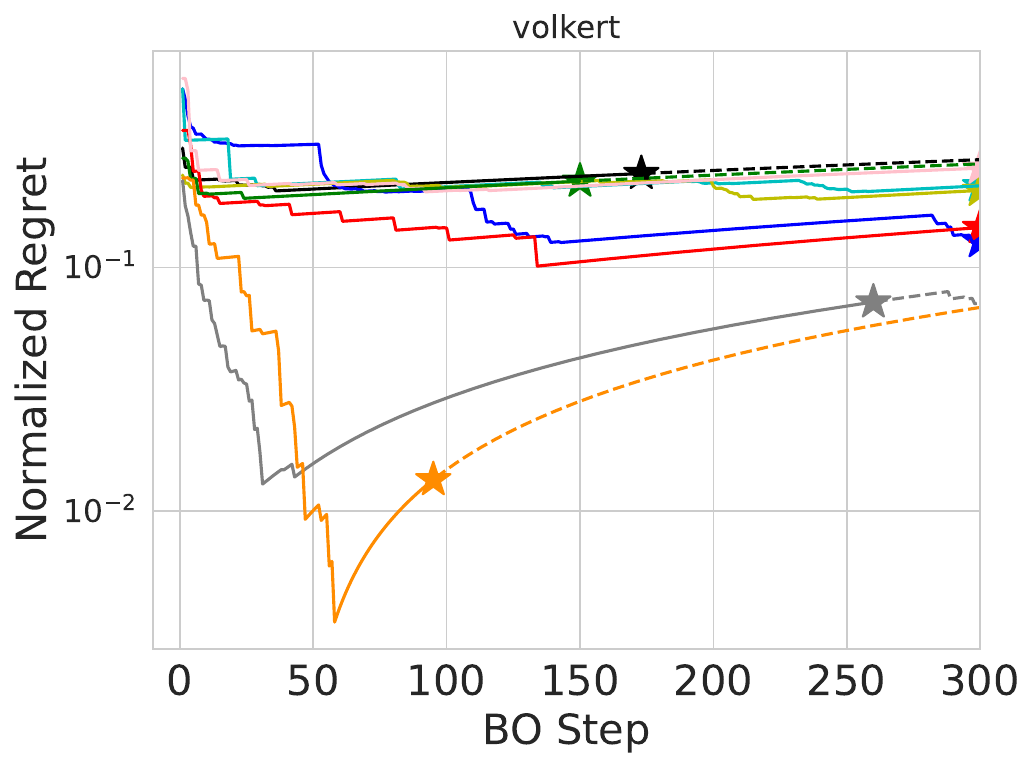}
\vspace{-0.17in}
\medskip
\includegraphics[width=1.0\textwidth]{figures/legend.pdf}
\caption{\small Visualization of the normalized regret over BO steps on \textbf{LCBench ($\alpha=$2e-04)}.}
\label{fig:0.0002_lcbench}
\vspace{-0.15in}
\end{figure}
\begin{figure}[H]
\vspace{-0.15in}
\centering
\includegraphics[width=0.32\textwidth]{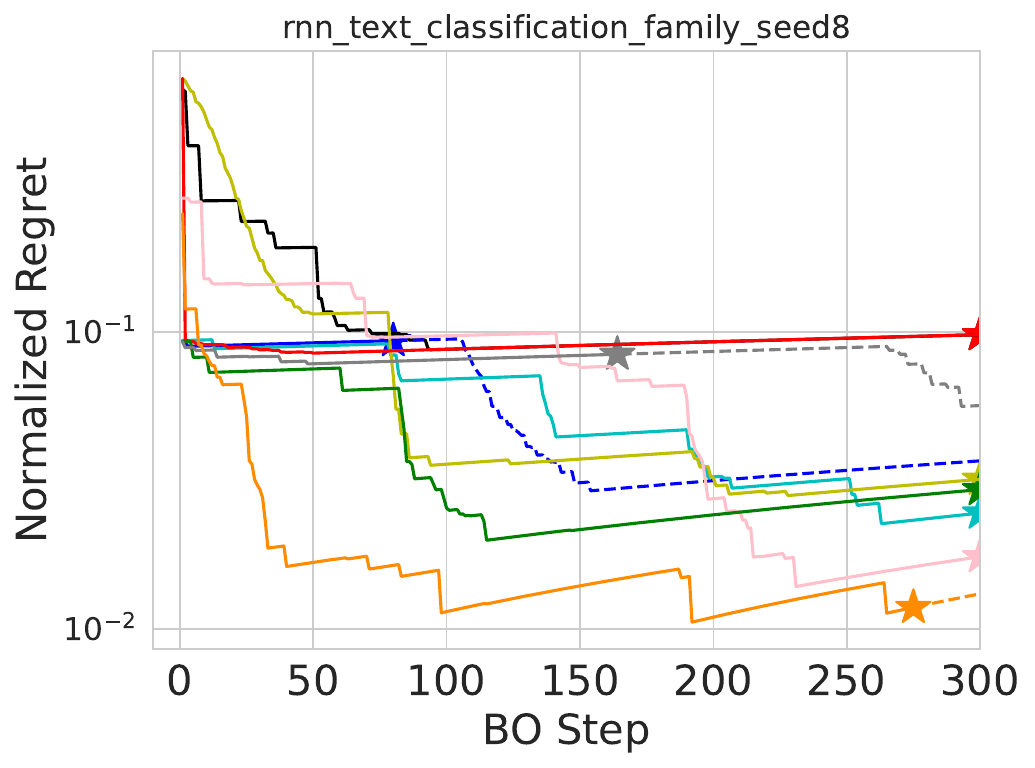}
\includegraphics[width=0.32\textwidth]{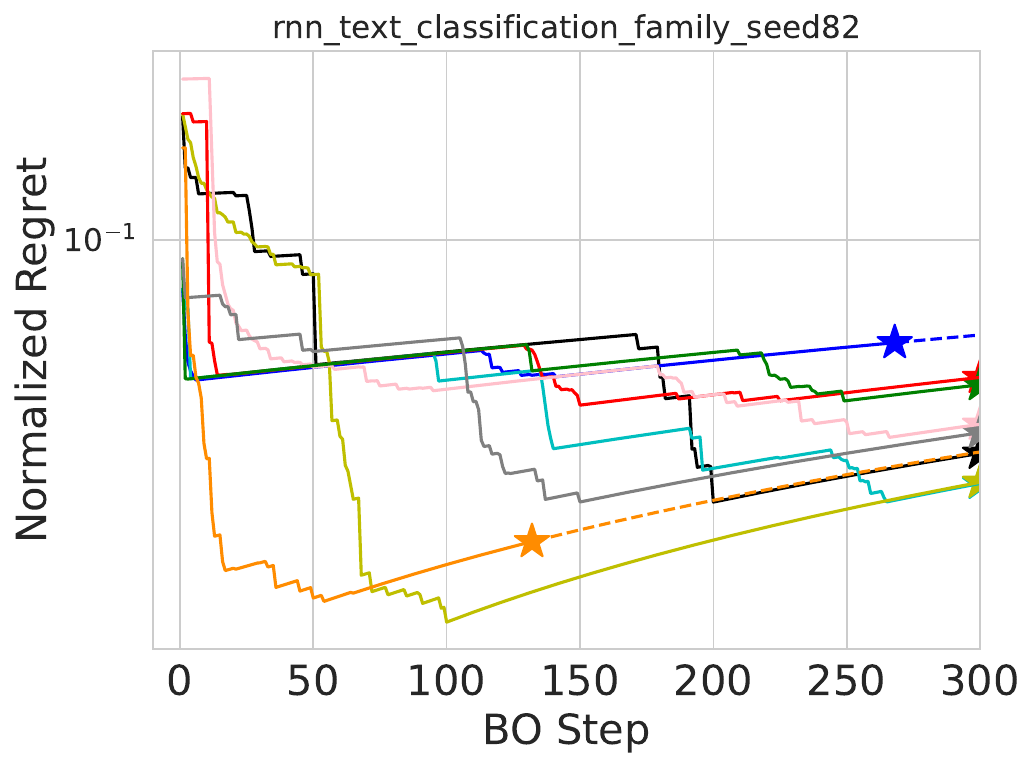}
\includegraphics[width=0.32\textwidth]{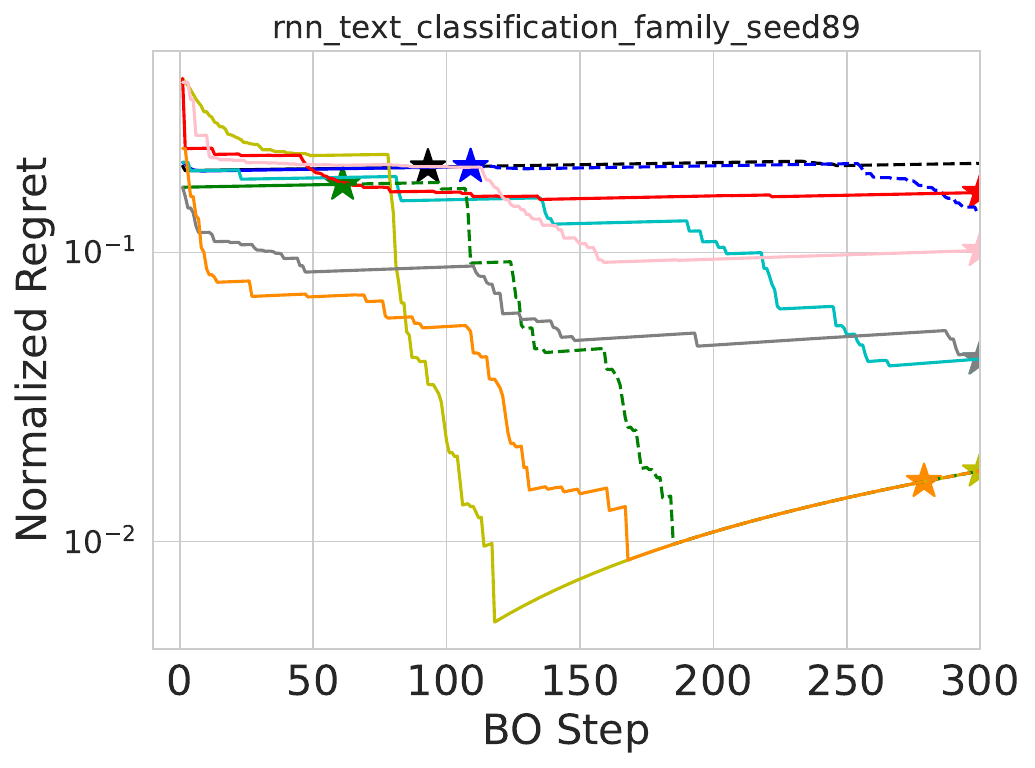}
\includegraphics[width=0.32\textwidth]{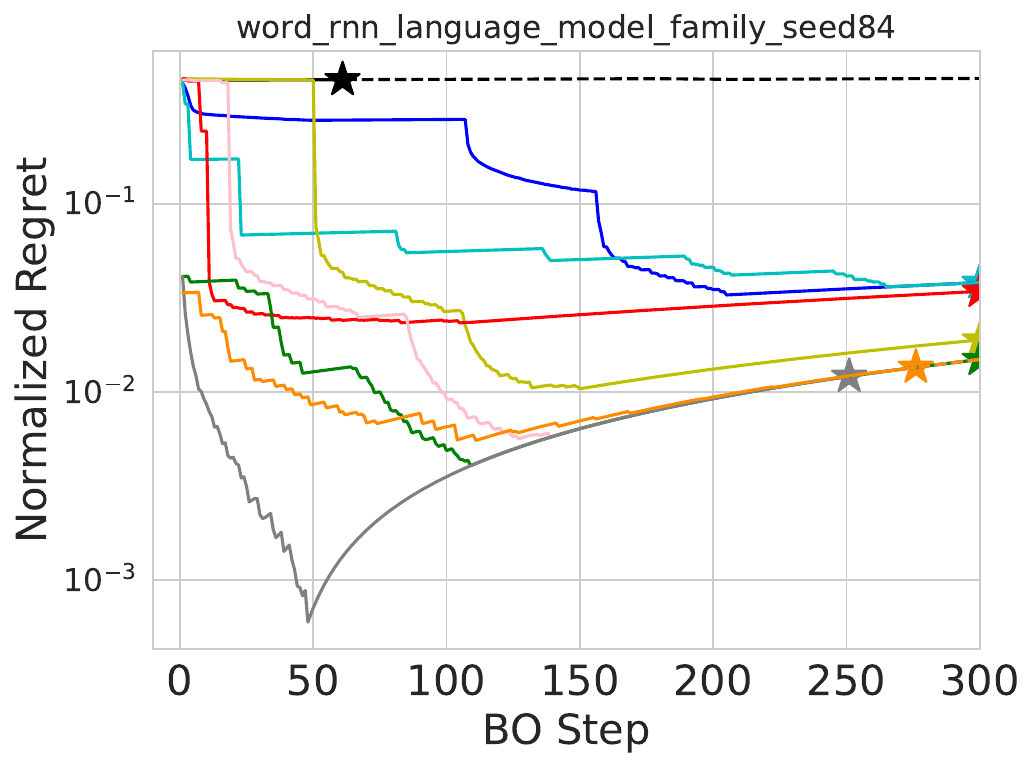}
\includegraphics[width=0.32\textwidth]{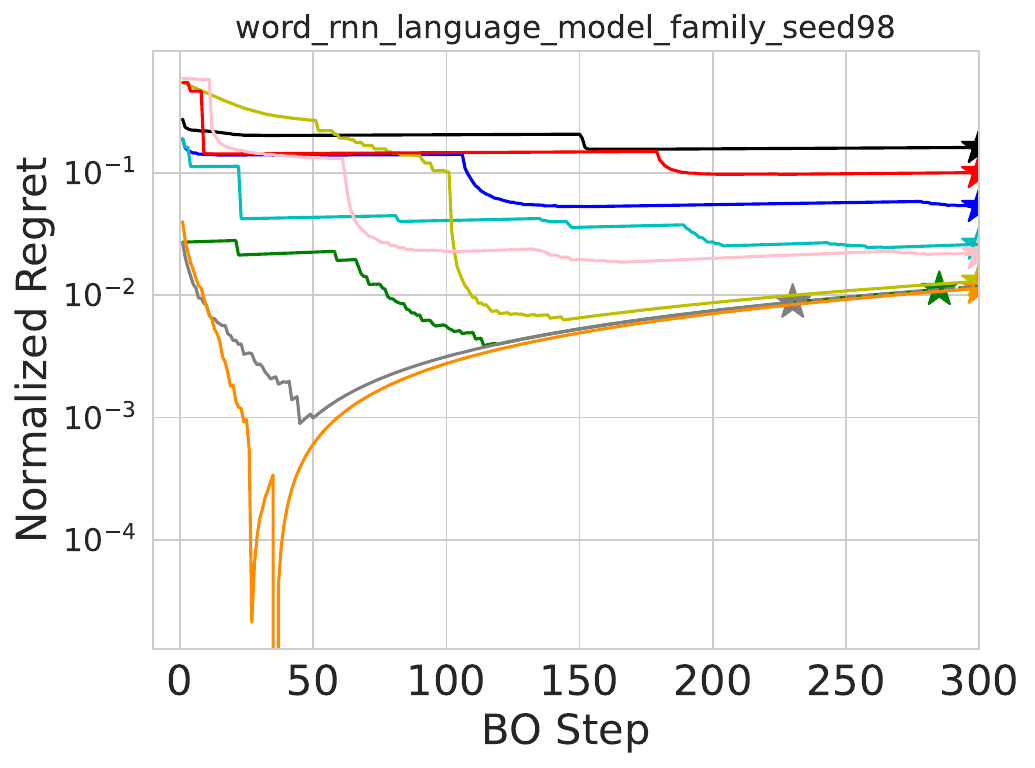}
\includegraphics[width=0.32\textwidth]{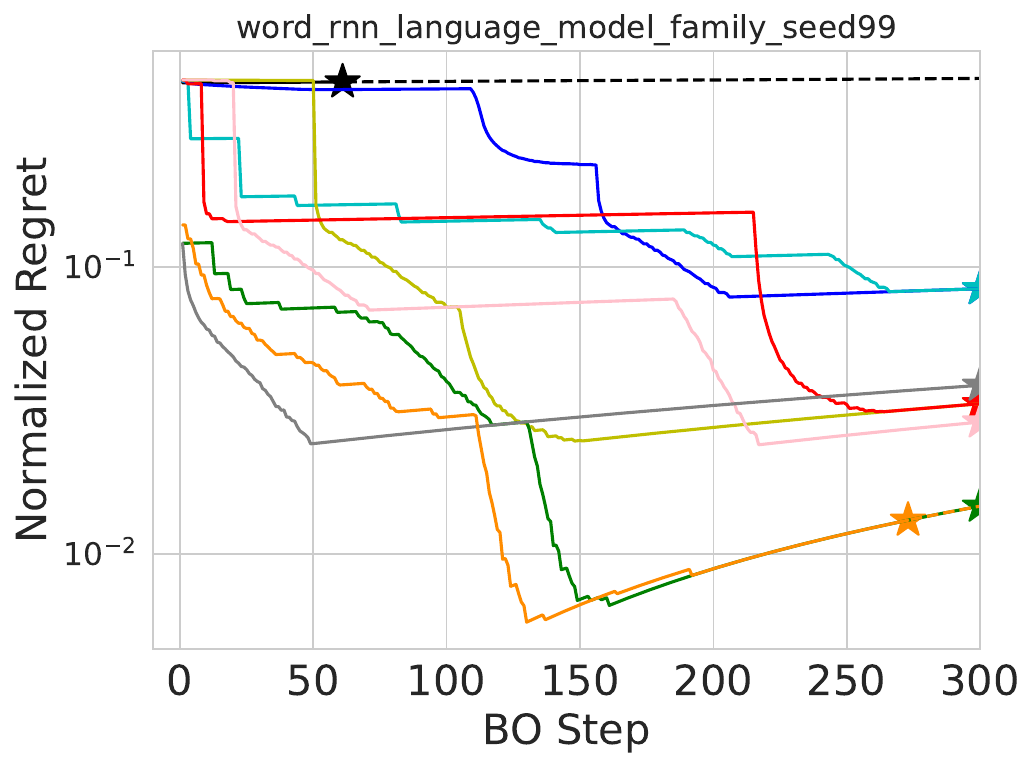}
\includegraphics[width=0.32\textwidth]{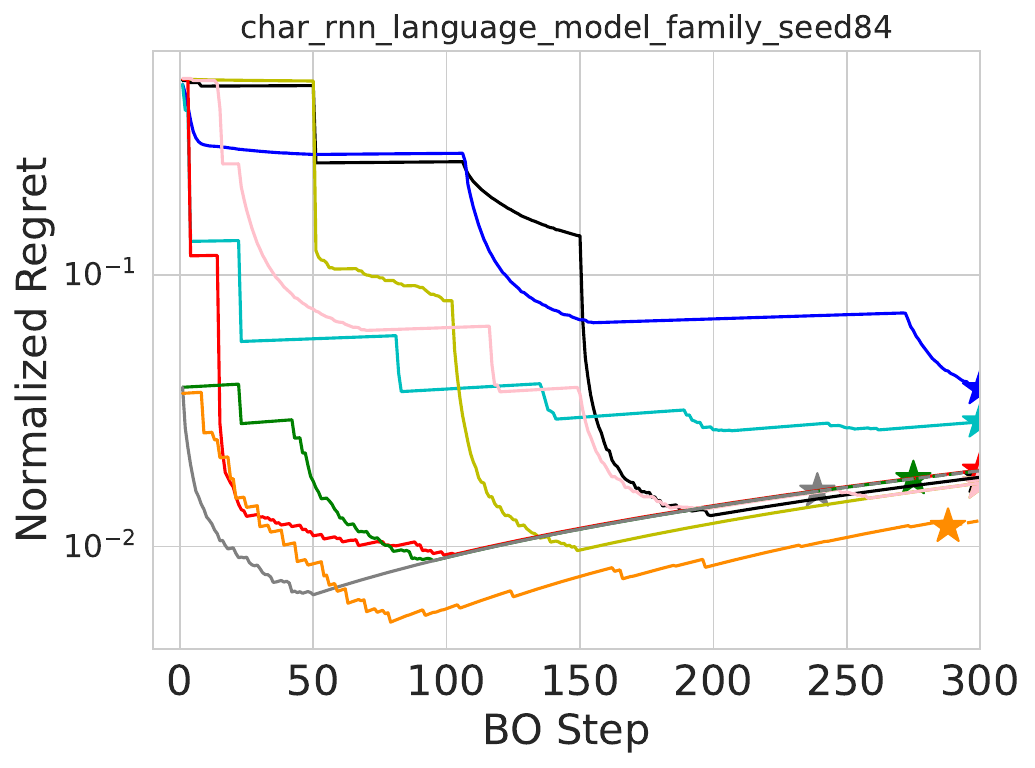}
\includegraphics[width=0.32\textwidth]{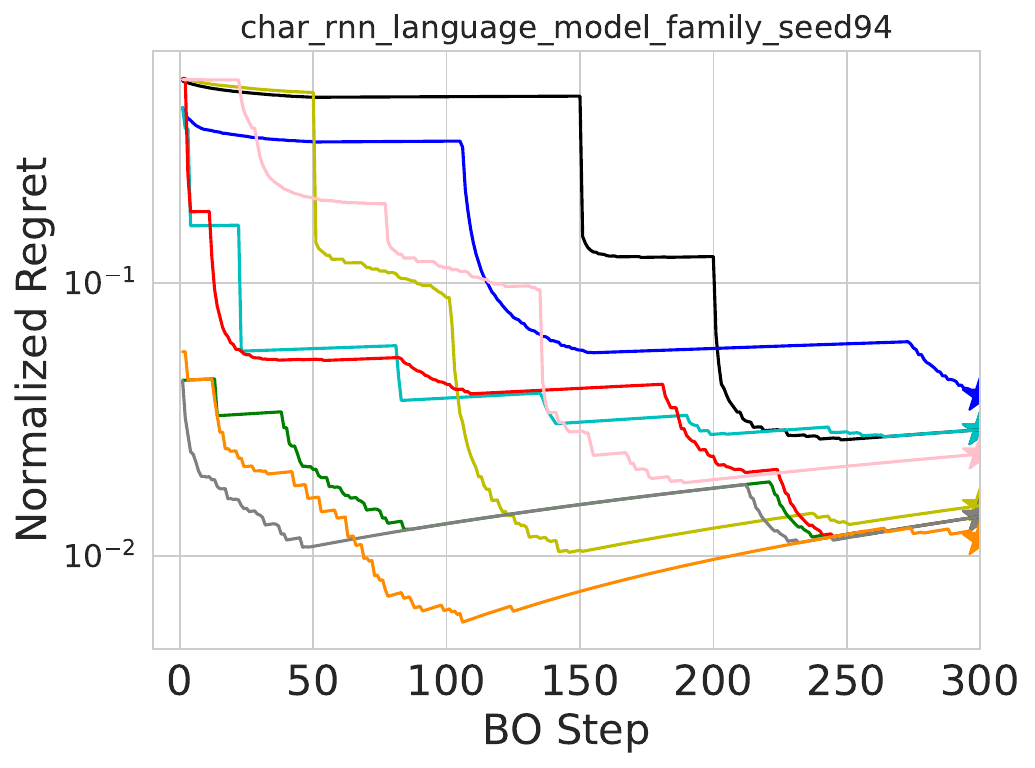}
\includegraphics[width=0.32\textwidth]{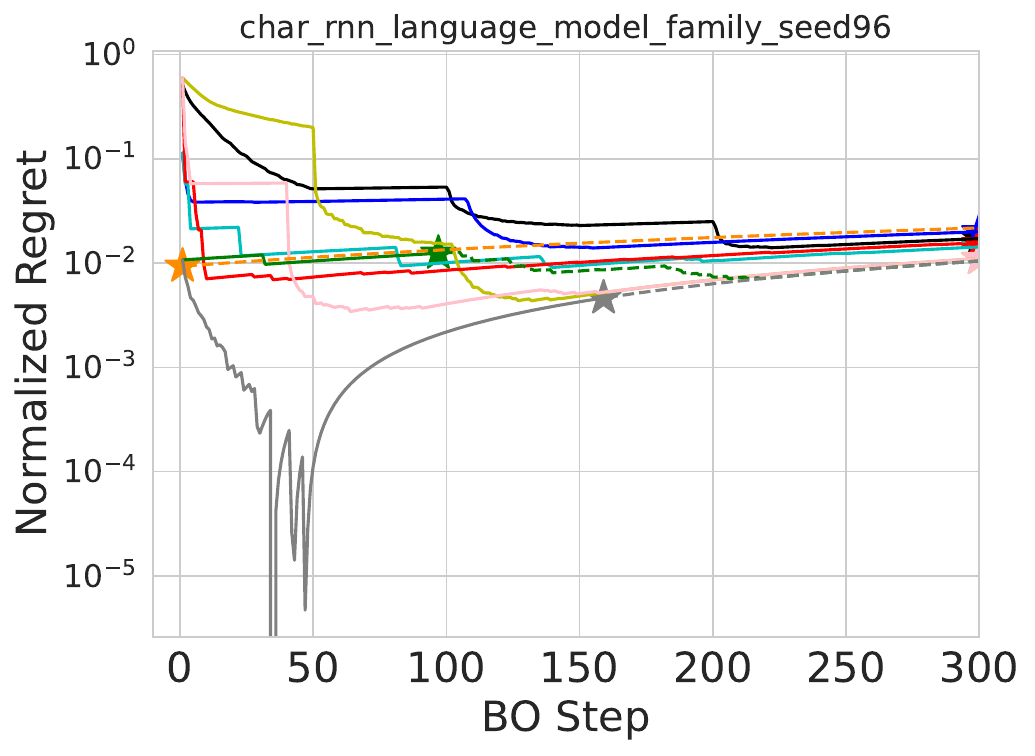}
\vspace{-0.17in}
\medskip
\includegraphics[width=1.0\textwidth]{figures/legend.pdf}
\caption{\small Visualization of the normalized regret over BO steps on \textbf{TaskSet ($\alpha=$4e-05)}.}
\label{fig:4e-05_taskset}
\vspace{-0.15in}
\end{figure}
\begin{figure}[H]
\vspace{-0.15in}
\centering
\includegraphics[width=0.32\textwidth]{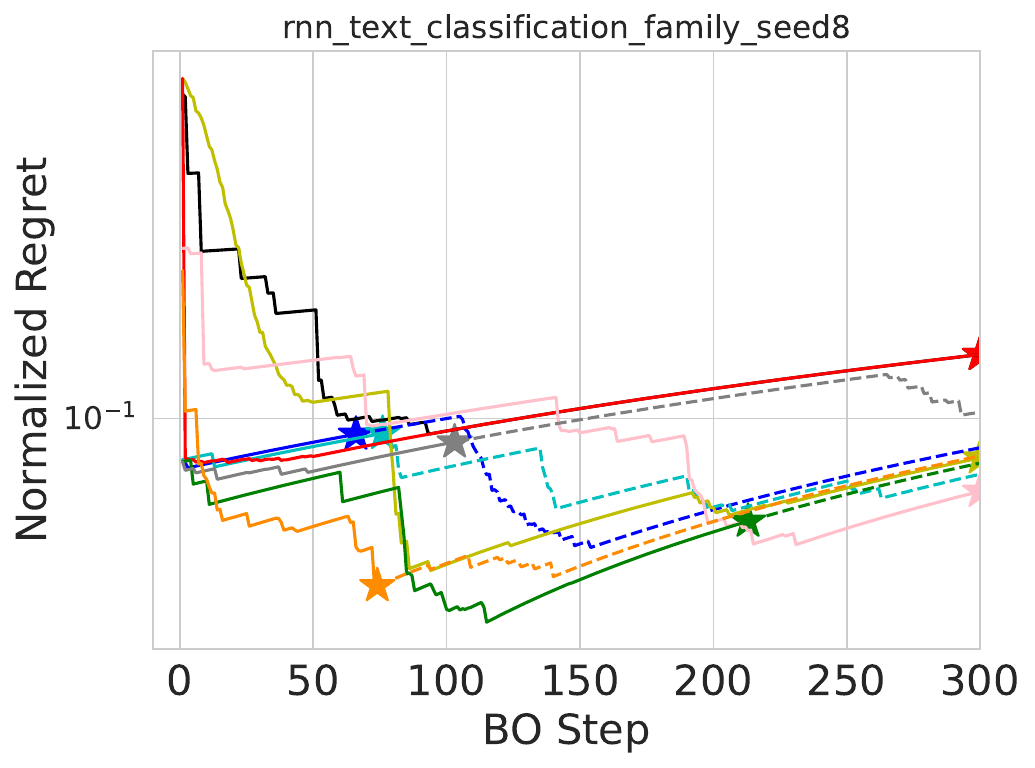}
\includegraphics[width=0.32\textwidth]{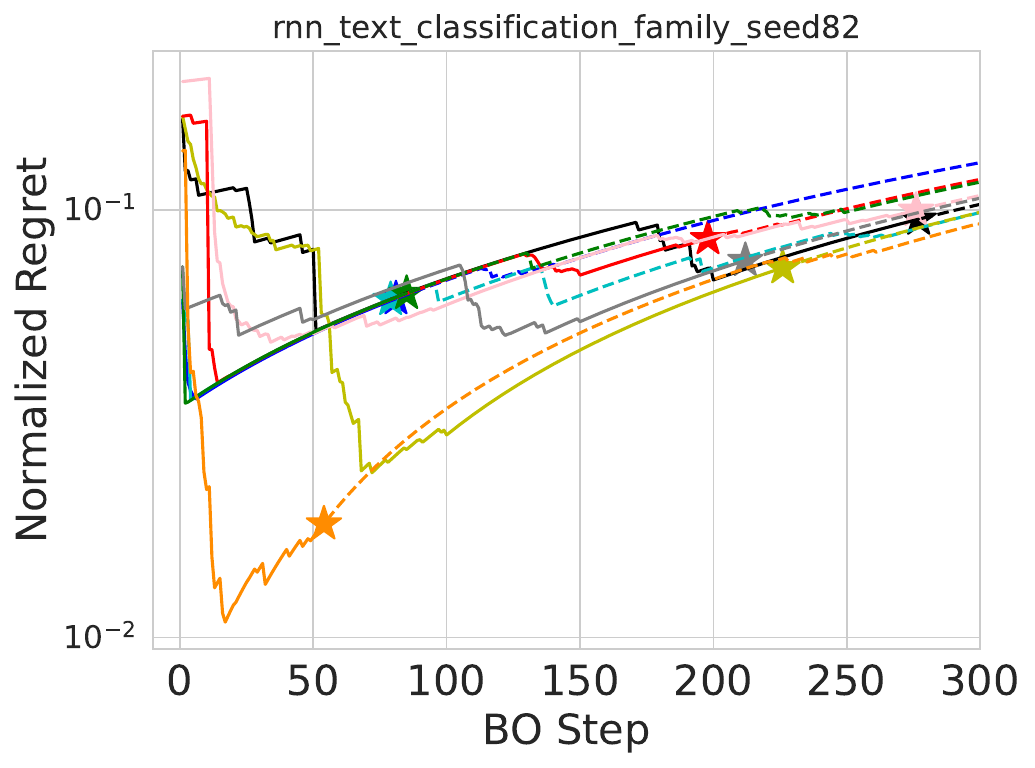}
\includegraphics[width=0.32\textwidth]{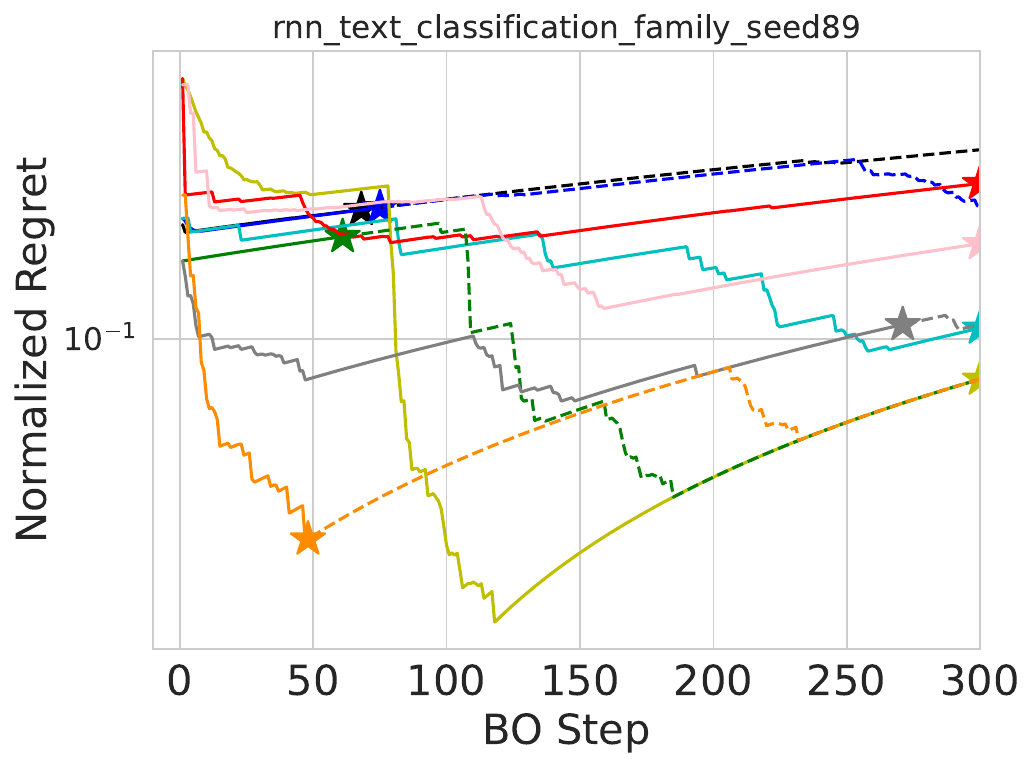}
\includegraphics[width=0.32\textwidth]{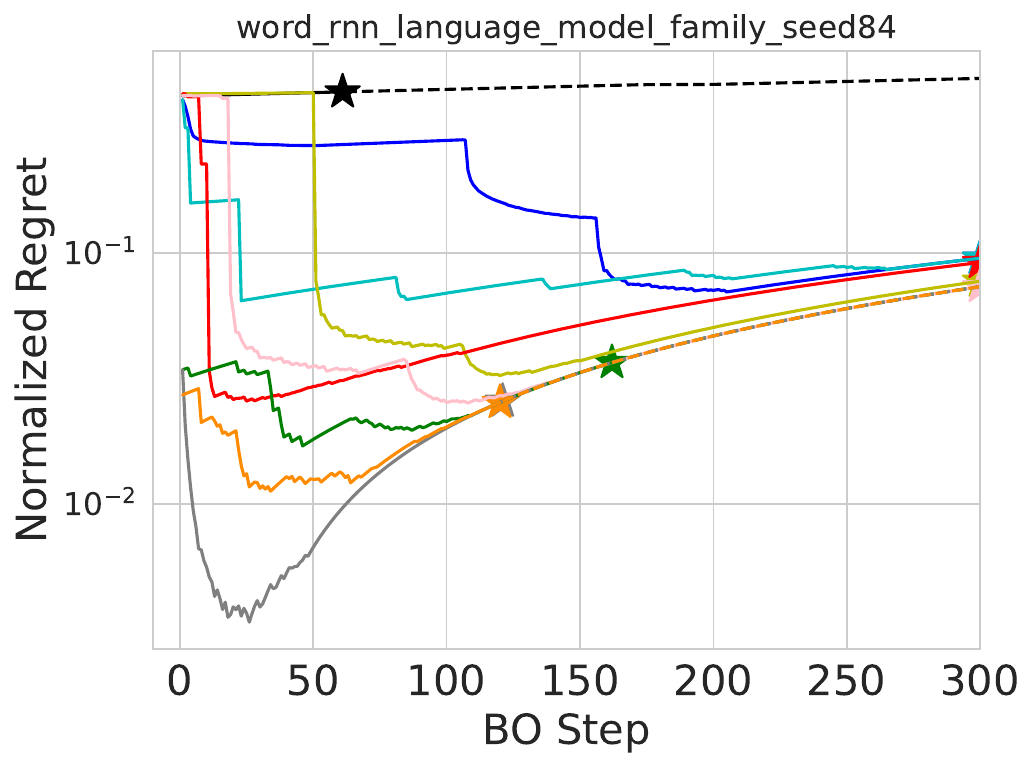}
\includegraphics[width=0.32\textwidth]{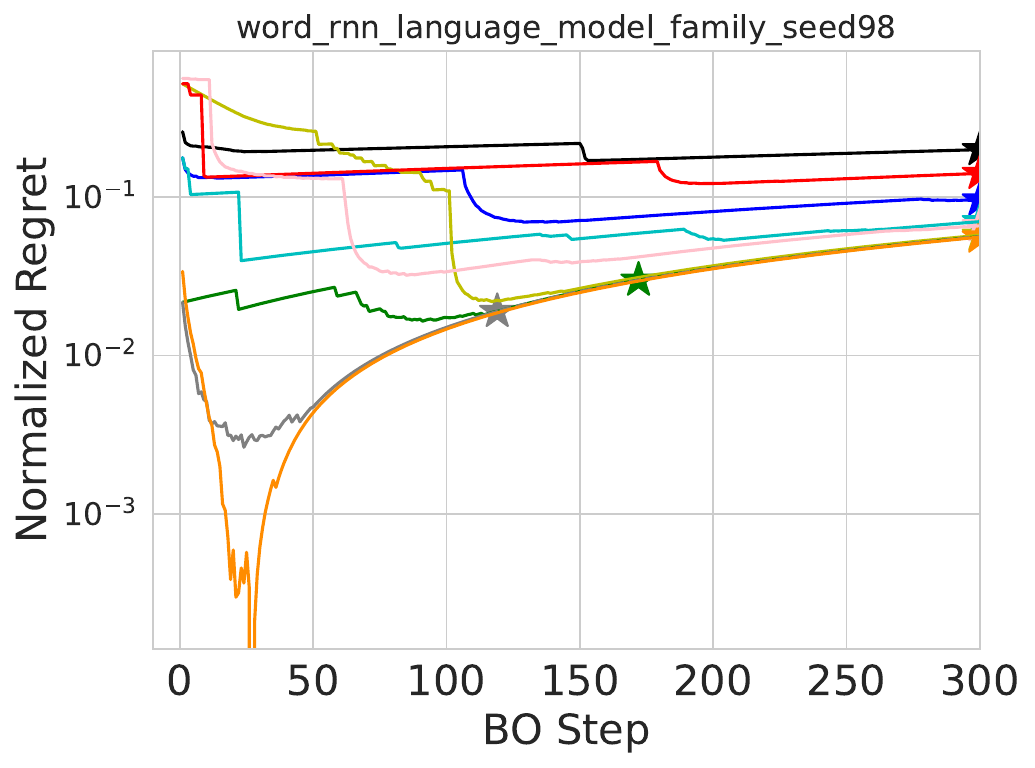}
\includegraphics[width=0.32\textwidth]{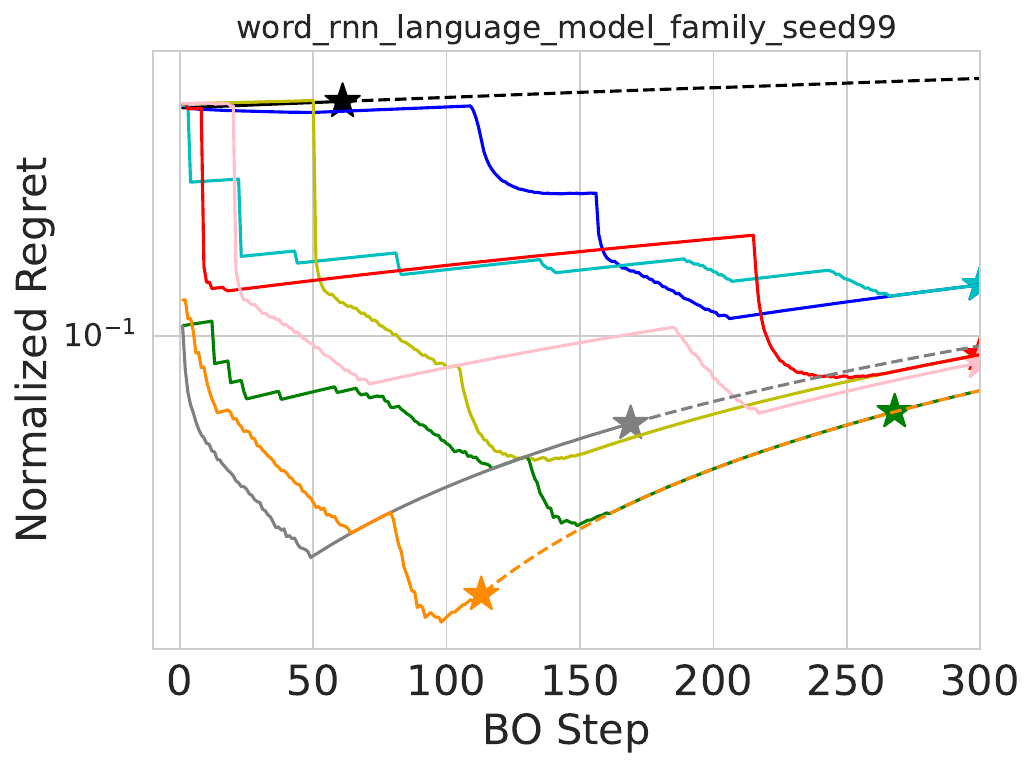}
\includegraphics[width=0.32\textwidth]{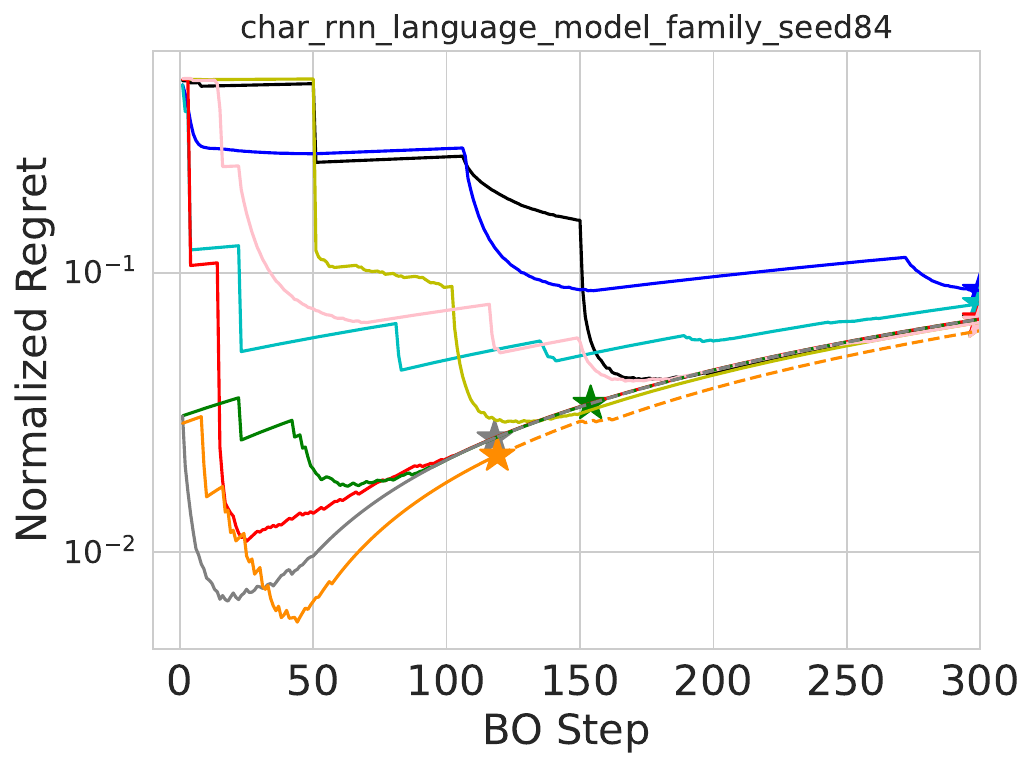}
\includegraphics[width=0.32\textwidth]{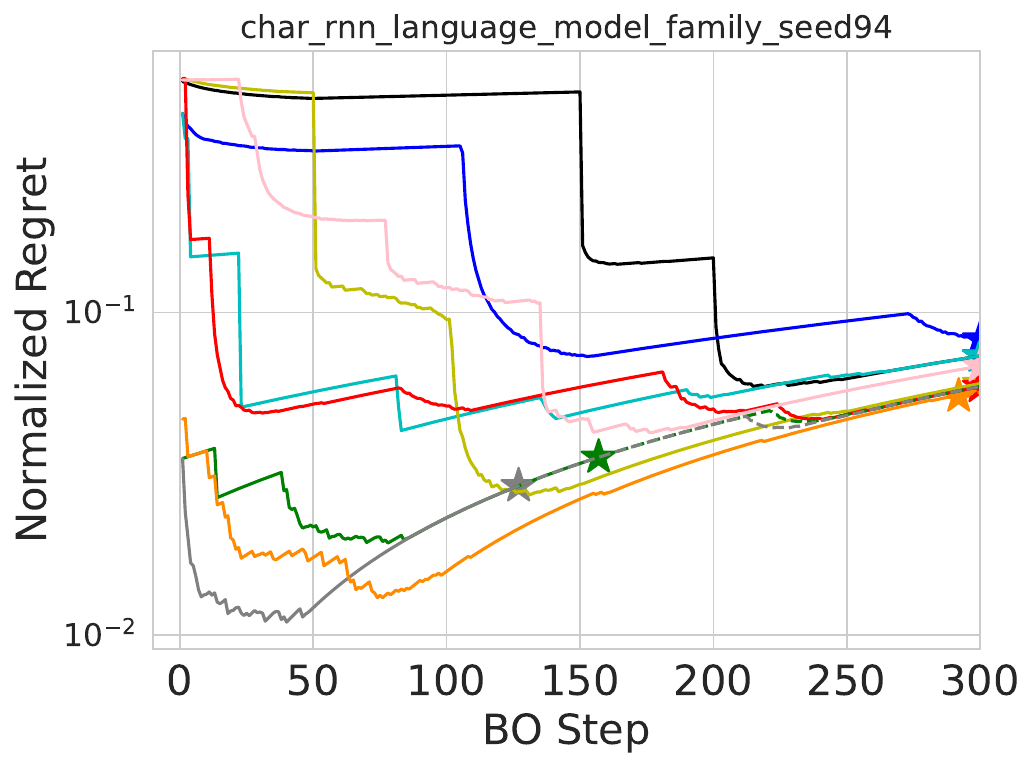}
\includegraphics[width=0.32\textwidth]{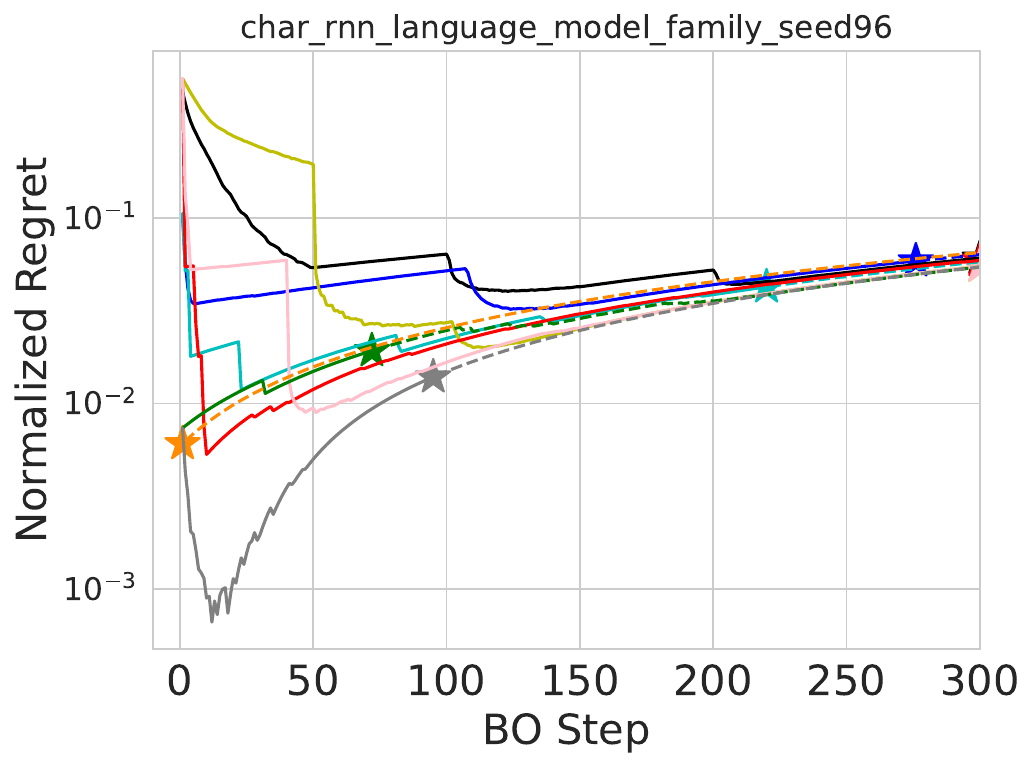}
\vspace{-0.17in}
\medskip
\includegraphics[width=1.0\textwidth]{figures/legend.pdf}
\caption{\small Visualization of the normalized regret over BO steps on \textbf{TaskSet ($\alpha=$2e-04)}.}
\label{fig:0.0002_taskset}
\vspace{-0.15in}
\end{figure}
\begin{figure}[H]
\vspace{-0.15in}
\centering
\includegraphics[width=0.32\textwidth]{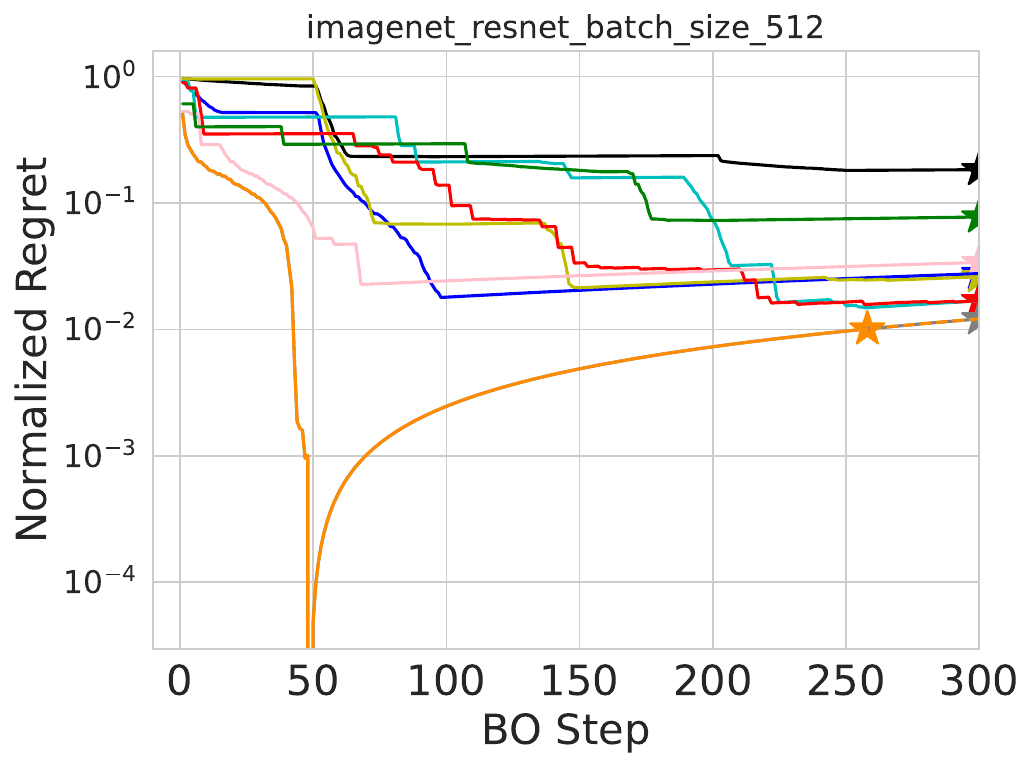}
\includegraphics[width=0.32\textwidth]{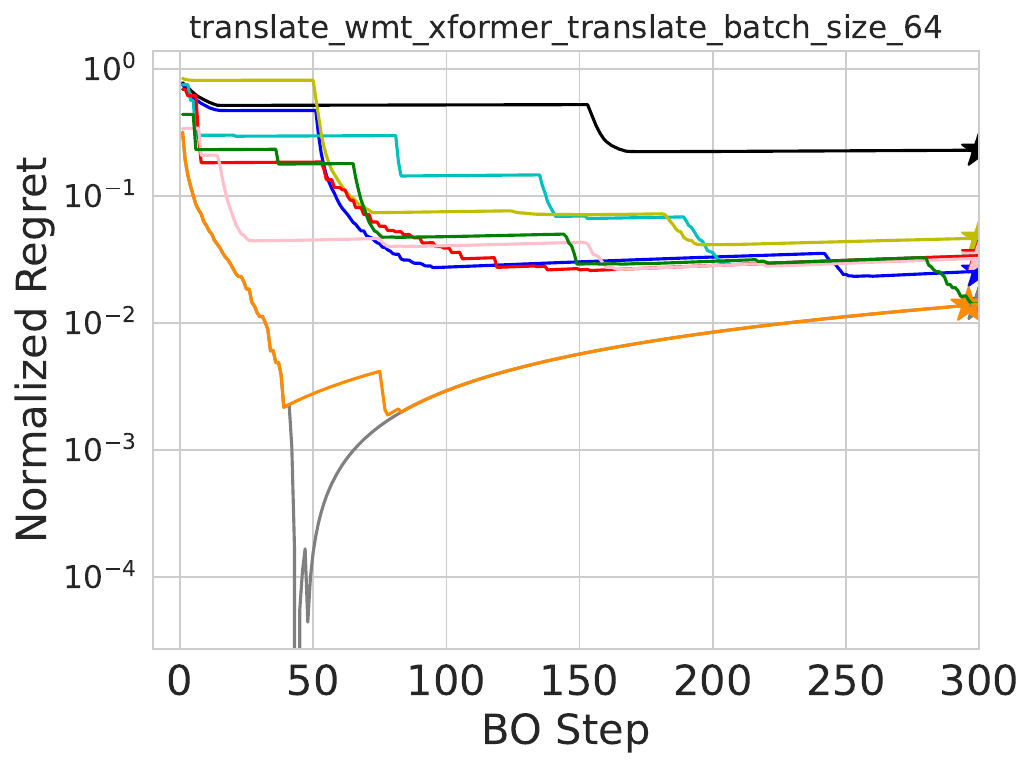}
\includegraphics[width=0.32\textwidth]{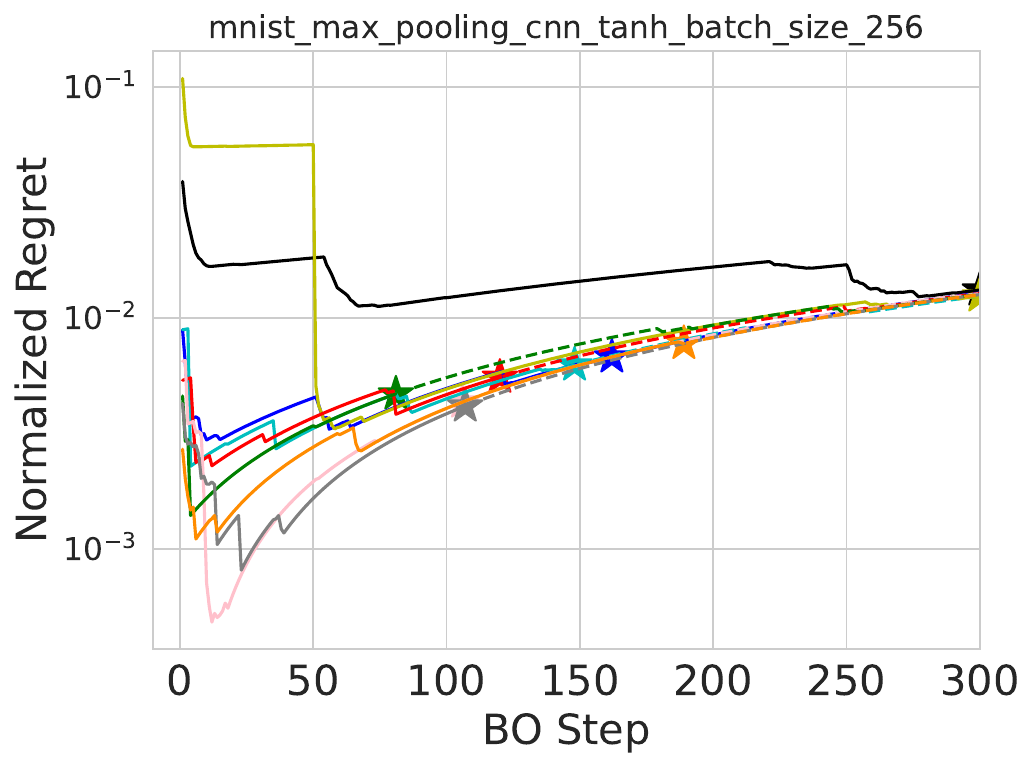}
\includegraphics[width=0.32\textwidth]{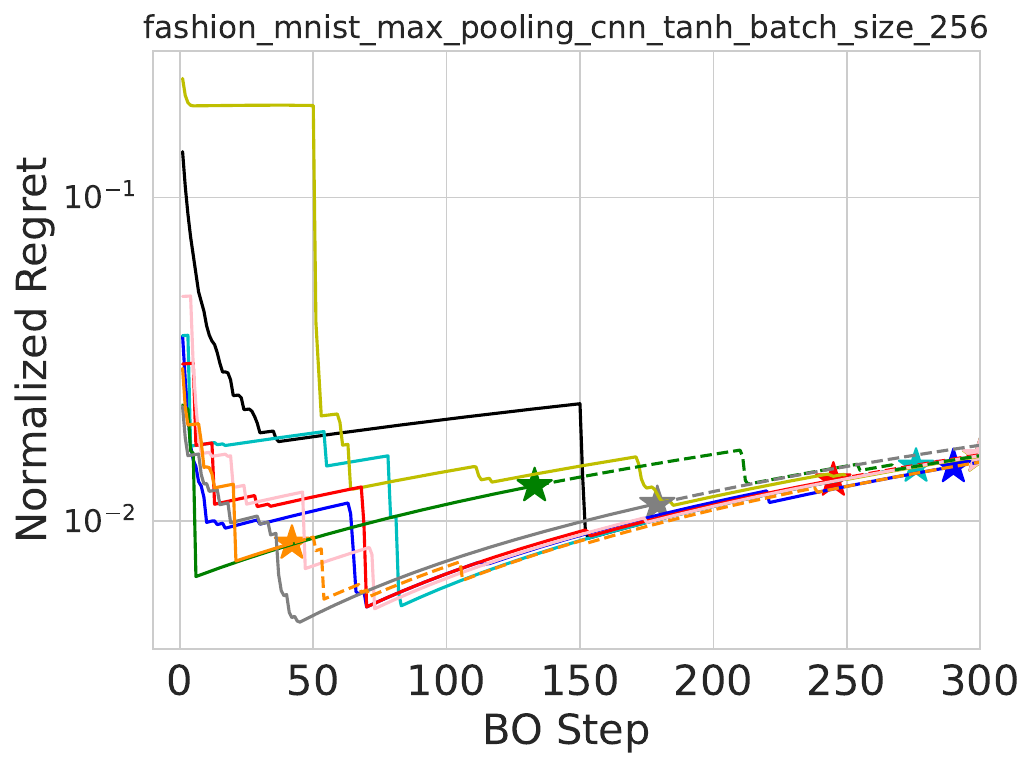}
\includegraphics[width=0.32\textwidth]{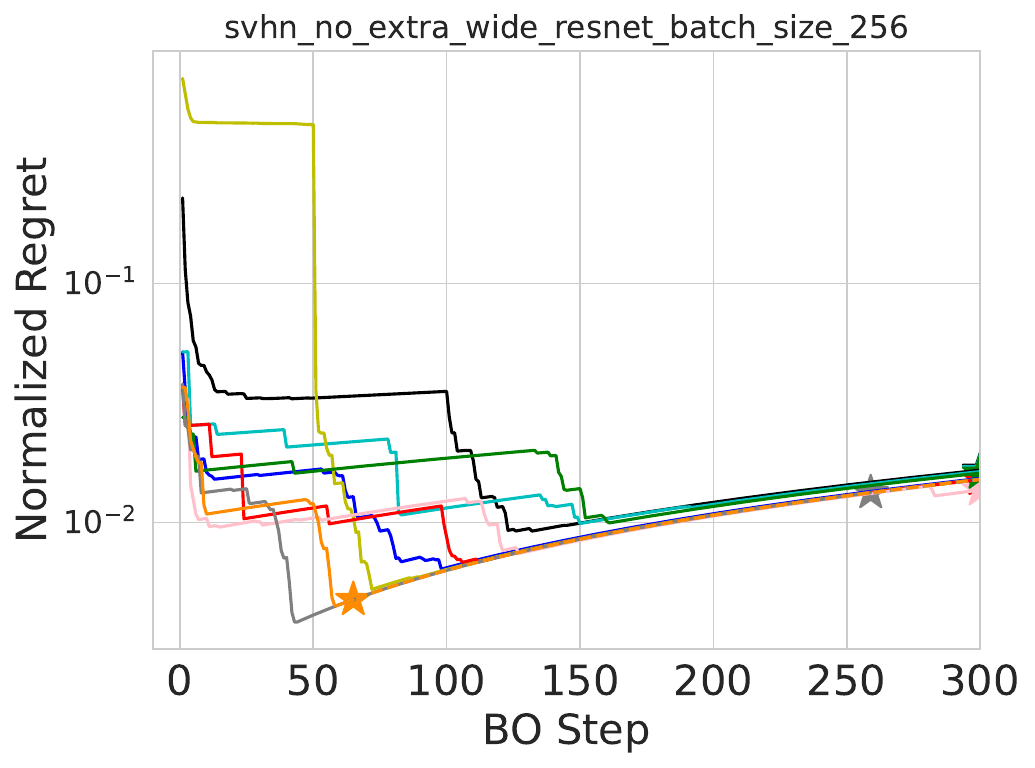}
\includegraphics[width=0.32\textwidth]{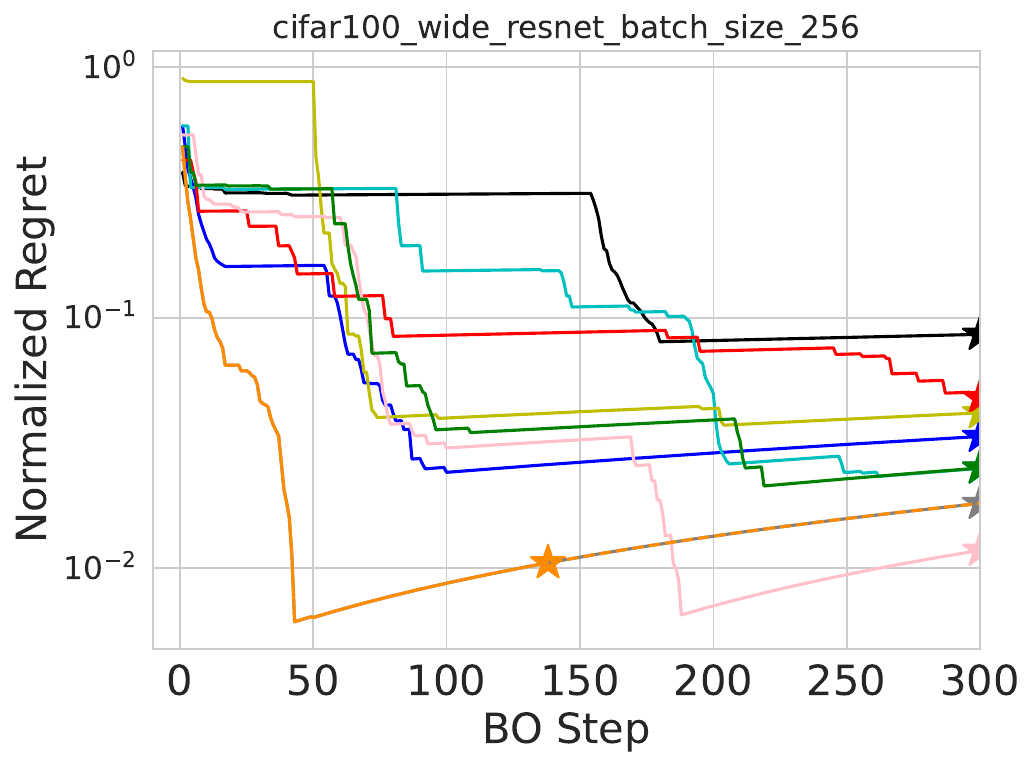}
\includegraphics[width=0.32\textwidth]{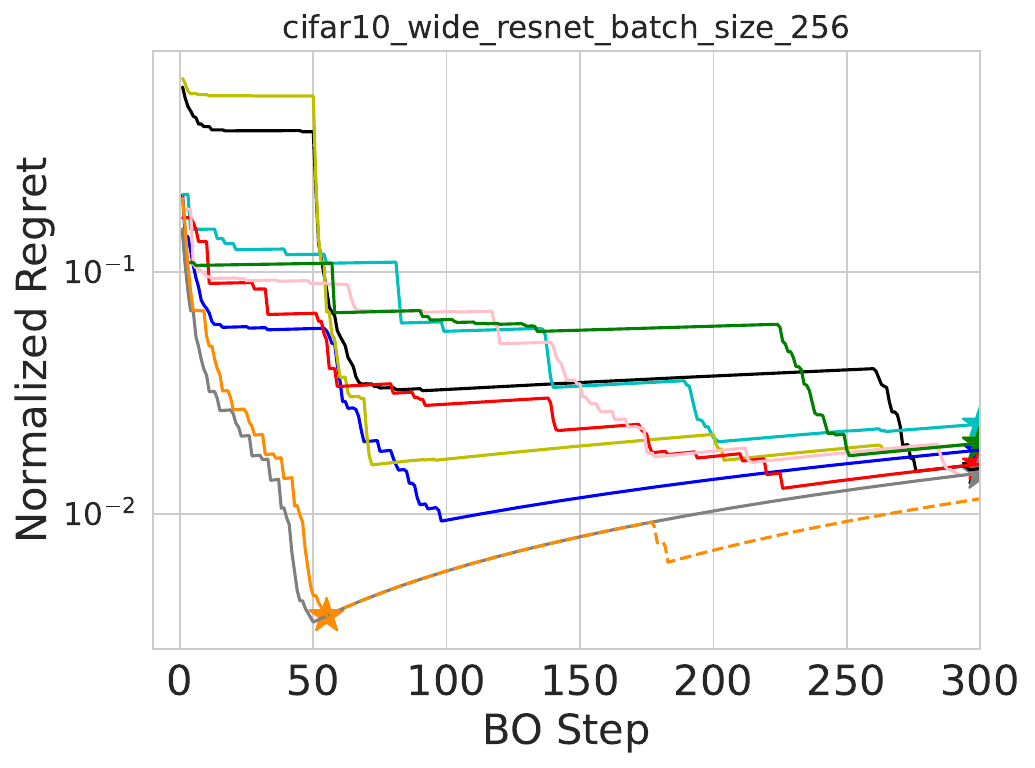}
\vspace{-0.17in}
\medskip
\includegraphics[width=1.0\textwidth]{figures/legend.pdf}
\caption{\small Visualization of the normalized regret over BO steps on \textbf{PD1 ($\alpha=$4e-05)}.}
\label{fig:4e-05_pd1}
\vspace{-0.15in}
\end{figure}
\begin{figure}[H]
\vspace{-0.15in}
\centering
\includegraphics[width=0.32\textwidth]{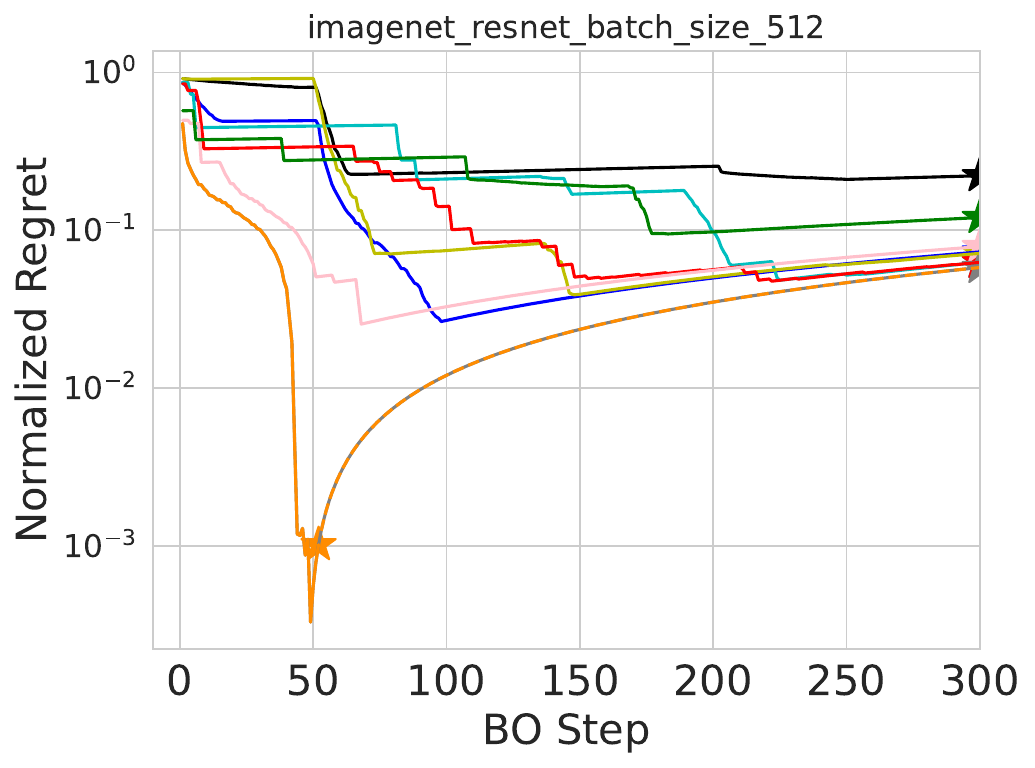}
\includegraphics[width=0.32\textwidth]{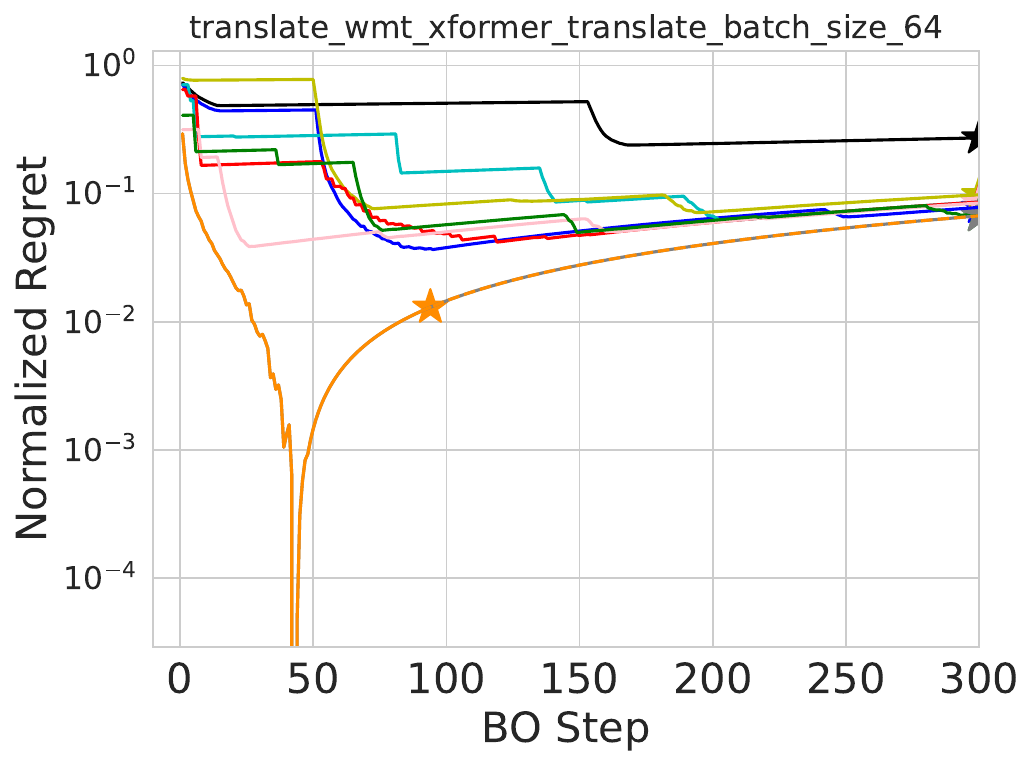}
\includegraphics[width=0.32\textwidth]{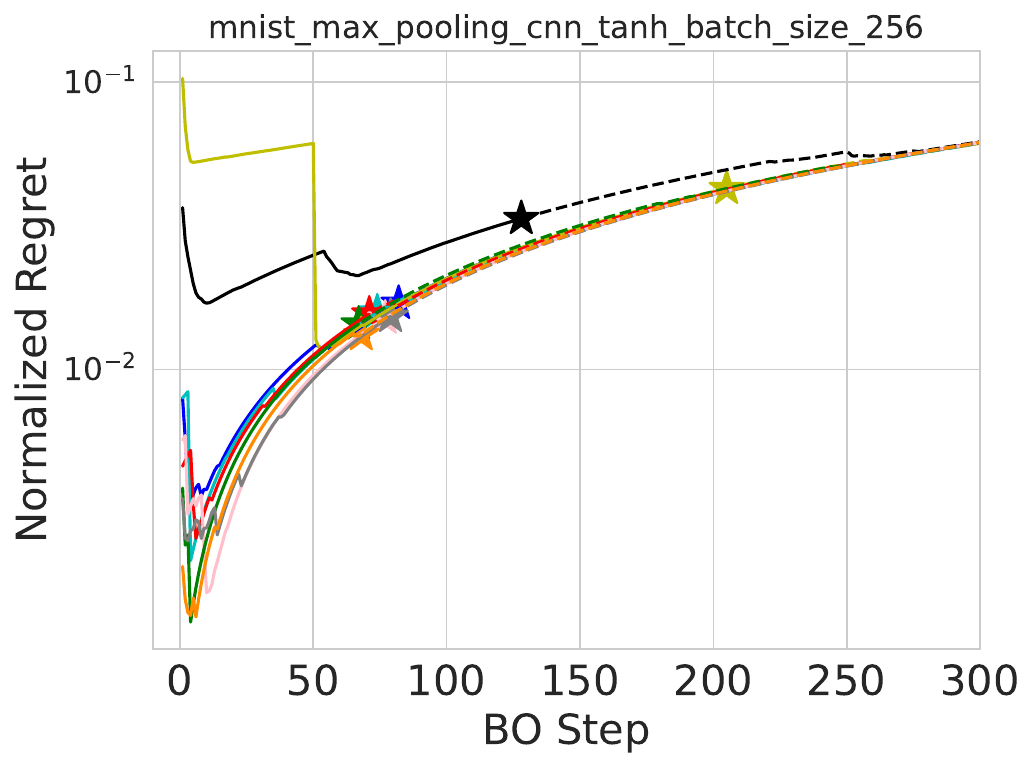}
\includegraphics[width=0.32\textwidth]{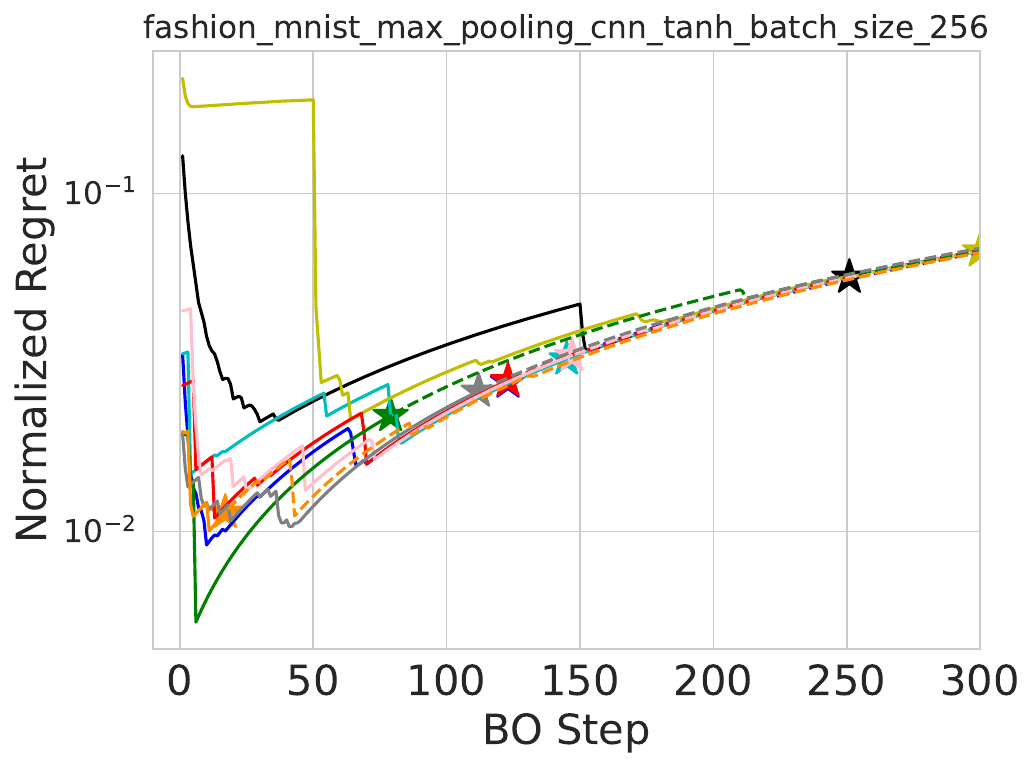}
\includegraphics[width=0.32\textwidth]{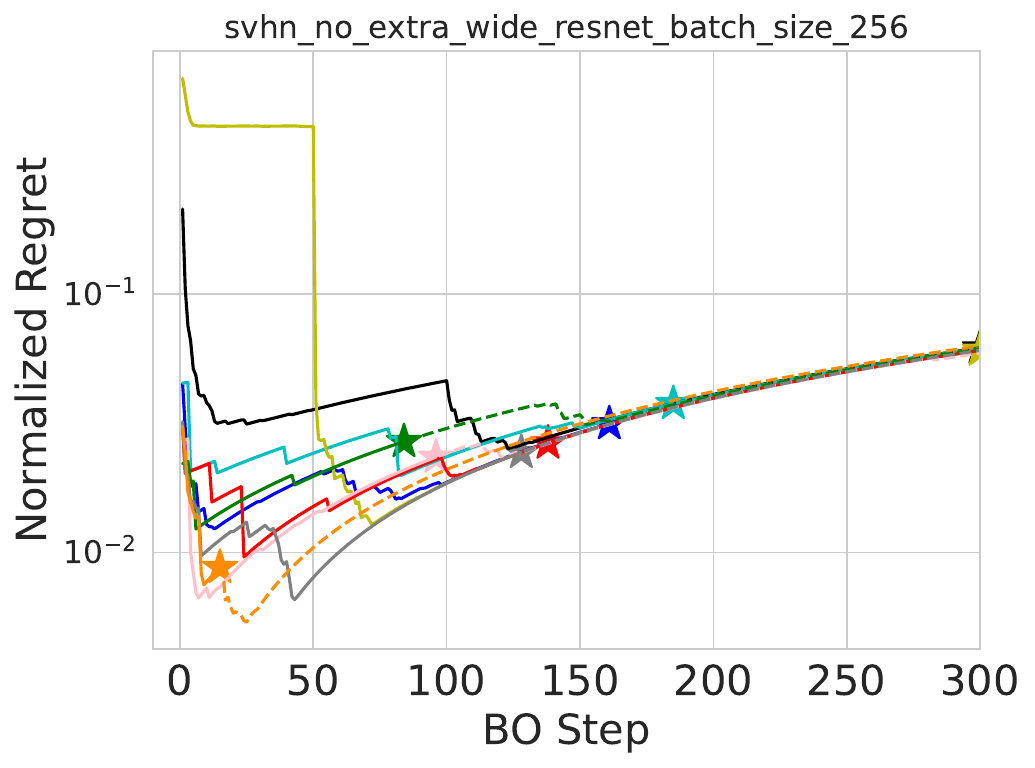}
\includegraphics[width=0.32\textwidth]{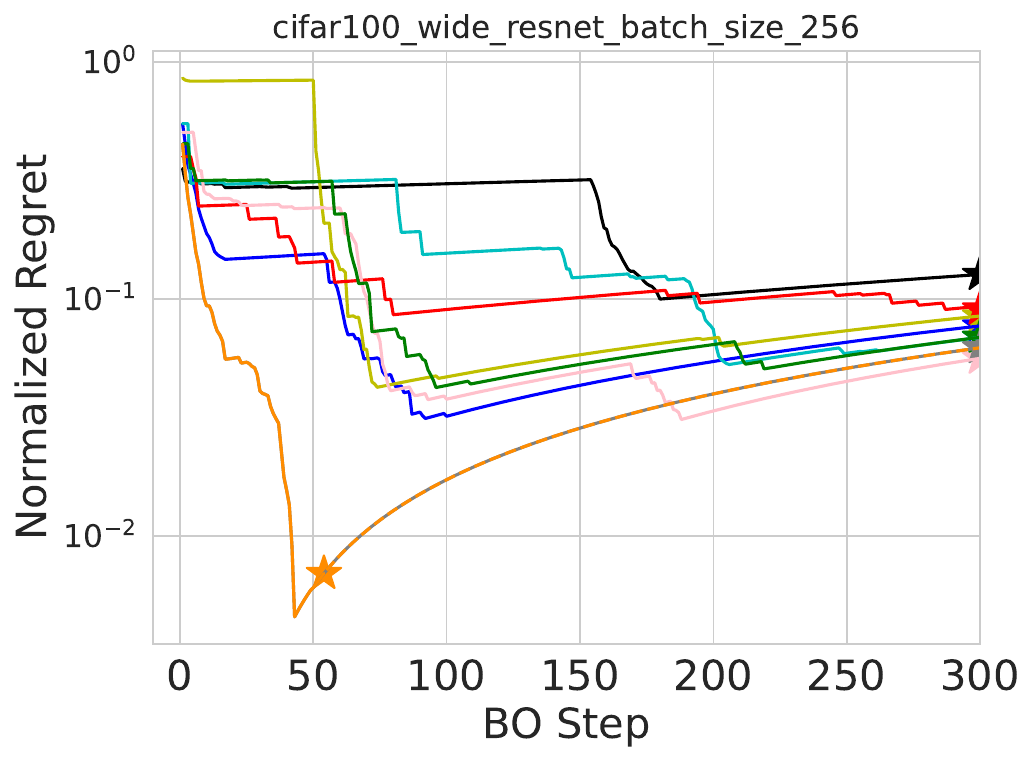}
\includegraphics[width=0.32\textwidth]{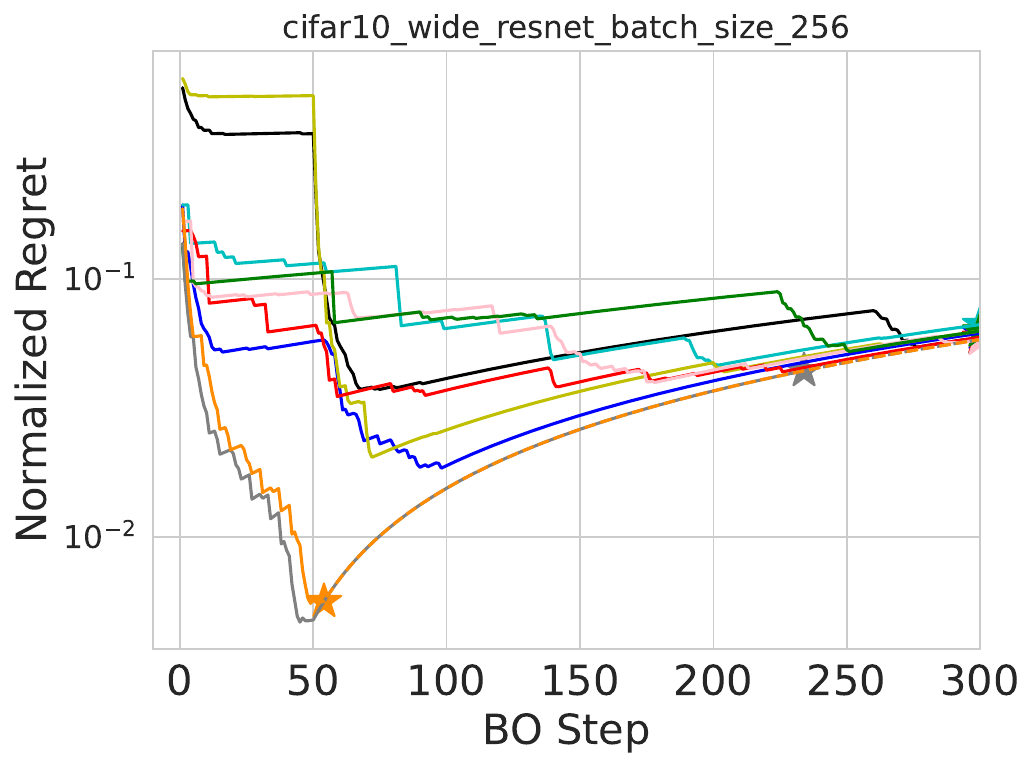}
\vspace{-0.17in}
\medskip
\includegraphics[width=1.0\textwidth]{figures/legend.pdf}
\caption{\small Visualization of the normalized regret over BO steps on \textbf{PD1 ($\alpha=$2e-04)}.}
\label{fig:0.0002_pd1}
\vspace{-0.15in}
\end{figure}

\paragraph{Visualizations of the normalized regret over BO steps} for LCBench ($\alpha=4e\text{-}05$), LCBench ($\alpha=2e\text{-}04$), TaskSet ($\alpha=4e\text{-}05$), TaskSet ($\alpha=2e\text{-}04$), PD1 ($\alpha=4e\text{-}05$), and PD1 ($\alpha=2e\text{-}04$) are provided Figure~\ref{fig:4e-05_lcbench}, \ref{fig:0.0002_lcbench}, \ref{fig:4e-05_taskset}, \ref{fig:0.0002_taskset}, \ref{fig:4e-05_pd1}, and \ref{fig:0.0002_pd1}, respectively.

\begin{figure}[H]
\vspace{-0.15in}
\centering
\includegraphics[width=1.0\textwidth]{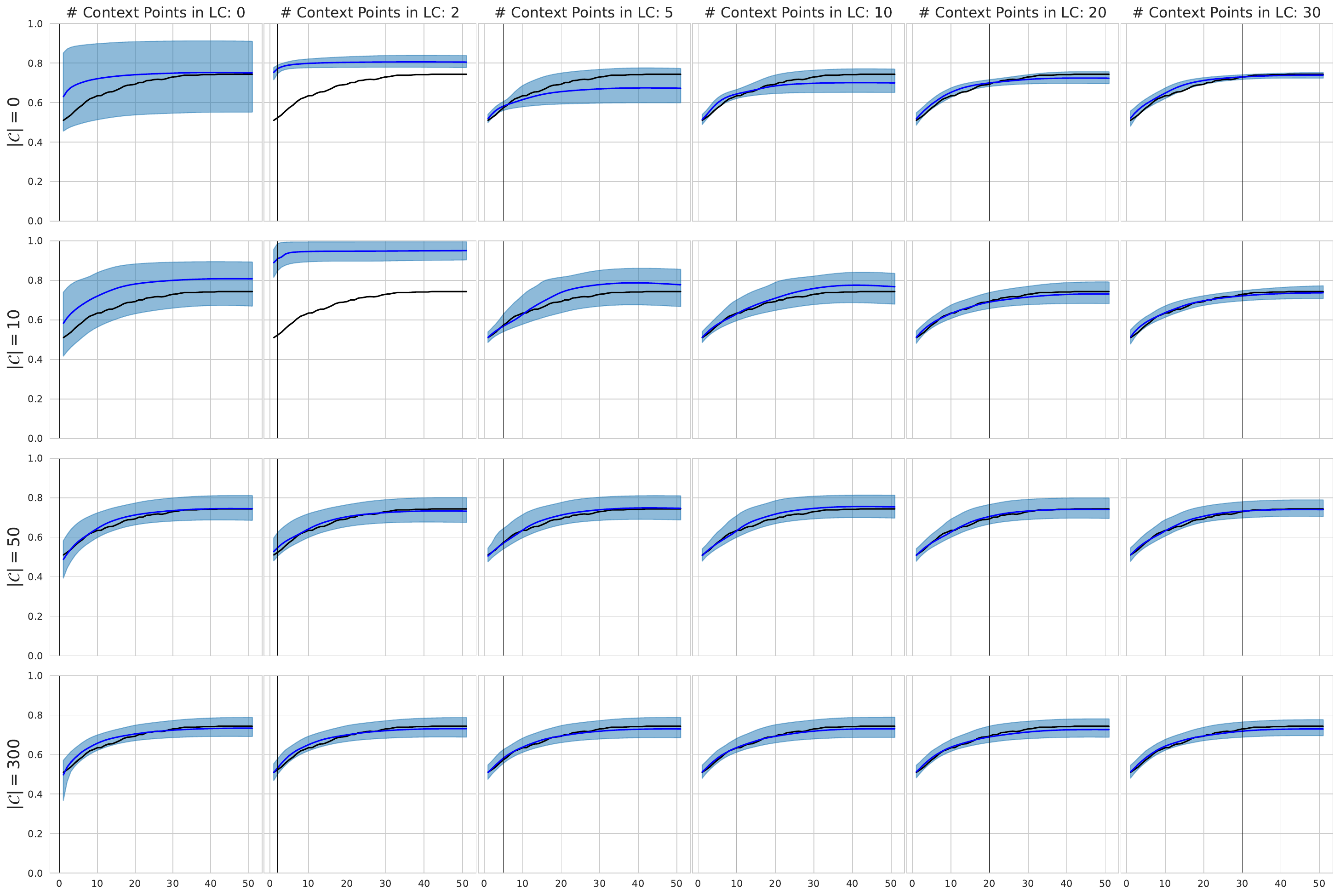}
\includegraphics[width=1.0\textwidth]{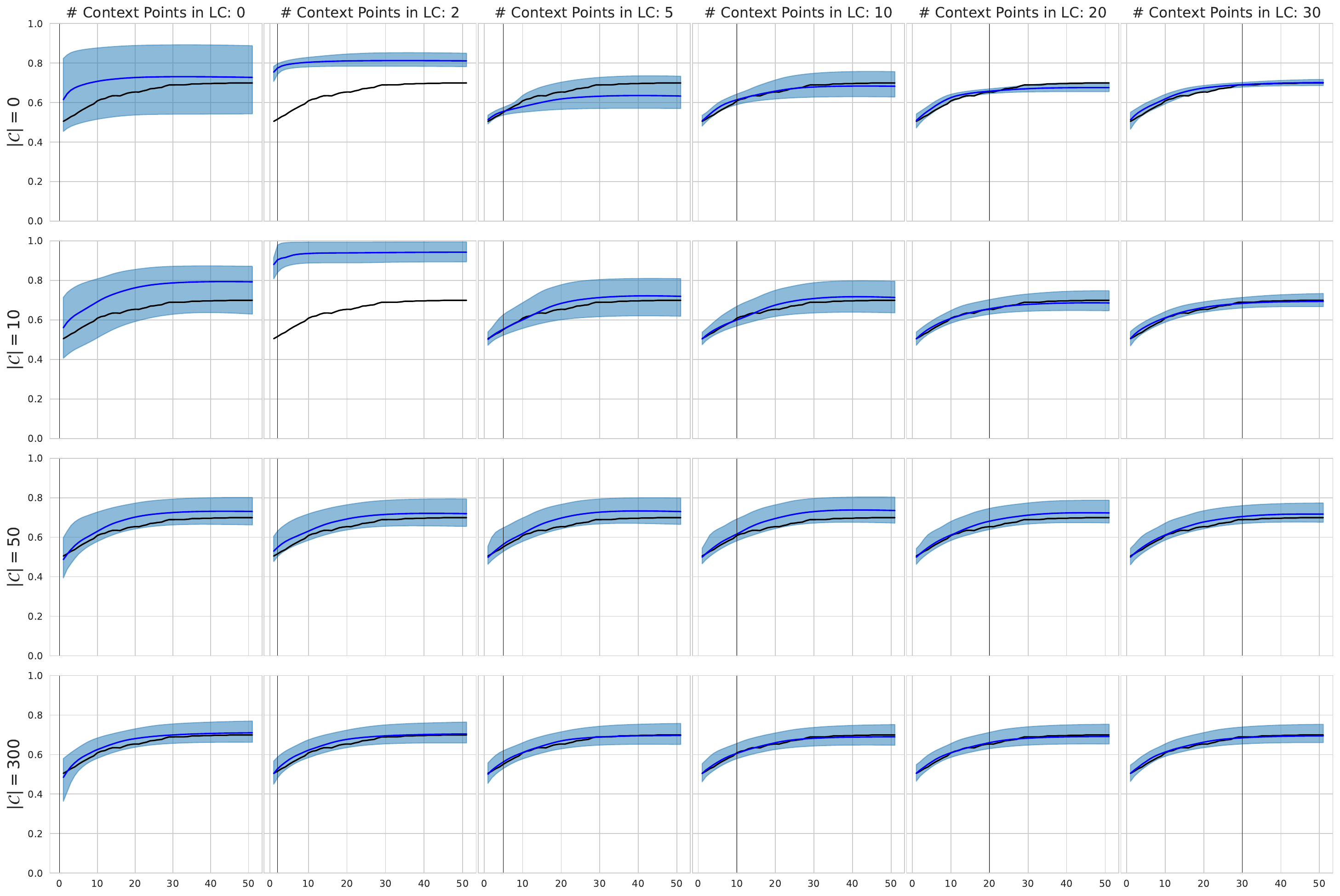}
\caption{\small Visualization of LC extrapolation over BO steps on \textbf{LCBench}.}
\label{fig:extrapolation_lcbench}
\vspace{-0.15in}
\end{figure}

\begin{figure}[H]
\vspace{-0.15in}
\centering
\includegraphics[width=1.0\textwidth]{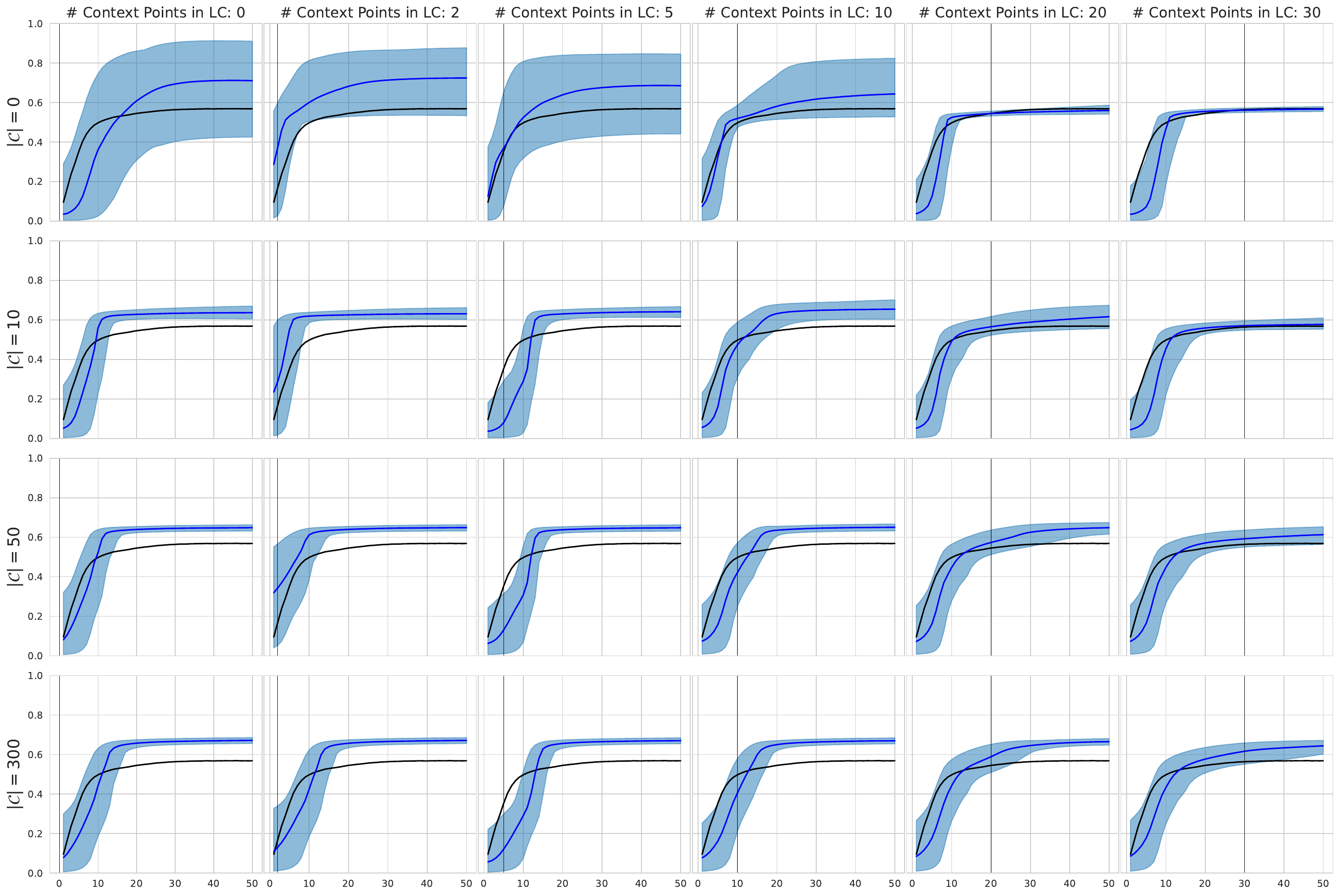}
\includegraphics[width=1.0\textwidth]{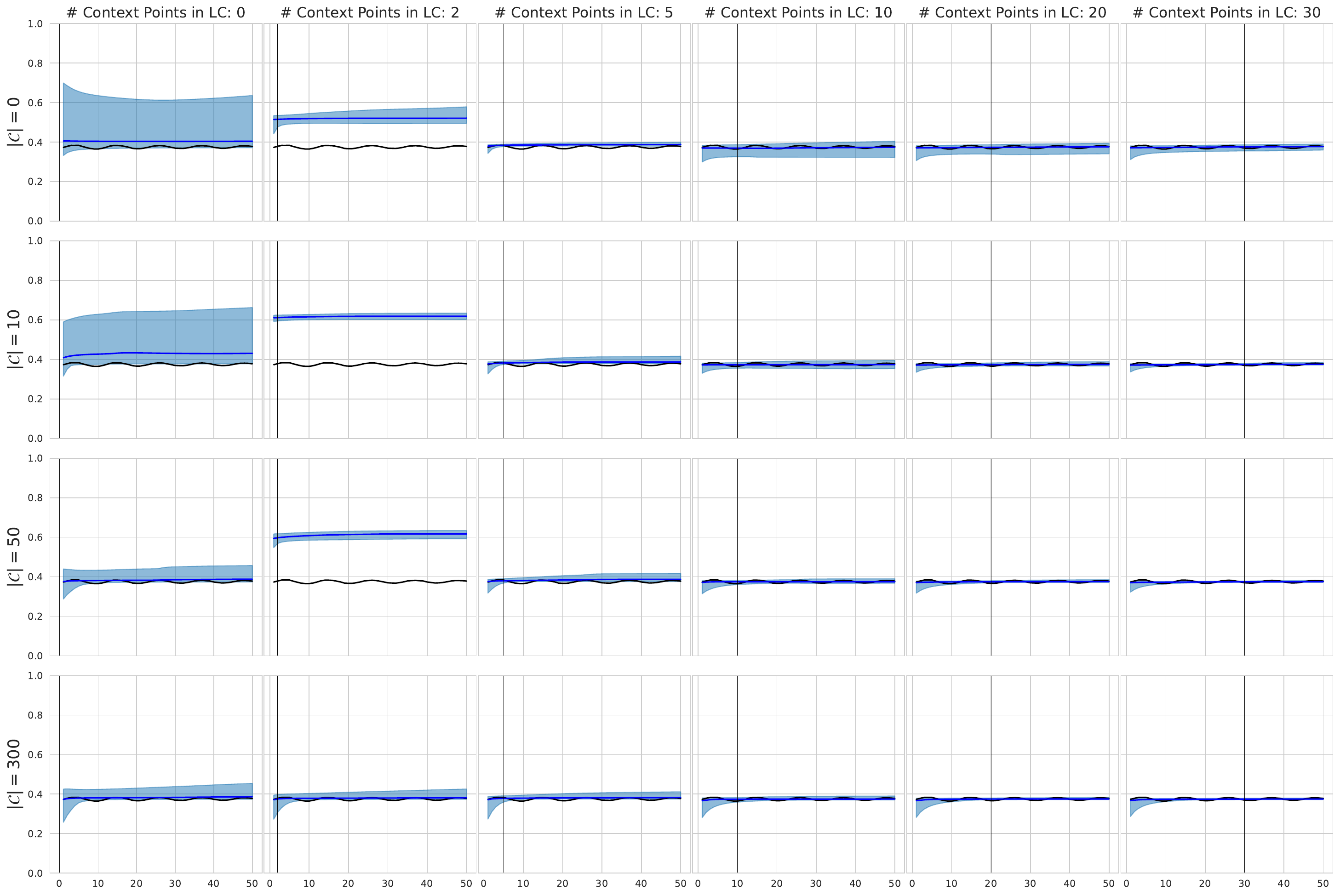}
\caption{\small Visualization of LC extrapolation over BO steps on \textbf{TaskSet}.}
\label{fig:extrapolation_taskset}
\vspace{-0.15in}
\end{figure}

\begin{figure}[H]
\vspace{-0.15in}
\centering
\includegraphics[width=1.0\textwidth]{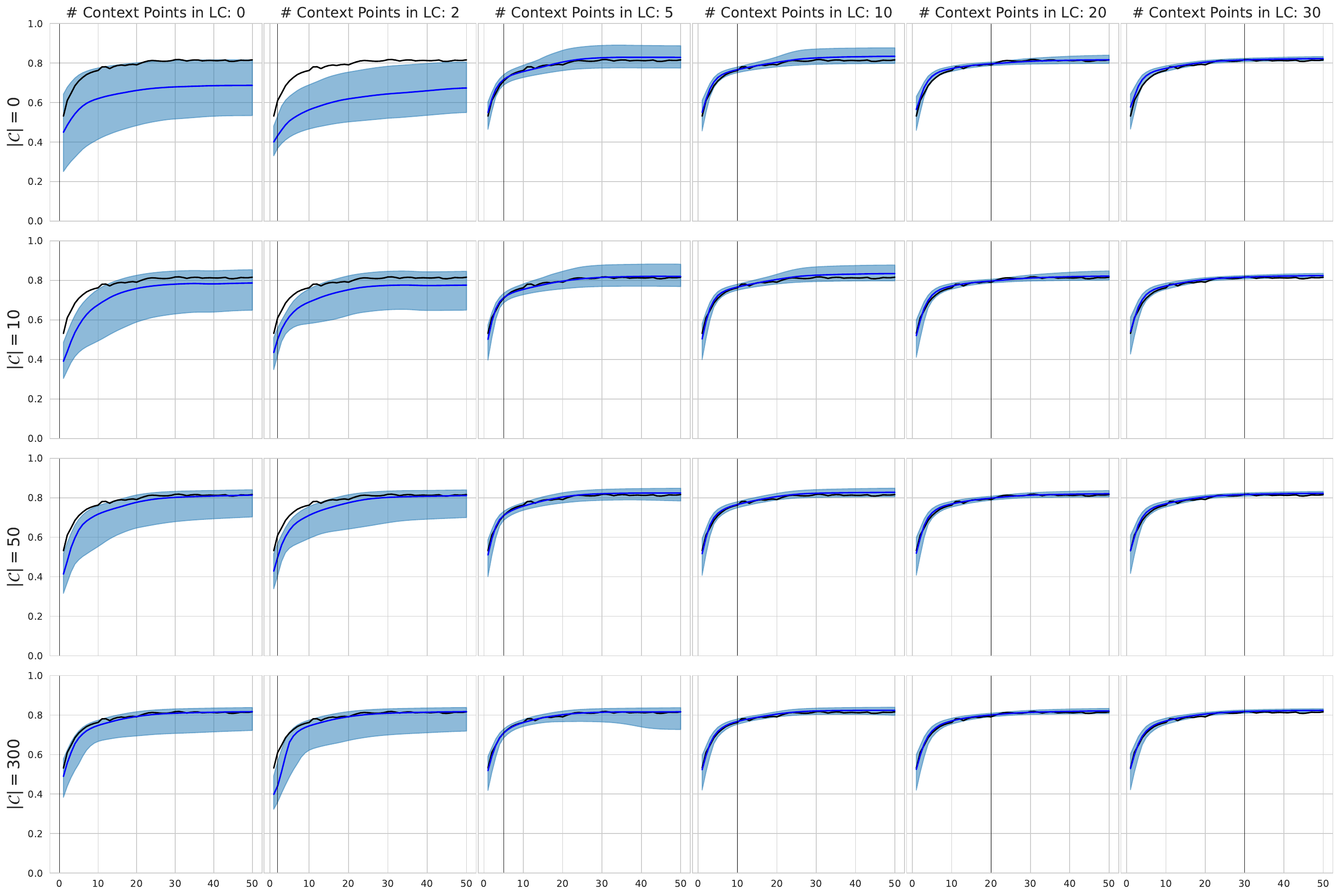}
\includegraphics[width=1.0\textwidth]{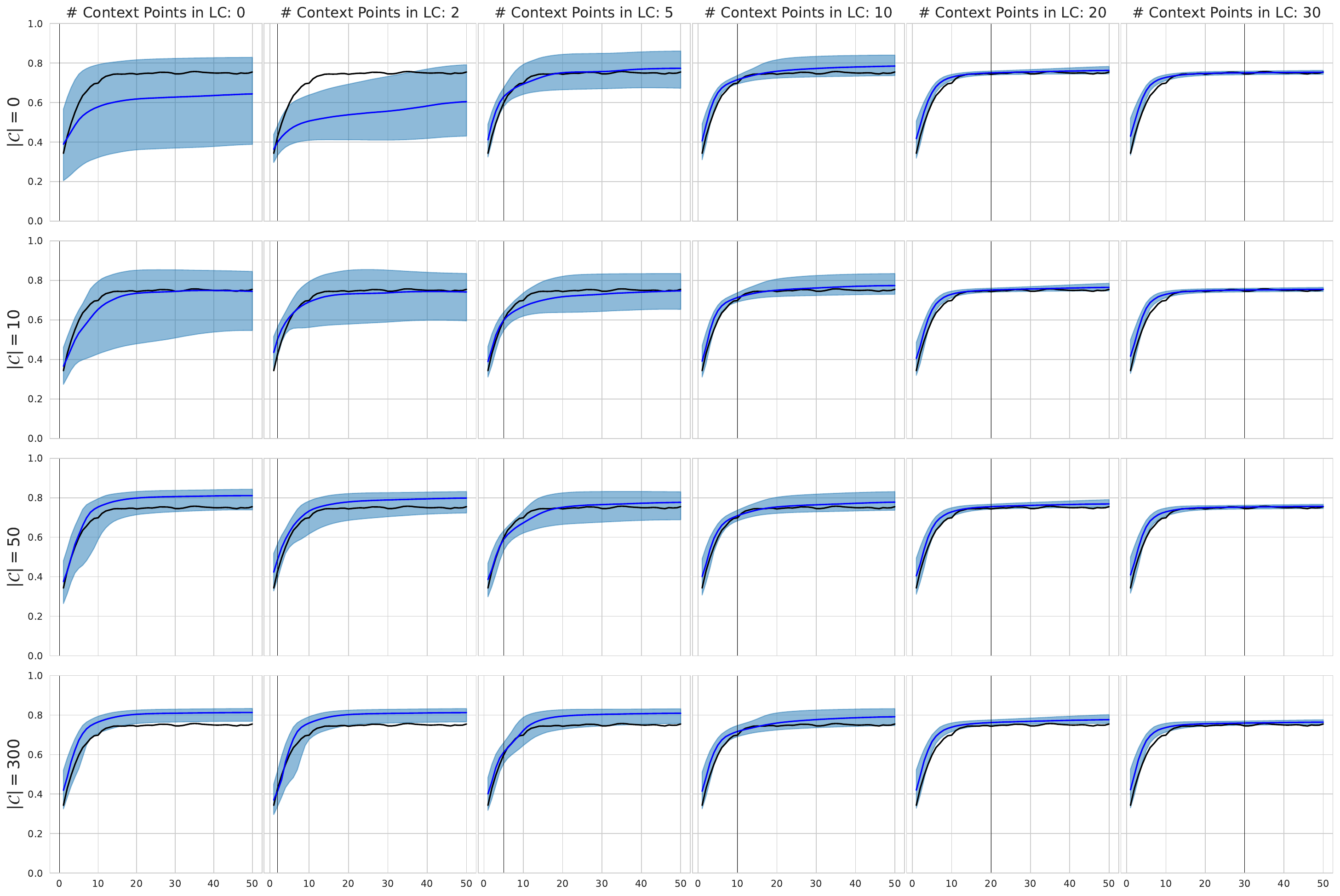}
\caption{\small Visualization of LC extrapolation over BO steps on \textbf{PD1}.}
\label{fig:extrapolation_pd1}
\vspace{-0.15in}
\end{figure}

\paragraph{Visualizations of the LC extrapolation over BO steps} for LCBench, TaskSet, and PD1 are provided Figure~\ref{fig:extrapolation_lcbench}, \ref{fig:extrapolation_taskset}, and \ref{fig:extrapolation_pd1}, respectively. Here, we plot the LC extrapolation results of unseen hyperparameter configurations through BO. Each row shows the results for a different size of the observation set ($|\mathcal{C}|=0, 10, 50$, and $300$), and each column shows a different size of context points in each LC (0, 2, 5, 10, 20, and 30).



\end{document}